\documentclass{article}

\PassOptionsToPackage{square, numbers}{natbib}
\usepackage[final]{neurips_2021}

\usepackage[utf8]{inputenc} 
\usepackage[T1]{fontenc}    
\usepackage{hyperref}       
\usepackage{url}            
\usepackage{booktabs}       
\usepackage{amsfonts}       
\usepackage{nicefrac}       
\usepackage{microtype}      

\usepackage{graphicx}
\usepackage{amsmath}
\usepackage[linesnumbered,ruled,vlined]{algorithm2e}
\SetKwInput{KwInput}{Input}  
\SetKwInput{KwOutput}{Output}
\SetKwInput{KwRequire}{Require}
\usepackage{multirow, subcaption}
\usepackage[flushleft]{threeparttable}
\usepackage{booktabs,caption}
\usepackage{wrapfig}
\usepackage[table]{xcolor}
\usepackage{booktabs}

\title{Revisit Multimodal Meta-Learning \\ through the Lens of Multi-Task Learning}

\author{%
 Milad Abdollahzadeh, Touba Malekzadeh, Ngai-Man Cheung \\
 Singapore University of Technology and Design\\
 \texttt{ \{milad\_abdollahzadeh, touba\_malekzadeh, ngaiman\_cheung\}@sutd.edu.sg
 }
 }

\begin{document}

\maketitle

\begin{abstract}

Multimodal meta-learning is a recent problem that extends conventional few-shot meta-learning by generalizing its setup to diverse multimodal task distributions. This setup 
makes a step towards mimicking 
how humans make use of a diverse set of prior skills to learn new skills. 
Previous work has achieved encouraging performance. In particular,  
in spite of the diversity of the multimodal tasks, 
previous work claims that a single meta-learner trained on a multimodal distribution can sometimes outperform multiple specialized meta-learners trained on individual unimodal distributions.
The improvement is attributed to knowledge transfer between different modes of task distributions. 
However, there is no deep investigation to verify and understand the knowledge transfer between multimodal tasks.
Our work makes two contributions to multimodal meta-learning.
First, we propose a method to {\em quantify knowledge transfer} between tasks of different modes at a micro-level. Our quantitative, task-level analysis is inspired by the recent transference idea from multi-task learning. 
Second, inspired by hard parameter sharing in multi-task learning and a new interpretation of related work, 
we propose  a {\em new multimodal meta-learner} that outperforms existing work by considerable margins.
While the major focus is on multimodal meta-learning, our work also attempts to shed light on task interaction in conventional meta-learning. The code for this project is available at \textcolor{magenta}{\url{https://miladabd.github.io/KML}}.

\end{abstract}

\section{Introduction}

{\bf Multimodal meta-learning} was recently proposed as an extension of conventional few-shot meta-learning.
In \cite{vuorio2019multimodal},  it is defined as a meta-learning problem that involves classification tasks from multiple different input and label domains. An example in their work is a 3-mode few-shot image classification which includes tasks to classify characters (Omniglot) and natural objects of different characteristics (FC100, mini-ImageNet). The multimodal extension of meta-learning is proposed with two objectives. First, it generalizes the conventional meta-learning setup for more diverse task distributions. Second, it 
makes a step towards mimicking
humans’ ability to acquire a new skill via prior knowledge of a set of diverse skills. For example, humans can quickly learn a novel snowboarding trick by exploiting not only fundamental snowboarding knowledge but also skiing and skateboarding experience \cite{vuorio2019multimodal}.

Multimodal Model-Agnostic Meta-Learning (MMAML) \cite{vuorio2019multimodal}
proposes a framework to better handle multimodal task distributions and achieves encouraging performance. As one of the most intriguing findings, MMAML claims that a single meta-learner trained on a multimodal distribution can sometimes outperform multiple specialized meta-learners trained on individual unimodal distributions. This was observed in spite of the diversity of the multimodal tasks.
In 
\cite{vuorio2019multimodal},
this observation is attributed to knowledge transfer across different modes of multimodal task distribution.

{\bf In our work,}
we delve into understanding knowledge transfer in multimodal meta-learning. While improved performance using a multimodal task distribution is reported in
\cite{vuorio2019multimodal}, there is no deep investigation to verify and understand how tasks from different modes benefit from each other. 
Towards understanding knowledge transfer in multimodal meta-learning at a micro-level, we propose a new quantification method inspired by the idea of {\em transference} recently proposed in multi-task learning (MTL)~\cite{fifty2020measuring}.
Despite the large number of meta-learning algorithms proposed in the literature, we remark that little work has been done in analyzing meta-learning at the 
task sample level. In particular, to the best of our knowledge, there is no previous work on understanding knowledge transfer among tasks and quantitative analysis of task samples in the context of meta-learning.  Because of the lack of such study, the interaction between task samples remains rather opaque in meta-learning.
Interestingly, despite the notion of a task in meta-learning and MTL, there has not been much intersections between these two branches of research, perhaps due to several fundamental differences between meta-learning and MTL\footnote{Here we follow the MTL definition in Zhang and Yang \cite{zhang2021survey} instead of a loose definition of MTL: learning problems that involve more that one task.} (e.g, meta-learning optimizes the risk over a large number of future tasks sampled from an unknown distribution of tasks, while MTL optimizes the average risk over a finite number of known tasks; this will be further discussed).
We note that because of these fundamental differences, we propose adaptations to develop our method to quantify knowledge transfer for multimodal tasks in meta-learning.

Another contribution of our work is a new multimodal meta-learner. Our idea is inspired by {\em hard parameter sharing} in MTL 
~\cite{ruder2017overview}.
Furthermore, we discuss our own
interpretation of the modulation mechanism in
\cite{vuorio2019multimodal}. These lead us to propose a new method that achieves substantial improvement over the best results in multimodal meta-learning \cite{vuorio2019multimodal}.
 While our major focus in this work is on multimodal meta-learning, we have also performed experiments on conventional meta-learning for our proposed knowledge transfer quantification. Our work makes an attempt to shed light on task interaction in conventional meta-learning.
 
{Our main contributions are
\begin{itemize}
    \item 
    Focusing on multimodal meta-learning, we propose a method to understand and quantify knowledge transfer across different modes at a micro-level.
    \item
    We propose a new multimodal meta-learner that outperforms existing state-of-the-art methods by substantial margins.
\end{itemize}
}

\begin{figure}[t]
\begin{center}
\includegraphics[ clip, trim={0.2cm 0.8cm 4.9cm 5.3cm}, width=14cm]{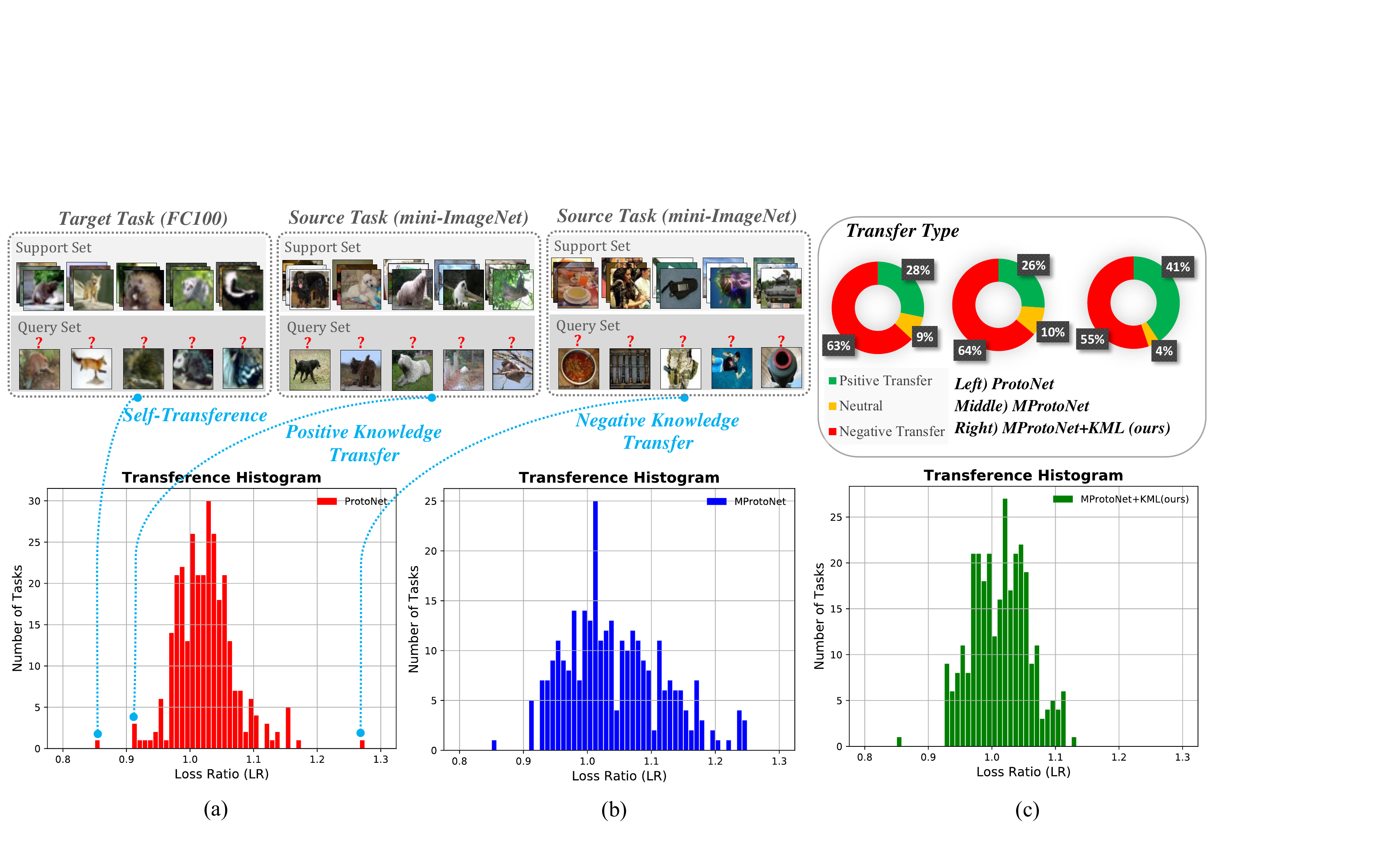}
   \caption{Information transfer (transference) from 300 meta-train mini-ImageNet tasks to a meta-test FC100 task. Transference Histogram for: (a) ProtoNet, (b) MProtoNet (with FiLM modulation), (c) MProtoNet with proposed KML method. For both positive knowledge transfer ($LR<1$), and negative knowledge transfer ($LR>1$) an exemplar task is shown. Proposed method increases the positive transfer from average of 27\% to 41\%. Here, we simply use the $LR$ threshold to classify the transference of a task as positive or negative.}  
   \label{fig:transference_analysis_1}
\end{center}
\end{figure}

\section{Related Work}
{\bf Few-Shot Learning.}
In a few-data regime, conventional learning methods mostly fail due to overfitting. Fine-tuning a pre-trained network ~\cite{sharif2014cnn, simonyan2014very, malekzadeh2017aircraft, Yan_2019_ICCV} sometimes prevents overfitting but at the cost of computation ~\cite{triantafillou2019meta}. Therefore, recent successful approaches tackle this problem by meta-learning ~\cite{thrun2012learning}. These methods can be classified into several categories. In \emph{metric-based} approaches, a similarity metric between support and query samples is learned by learning an embedding space, in which samples from similar classes are close and samples from different classes are further apart ~\cite{snell2017prototypical, vinyals2016matching, sung2018learning, chen2019closer, kye2020meta, ravichandran2019few}. \emph{Optimization-based} methods focus on learning an optimizer, including an LSTM meta-learner for replacing stochastic gradient descent optimizer ~\cite{Ravi2017OptimizationAA}, a mechanism to update weights using an external memory ~\cite{munkhdalai2017meta}, or finding a good initialization point for model parameters for fast adaptation ~\cite{finn2017model, rusu2018meta, li2017meta, Baik_2020_CVPR}. \emph{Augmentation-based} methods learn a generator from the existing labeled data to further use it for data augmentation in novel classes ~\cite{antoniou2017data, hariharan2017low}. Finally, \emph{weight-generation} methods directly generate the classification weights for unseen classes ~\cite{qiao2018few, guo2020attentive}.

{\bf Multi-Task Learning.} MTL algorithms can generally be divided into \emph{hard} or \emph{soft parameter sharing} methods ~\cite{ruder2017overview}. Hard parameter sharing is the most prevalent design to MTL where a subset of the hidden layers are shared among all tasks, while several task-specific layers are stacked on top of the shared base ~\cite{Caruana93multitasklearning}. Hard parameter sharing reduces the risk of overfitting and enables parameter efficiency across tasks ~\cite{zhang2021survey, baxter2000model}. In soft parameter sharing, each task has a separate model. Then, either the parameters of the models are encouraged to be similar by regularizing a distance ~\cite{duong2015low, yang2016trace} or the knowledge of tasks are linearly combined to produce output for each task ~\cite{misra2016cross}. Other works try to improve the multi-task performance by addressing \emph{what to share} ~\cite{guo2020learning, sun2019adashare, vandenhende2019branched}, \emph{which tasks to train together} ~\cite{bingel2017identifying, standley2020tasks}, or {\em inferring task-specifc model weights} ~\cite{yang2016deep, yang2016trace, maninis2019attentive}. Recently, the
\emph{transference}
is proposed to analyze the information transfer in a general MTL framework ~\cite{fifty2020measuring}. This metric is then used for selecting the tasks for updating shared parameters.

\section{Preliminaries}
In the \emph{N-way}, \emph{K-shot} image classification task, given \emph{N} classes and \emph{K} labeled examples per class, the final goal is to learn a model that  can generalize well to unseen examples from the same classes. Each task $\mathcal{T}$ consists of a \emph{support} set $\mathcal{S}_{\mathcal{T}}=\{(x_i, y_i) \}_{i=1}^{N \times K}$ and a query set $\mathcal{Q_{\mathcal{T}}} = \{(\Tilde{x}_i, \Tilde{y}_i)\}_{i=1}^{N\times M}$. The model learns from the samples in the support set $\mathcal{S}_{\mathcal{T}}$ and is evaluated on the query set $\mathcal{Q}_{\mathcal{T}}$. As a solution, meta-learning assumes that tasks are sampled from a distribution of tasks $\emph{p}(\mathcal{T})$, and are split into meta-training and meta-test sets. The meta-learner learns the prior knowledge about the underlying structure of tasks using the meta-training set, such that later, the meta-test tasks can benefit from this prior knowledge. While early meta-learning algorithms directly minimize the average training error of a set of training tasks, ~\citet{vinyals2016matching} propose a novel training strategy called \emph{episodic training}. Episodic training utilizes sampled mini-batches called \emph{episodes} during training, where each episode is designed to mimic the few-shot task and generated by subsampling of classes and samples from the training dataset. The use of episodes improves the generalization  by making the training process more faithful to the test environment and has become a dominant approach for the few-shot learning \cite{chen2020closer, finn2017model, snell2017prototypical}.  Meta-learning algorithms  usually suppose the task distribution $p(\mathcal{T})$ is \emph{unimodal}, meaning that all generated classification tasks belong to a single input-label domain (e.g.\ classification tasks with different combination of digits). Then, a \emph{multimodal} counterpart can be considered where it contains classification  tasks from multiple different input-label domains (e.g.\ few-shot digits classification and birds classification). In our work, multimodal meta-learning refers to the multimodality occurs in task distribution $p(\mathcal{T})$ due to using tasks from multiple domains. This should not be confused with multimodality in data type~\cite{baltruvsaitis2018multimodal} (e.g., combination of image and text).

\section{Information Transfer Among Tasks}

{\bf Transference in Episodic Training.}
The episodic training process can be viewed from a multi-task learning point of view, where multiple tasks (episodes) collaborate to build a shared feature representation that is broadly suitable for many tasks. Consequently, training episodes implicitly transfer information to each other by updating this shared feature representation with successive gradient updates. Then, the information transfer (transference) in episodic training can be viewed as the effect of gradients update from one episode (or a group of episodes) to the network parameters on the generalization performance on other episodes. Like MTL, in an episodic training scenario, some learning episodes can be constructive regarding a target task and some can be destructive ~\cite{liu2019loss}. 

{\bf Analysis of Transference.}
To gain insights on the interaction between different episodes during episodic training, we adapt the \emph{transference} idea ~\cite{fifty2020measuring} from multi-task learning to episodic training scenario of meta-learning. Consider a meta-learner parameterized by $\theta$. At time-step $t$ of episodic training, a batch of tasks are sampled from task distribution $p(\mathcal{T})$. Then for task $i$ denoted by $\mathcal{T}_i$, the model uses its support set $\mathcal{S}_{i}$ to adapt to it, and the quality of adaption is evaluated by the loss on query set $\mathcal{Q}_{i}$ denoted by $\mathcal{L}_{\mathcal{T}_{i}}$. The quantity $\theta^{t+1}_{i}$ is defined as updated model parameters after a SGD step with respect to task $i$:

\begin{equation}
\label{eq: model_update_task_i}
    \theta^{t+1}_{i} = \theta^{t} - \alpha \nabla_{\theta^{t}}
    \mathcal{L}_{\mathcal{T}_{i}}(\mathcal{Q}_{i}; \theta^{t}, \mathcal{S}_{i})
\end{equation}

We can use this quantity to calculate a  loss which reflects the effect of task $i$ on the performance of others. More specifically, in order to assess the effect of the gradient update of task $i$ on a given target task $j$, we calculate the ratio between loss of task $j$ before and after applying the gradient update on the shared parameters with respect to $i$:

\begin{equation}
    \label{eq:lookahead_loss}
    LR_{i \rightarrow j} = \frac{\mathcal{L}_{\mathcal{T}_{j}}(\mathcal{Q}_{j}; {\theta^{t+1}_{i}}, \mathcal{S}_{j})}{\mathcal{L}_{\mathcal{T}_{j}}(\mathcal{Q}_{j}; \theta^{t}, \mathcal{S}_{j})}
\end{equation}

\begin{algorithm}[t]
	\DontPrintSemicolon
	\SetAlgoLined
	
\KwRequire{task distribution $p(\mathcal{T})$, learning rate $\alpha$, current state of network parameters $\theta^{t}$}
		
Sample a batch of tasks $\xi = \{\mathcal{T}_i\} \sim p(\mathcal{T})$\\

Sample a target task $\mathcal{T}_{j}$\\

Use support set of target task $\mathcal{S}_j$ for adaption\\

Use query set $\mathcal{Q}_{j}$ to evaluate the adapted model and compute the loss ${\mathcal{L}_{\mathcal{T}_{j}}(\mathcal{Q}_{j}; \theta^{t}, \mathcal{S}_{j})}$\\ 

\For{all $\mathcal{T}_{i} \in \xi$}
    {
    Use support set $\mathcal{S}_{i}$ to adapt to that task\\
    
    Use query set $\mathcal{Q}_{i}$ to evaluate the adapted model and compute the loss $\mathcal{L}_{\mathcal{T}_{i}}(\mathcal{Q}_{i}; \theta^{t}, \mathcal{S}_{i})$\\
		    
	Update model parameters with respect to task $i$ using (\ref{eq: model_update_task_i})\\
	
	Compute the loss of target task using updated parameters $\mathcal{L}_{\mathcal{T}_{j}}(\mathcal{Q}_{j}; \theta^{t+1}_{i}, \mathcal{S}_{j})$\\
	
	Compute transference from task $i$ to target task $LR_{i \rightarrow j}$ using (\ref{eq:lookahead_loss})\\
		}

	\caption{Measuring Transference on a Target Task.}
	\label{alg:transference}
\end{algorithm}


In episodic training, the loss ratio $LR_{i \rightarrow j}$ can be considered as a measure of transference from meta-train task $i$ to meta-test task $j$. 
If $LR_{i \rightarrow j}$ has a value smaller than one, the update on the network parameters results in a lower loss on target task $j$ than the original parameter values, meaning that task $i$ has a constructive effect on the generalization of the model on task $j$. On the other hand, if $LR_{i \rightarrow j}$ has a value greater than one, it indicates the destructive effect of task $i$ on the target task. For $i=j$, the loss ratio denotes the \emph{self-transference}, i.e. the effect of a task's gradient update in its own loss which can be used as a baseline for transference.
One major limitation of the transference algorithm proposed in ~\cite{fifty2020measuring} is that while the knowledge transfer is measured by generalization performance, they have only considered the LR improvement on the target training tasks. This can not necessarily guarantee the improvement in the generalization performance. We adrress this by sampling source tasks from meta-train dataset and target tasks from meta-test dataset.
Then, our transference metric can be considered as a generalization metric in micro-level. So we expect the algorithms with better generalization to have a more positive knowledge transfer in terms of our transference metric, and vice versa.
The overall procedure of calculating transference within an episodic training framework is summarized in Algorithm \ref{alg:transference}. 

{\bf Experiments.}
To investigate the interaction between tasks during training in a multimodal task distribution, we have sampled 300 meta-train tasks from the mini-ImageNet dataset as source tasks. Then we have analyzed the transference from these tasks to a single FC100 meta-test task using algorithm \ref{alg:transference}. For this experiment, we have used ProtoNet ~\cite{snell2017prototypical} as meta-learner, and the analysis is performed in the middle of training on the combination of mini-ImageNet and FC100 datasets. The details of the experimental setup can be found in the supplementary. The histogram of the transference is shown in Figure \ref{fig:transference_analysis_1}a which indicates both positive and negative knowledge transfer from mini-ImageNet tasks to the target task. An exemplar task for both positive and negative knowledge transfer is shown in figure \ref{fig:transference_analysis_1}. The task including animal classes has positive knowledge transfer to target task while the task including non-animal classes has negative transfer. 

\begin{figure*}
\begin{center}
\includegraphics[ clip, trim={2cm 2.4cm 0.7cm 3cm}, width=14cm]{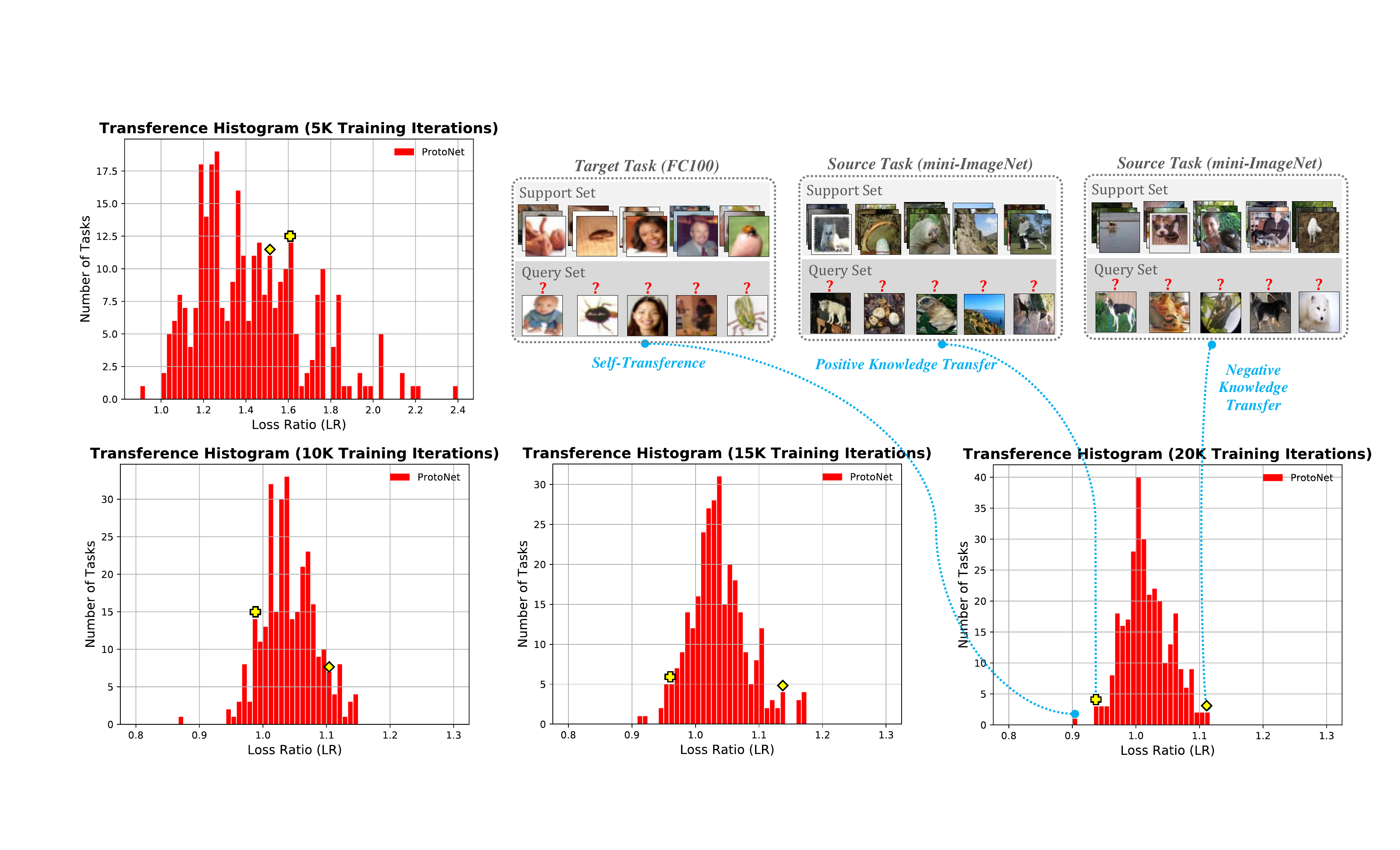}
   \caption{Transference from 300 mini-ImageNet meta-train tasks to a FC100 meta-test task. In the beginning of the training, when network learns low-level features, both tasks transfer negative knowledge. As the training proceeds, one becomes more and more positive, and then, consistently transfers positive knowledge to target task.}  
   \label{fig:transference_analysis_2}
\end{center}
\end{figure*}

In figure \ref{fig:transference_analysis_2}, the transference to a different meta-test FC100 target task from mini-ImageNet meta-train tasks is shown. While the target task includes classification from people and insect classes, two source tasks with animal classes are among the best and worst knowledge transferring source tasks. This can be attributed to the quality of samples in these tasks. When a task includes noisy data samples, it is much harder to solve meaning that the transference can also happen based on task hardness ~\cite{tran2019transferability}. Figure \ref{fig:transference_analysis_2} also indicates that in the cross mode knowledge transfer, the negative transference occurs at the beginning iterations and increasingly more positive transference occurs as training proceeds. Based on the experience from MTL literature, the negative knowledge transfer occurs when different tasks fight for the capacity ~\cite{liu2019loss}. In the next section we will propose a new modulation scheme to reduce negative transfer and improve generalization (Figure \ref{fig:transference_analysis_1}c).

\section{Proposed Multimodal Meta-Learner}
In the previous section, the transference analysis shows that different tasks in episodic training can have a positive or negative impact on the learning process of other tasks. Recent works have analyzed the compatibility of tasks in multi-task learning ~\cite{standley2020tasks, fifty2020measuring} during training to select some grouping of tasks for co-training that can improve performance. In our episodic training scenario, ideally, we would like to select the grouping of learning episodes that are compatible with the meta-test task. However, direct application of the methods in MTL to episodic training is not possible, due to two main reasons. \emph{First}, the novel tasks in meta-testing are unseen and unknown during meta-training. \emph{Second,} even if the tasks in meta-testing were known, it can be very computationally expensive to determine the group of cooperative tasks, and then assign a task-specific layer for each task, because episodic training involves tens of thousands of training tasks. Here, we propose a new interpretation of the modulation mechanism in MMAML. These lead us to propose pseudo-task-specific (task-aware) layers inspired by hard parameter sharing in MTL. 

\begin{figure*}
\begin{center}
\includegraphics[ clip, trim={1cm 4.2cm 3.2cm 3.1cm}, width=14cm]{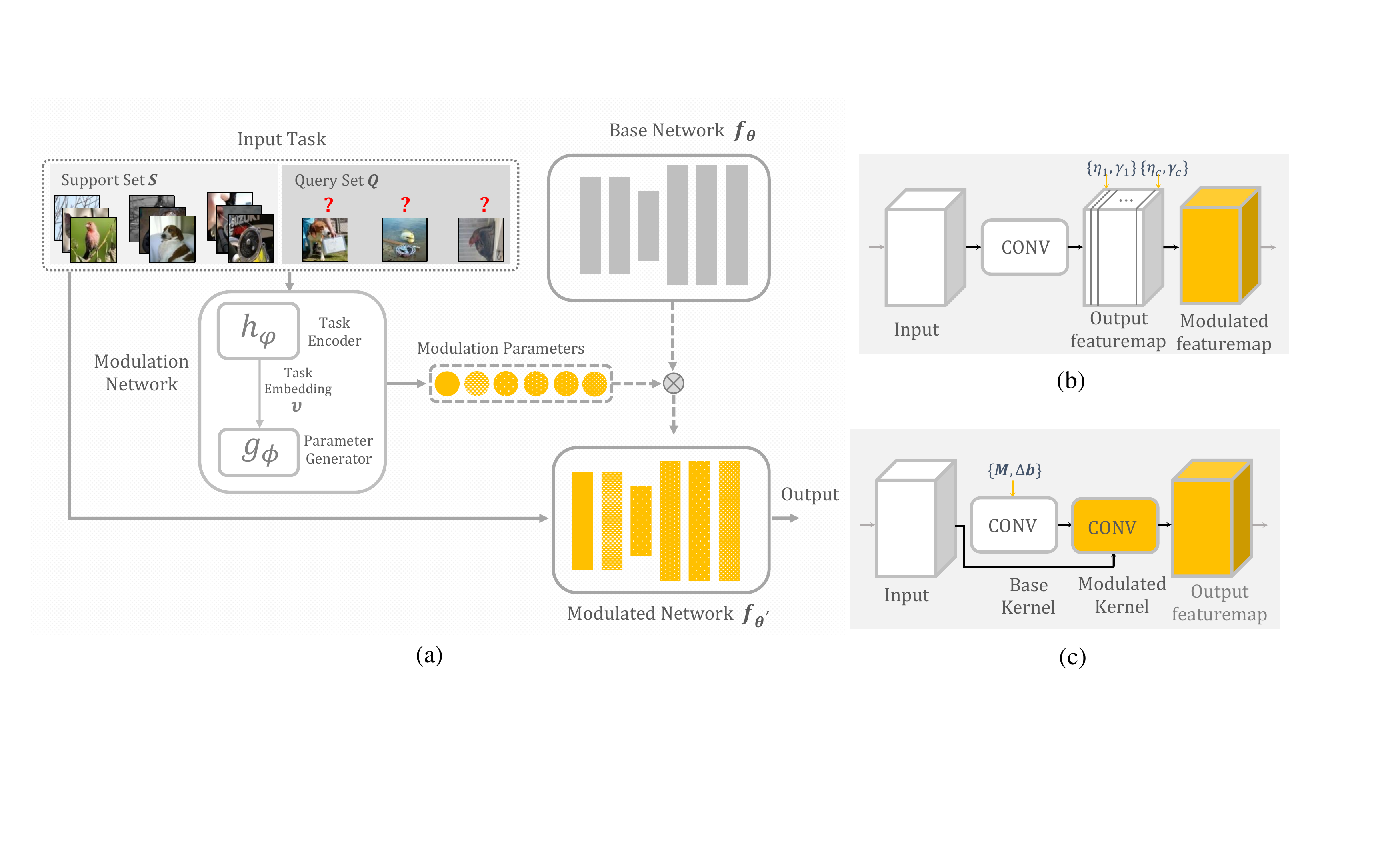}
   \caption{(a) General multimodal meta-Learning framework; Modulation scheme in (b) MMAML, (c) Proposed KML method. }  
   \label{fig:KML_Algorithm}
\end{center}
\end{figure*}

\subsection{Multimodal Model-Agnostic Meta-Learning}

To tackle multimodal few-shot tasks, MMAML ~\cite{vuorio2019multimodal} proposes a multimodal meta-learning  framework which consists of a modulation network and a base network $f_{\theta}$ (Figure \ref{fig:KML_Algorithm}a). The modulation network, predicts the mode of a task and generates a set of task-specific modulation parameters to help the base network to better fit to the identified mode. It includes a task encoder network and a modulation parameter generator. Task encoder takes the support samples of a task $\mathcal{T}$ as input and produces an embedding vector $\boldsymbol{\upsilon}_{\mathcal{T}} = h_{\varphi}(\mathcal{S}_{\mathcal{T}})$ to encode its characteristics. Then, task-specific modulation parameters are generated using the embedding vector of task by the parameter generator network $\boldsymbol{\omega}_{\mathcal{T}} = g_{\phi}(\boldsymbol{\upsilon}_{\mathcal{T}})$. Since MMAML uses feature-wise linear modulation (FiLM)~\cite{perez2018film}, the generated modulation parameters are split into scaling and shifting parameters $\boldsymbol{\omega}_{\mathcal{T}} = \left \{ \boldsymbol{\eta}_{\mathcal{T}}, \boldsymbol{\gamma}_{\mathcal{T}} \right \} $. Let $\mathbf{Y}_{i}$ denote the \emph{i\textsuperscript{th}}  output channel of a layer in the base network $f_{\theta}$. Then, the corresponding modulated output channel in MMAML is calculated as:

\begin{equation}
   \label{eq:FiLM}
   \hat{\mathbf{Y}}_{i} = \eta_{i}\mathbf{Y}_{i}+\gamma_{i}
\end{equation}

where $\eta_{i} \in \boldsymbol{\eta}_{\mathcal{T}}$ and $\gamma_{i} \in \boldsymbol{\gamma}_{\mathcal{T}}$ are scalar values. For task encoder, MMAML uses a 4-layer convolutional network which is followed by multiple MLPs as parameter generator. MMAML uses the model-agnostic meta-learning (MAML) algorithm ~\cite{finn2017model} as meta-learner in base network.

{\bf Limitations of Feature-Wise Modulation.} We believe the major limitation of MMAML results from using FiLM modulation scheme. For each convolutional layer of a CNN, the \emph{i\textsuperscript{th}} output channel (feature map) is computed by convolving the \emph{i\textsuperscript{th}} kernel of that layer $\mathbf{W}_{i}$ (the layer index is removed for the sake of simplicity) with input to the layer $\mathbf{X}$ and adding the bias term $b_{i}$:

\begin{equation}
   \label{eq:convolution}
   \mathbf{Y}_{i} = \mathbf{W}_{i} * \mathbf{X} + b_{i}
\end{equation}

where $*$ denotes the convolution operator. Considering this, the modulation in (\ref{eq:FiLM}) can be rewritten as:

\begin{equation}
\label{eq:interpretation}
\hat{\mathbf{Y}}_{i} = (\eta_{i}\mathbf{W}_{i}) * \mathbf{X} + (\eta_{i}b_{i} + \gamma_{i})
\end{equation}

We define the modulated kernel, and the modulated bias as $\hat{\mathbf{W}}_{i} = \eta_{i}\mathbf{W}_{i}$, and $\hat{b}_{i} = \eta_{i} b_{i} + \gamma_{i}$. Note that the modulated kernel is calculated by scaling the whole elements of the original kernel with a single scalar number. Then considering (\ref{eq:FiLM})-(\ref{eq:interpretation}), the modulation in MMAML can be interpreted as convolving the input with this modulated kernel and adding the modulated bias term. The validity of this new interpretation is verified with experimental results (please refer to supplementary material). Based on this new interpretation, the modulated kernels used for different tasks are heavily bound together by $\mathbf{W}$ and are just scaled versions of it. Since the base kernel is shared across all tasks and scaling gives not many degrees of freedom for each task to produce its desired modulated kernel, different tasks fight for the model capacity, and the negative transfer between tasks from different distributions can lead to training a compromised base kernel.

\begin{figure*}[ttt!]
\begin{tabular}{cc}

\scalebox{0.90}{
\begin{minipage}{0.66\textwidth}
\begin{algorithm}[H]
\captionof{algocf}{Kernel Modulation Algorithm} \label{alg:1}
\DontPrintSemicolon
\SetAlgoLined
	
\KwRequire{task distribution $p(\mathcal{T})$, learning rate $\alpha$}
Randomly initialize $\theta$, $\varphi$, $\phi$\\
\While{not done}
	{
		Sample batch of tasks $\mathcal{T}_i \sim p(\mathcal{T})$\\
		\For{all $\mathcal{T}_{i}$}{
		    Infer $\boldsymbol{\upsilon}_{\mathcal{T}_i} = h_{\varphi}(\mathcal{S}_{i})$\\
		    
		    Generate $M_{\mathcal{T}_i}^{(l)}$ and $\Delta b_{\mathcal{T}_i}^{(l)}$ using (\ref{eq:modulation_matrix_generator}) and (\ref{eq:bias_generator})\\
		    
		    Modulate $\theta$ using (\ref{eq:kernel_mod}) and (\ref{eq:bias_mod}) to obtain $\hat{\theta}_{\mathcal{T}_{i}}$\\
		    
			Compute $\mathcal{L}_{\mathcal{T}_{i}}(\mathcal{Q}_{i};\hat{\theta}_{\mathcal{T}_{i}}, \mathcal{S}_{i})$ using $f_{\hat{\theta}_{\mathcal{T}_{i}}}$\\
			
		}
		
		Update $\theta \leftarrow \theta - \alpha \nabla_{\theta} \sum_{\mathcal{T}_{i}\sim p(\mathcal{T})} \mathcal{L}_{\mathcal{T}_{i}}(\mathcal{Q}_{i};\hat{\theta}_{\mathcal{T}_{i}}, \mathcal{S}_{i})$\\
		
		Update $\varphi \leftarrow \varphi - \alpha \nabla_{\varphi} \sum_{\mathcal{T}_{i}\sim p(\mathcal{T})} \mathcal{L}_{\mathcal{T}_{i}}(\mathcal{Q}_{i};\hat{\theta}_{\mathcal{T}_{i}}, \mathcal{S}_{i})$\\
		
		Update $\phi \leftarrow \phi - \alpha \nabla_{\phi} \sum_{\mathcal{T}_{i}\sim p(\mathcal{T})} \mathcal{L}_{\mathcal{T}_{i}}(\mathcal{Q}_{i};\hat{\theta}_{\mathcal{T}_{i}}, \mathcal{S}_{i})$\\
	}
\end{algorithm}
\end{minipage}} &
\hfill
\begin{minipage}{0.30\textwidth}
\begin{subfigure}{.35\textwidth}
  \includegraphics[ clip, trim={1.85cm 18cm 9cm 1.65cm}, width=5cm]{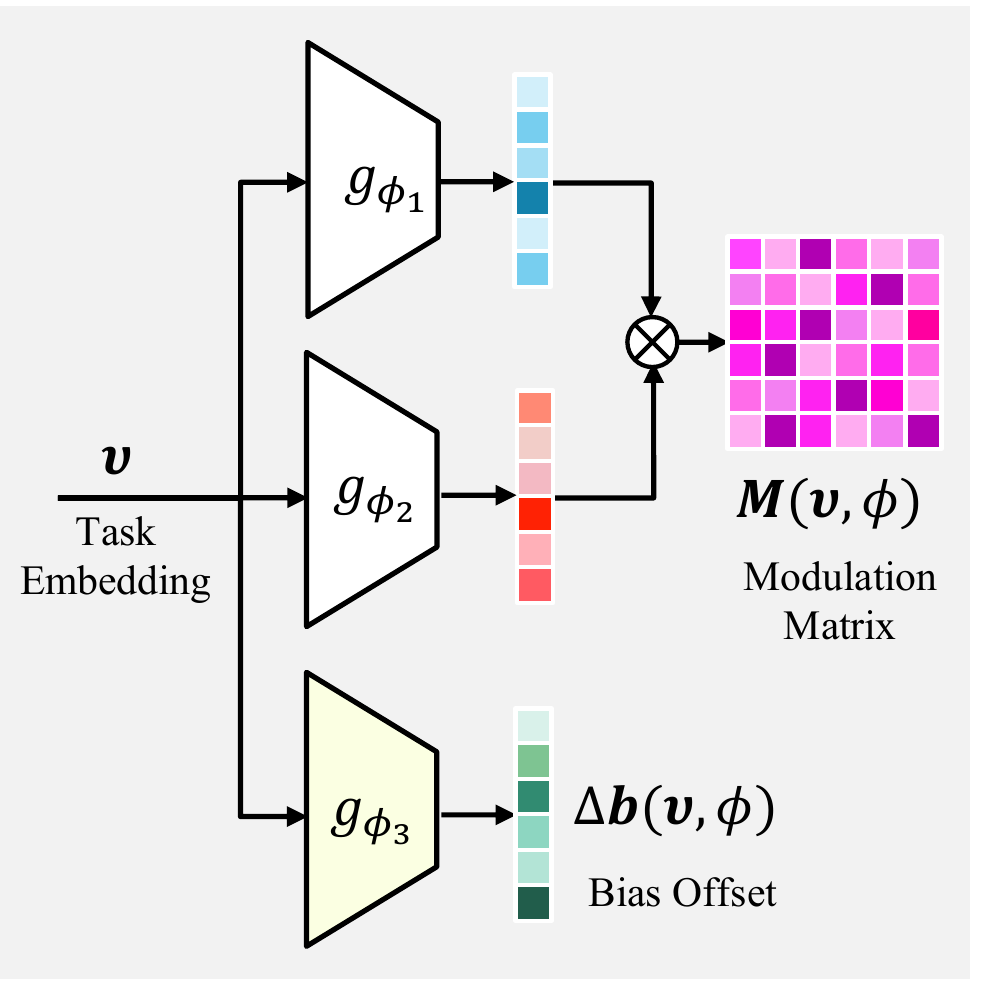}
  \label{fig:sfig1}
\end{subfigure}

\captionof{figure}{\small Simplified design for modulation parameter generator.} \label{fig:modulation_parameter_generator}
\end{minipage}

\end{tabular}
\end{figure*}

\subsection{Kernel Modulation}
 
We propose to modulate the base network by changing the whole parameters within each layer using  the modulation parameters conditioned on the task. This gives more capacity for each task to generate its desired kernel. To do this, for every single parameter within the network, a modulation parameter is generated. For a task $\mathcal{T}$, and for each layer $l = 1, \dots, L$ of the network, let $\mathbf{M}^{l}(\boldsymbol{\upsilon}_{\mathcal{T}}, \phi)$ denote the modulation matrix generated for modulating the base kernel $\mathbf{W}^{l}$ of that layer. Among different design choices, intuitively, if we consider the modulated kernels as perturbations around the base kernel, by learning to generate the residual weights, a more powerful base kernel can be learned via enabling proper knowledge transfer across different modes of the task distribution. In our case, this can be achieved by adding the unit matrix $\mathbf{J}$ (all-ones matrix) to the modulation matrix and then perform Hadamard multiplication between the result and the base kernel as follows:
 
\begin{equation}
    \label{eq:kernel_mod}
    \hat{\mathbf{W}}_{\mathcal{T}}^{l} = \mathbf{W}^{l} \odot (\mathbf{J} + \mathbf{M}^{l}(\boldsymbol{\upsilon}_{\mathcal{T}}, \phi))
\end{equation}
 
For the bias of each layer we simply add an offset term $\Delta \mathbf{b}^{l}(\boldsymbol{\upsilon}_{\mathcal{T}}, \phi)$:

\begin{equation}
    \label{eq:bias_mod}
    \hat{\mathbf{b}}_{\mathcal{T}}^{l} = \mathbf{b}^{l} + \Delta \mathbf{b}^{l}(\boldsymbol{\upsilon}_{\mathcal{T}}, \phi) 
\end{equation}

The modulated parameters for task $\mathcal{T}$ are therefore a collection of modulated kernels and biases, i.e.,
$\hat{\theta}_{\mathcal{T}}=\{ \hat{\mathbf{W}}_{\mathcal{T}}^{1}, \dots, \hat{\mathbf{W}}_{\mathcal{T}}^{L}, \hat{\mathbf{b}}_{\mathcal{T}}^{1} , \dots,
\hat{\mathbf{b}}_{\mathcal{T}}^{L}
\}$.
The modulation parameters for each task are generated through processing task embeddings of that task $\boldsymbol{\upsilon}_{\mathcal{T}}$ by parameter generator network $\mathbf{g}_{\phi}$ (Figure \ref{fig:KML_Algorithm}). Similar to ~\cite{vuorio2019multimodal}, assuming $\mathbf{g}_{\phi}$ to be MLP, saves lots of computations where a separate parameter generation network $\mathbf{g}_{\phi}^{l}$ used for each layer. The main concern with using an MLP is the large number of parameters required in $\mathbf{g}_{\phi}^{l}$ to generate the modulation parameters $\Delta \mathbf{b}^{l}$ and $ \mathbf{M}^{l}$. Because we want to produce the modulation parameters for the whole parameters of a deep neural network, we need to employ some sort of parameter reduction in $\mathbf{g}_{\phi}$ to prevent an explosion of parameters. Instead of following current complicated algorithms such as pruning the weights of network ~\cite{blalock2020state} or redesigning the operations ~\cite{Sandler_2018_CVPR}, we propose a simple design for $\mathbf{g}_{\phi}$ which substitutes an MLP module with three smaller ones (Figure \ref{fig:modulation_parameter_generator}). In this design, the modulation matrix is generated by processing task embeddings $\boldsymbol{\upsilon}_{\mathcal{T}}$ using two modules, and then performing the outer product operation on the output of these two modules:

\begin{equation}
    \label{eq:modulation_matrix_generator}
    \mathbf{M}^{l}(\boldsymbol{\upsilon}_{\mathcal{T}}, \phi) = {\mathbf{g}_{\phi_{1}}^{l}(\boldsymbol{\upsilon}_{\mathcal{T}})} \otimes  {\mathbf{g}_{\phi_{2}}^{l}(\boldsymbol{\upsilon}_{\mathcal{T}})}
\end{equation}

The produced matrix is then reshaped to match the kernel shape in that layer. The bias offset of each layer is generated by processing the embedding vector $\boldsymbol{\upsilon}_{\mathcal{T}}$ using the third module:

\begin{equation}
    \label{eq:bias_generator}
    \Delta \mathbf{b}^{l}(\boldsymbol{\upsilon}_{\mathcal{T}}, \phi) = \mathbf{g}_{\phi_{3}}^{l}(\boldsymbol{\upsilon}_{\mathcal{T}})
\end{equation}

Our experiments show that compared to a single MLP, the proposed simplified structure decreases the number of parameters in $\mathbf{g}_{\phi}$ by a factor of {\bf 150}, and also provides better generalization performance. A detailed discussion can be found in supplementary. A similar approach to our simplified structure for parameter generator is proposed in ~\cite{simon2020modulating}. However in ~\cite{simon2020modulating}, it is used to filter out noisy gradients rather than generating task-aware parameter for few-shot learning done in our work.
The overall multimodal meta-learning algorithm using the proposed \textbf{K}ernel {\bf M}odu{\bf L}ation (KML) scheme is summarized in Algorithm \ref{alg:1}. 
Note that considering the new interpretation presented in (\ref{eq:interpretation}), our KML can be thought of as a generalization of FiLM. So we can expect that applying KML to the areas improved by FiLM may bring some further improvements. In addition to the few-shot classification results provided in section \ref{section:experiments}, we have also provided some results on visual reasoning in the supplementary which shows better performance of proposed KML over FiLM.

\section{Experiments and Analysis}
\label{section:experiments}
We evaluate the proposed model in both multimodal and unimodal few-shot classification scenarios. Following ~\cite{vuorio2019multimodal}, to create a multimodal few-shot image classification meta-dataset, we combine multiple widely used datasets (Omniglot ~\cite{lake2011one}, mini-Imagenet ~\cite{vinyals2016matching}, FC100 ~\cite{oreshkin2018tadam}, CUB ~\cite{wah2011caltech}, and Aircraft ~\cite{maji2013fine}). Most of our experiments were performed by modifying the code accompanying ~\cite{vuorio2019multimodal}, and for a fair comparison, we follow their experimental protocol unless specified. The details of all datasets, experimental setups, and network structure can be found in the supplementary material. Note that when using an optimization-based meta-learner ($f_\theta$) like MAML, our KML idea can easily be extended to Reinforcement Learning (RL) environments. More details and experimental results on RL can be found in supplementary.

\begin{table*}
\caption{Meta-test accuracies on the multimodal few-shot image classification with 2, 3, and 5 modes. The accuracy values are mean of 1000 randomly generated test tasks, and the $\pm$ shows 95\% confidence interval over tasks. 2Mode\textsuperscript{\dag} and 2Mode indicate the combination of mini-ImageNet with FC100 and Omniglot, respectively.  \textsuperscript{*} Results produced by code provided in ~\cite{vuorio2019multimodal}.  \textsuperscript{**} Our Implementation.}
	\label{table:Multimodal_Classification}
\begin{threeparttable}
	\begin{center}
		\resizebox{0.95\linewidth}{!}{
			\begin{tabular}{llcccc}
				\toprule
				\multirow{2}{*}{\textbf{Setup}} &
				\multirow{2}{*}{\textbf{}} &
				\multicolumn{4}{c}{\textbf{Method}} \\
				\cmidrule(r){3-6}
				& & \textbf{MAML ~\cite{finn2017model}}\textsuperscript{*}  & \textbf{Multi-MAML} & \textbf{MMAML~\cite{vuorio2019multimodal}}\textsuperscript{*} & \textbf{MMAML+KML (ours)} \\ 
				\midrule
				
				\multirow{2}{*}{\textbf{2Mode\textsuperscript{\dag}}} & 1-shot & 
				{40.53$\pm$68\%}&
				39.27$\pm$0.76\%& 
				39.11$\pm$0.62\%&
				\textbf{40.73$\pm$0.66\%}\\
				
				\arrayrulecolor{black!30}
				\cmidrule(r){3-6}
				&
				{5-shot} &
				\textbf{54.11$\pm$0.63\%}&
				{53.51$\pm$0.72\%}& 
				52.02$\pm$0.63\%&
				{53.72$\pm$0.60\%}\\
				\arrayrulecolor{black!100}\midrule
				
				\multirow{2}{*}{\textbf{2Mode}} &
				1-shot &
				65.18$\pm$0.61\%&
				66.77$\pm$0.68\%&
				67.67$\pm$0.63\%&
				\textbf{68.01$\pm$0.59\%}\\
				\arrayrulecolor{black!30}
				\cmidrule(r){3-6}
				
				&
				{5-shot} &
				74.18$\pm$0.57\%&
				73.07$\pm$0.61\%& 
				73.52$\pm$0.71\%&
				\textbf{77.02$\pm$0.66\%}\\
				\arrayrulecolor{black!100}\midrule

				\multirow{2}{*}{\textbf{3 Mode}} &
				{1-shot} &
				54.40$\pm$0.56\%&
				56.01$\pm$0.66\%& 
				57.35$\pm$0.61\%& 
				\textbf{57.68$\pm$0.59\%}\\
				
				\arrayrulecolor{black!30}
				\cmidrule(r){3-6}
				
				& {5-shot} &
				66.51$\pm$0.54\%&
				65.92$\pm$0.62\%& 
				64.21$\pm$0.57\%& 
				\textbf{67.12$\pm$0.55\%}\\
				
				\arrayrulecolor{black!100}\midrule
				
				{\textbf{5Mode}} &
				{1-shot} &
				47.19$\pm$0.49\%&
				48.33$\pm$0.58\%& 
				49.53$\pm$0.50\%& 
				\textbf{50.31$\pm$0.49\%}\\
				
				\arrayrulecolor{black!30}\cmidrule(r){3-6}
				
				&
				{5-shot} &
				58.13$\pm$0.48\%&
				59.20$\pm$0.52\%& 
				58.89$\pm$0.47\%&
				\textbf{60.51$\pm$0.47\%}\\
				
				\arrayrulecolor{black!100}
				\bottomrule
				\rule{0pt}{2.5ex}  
				& & \textbf{ProtoNet ~\cite{snell2017prototypical}}\textsuperscript{**} &
				\textbf{Multi-ProtoNet} & \textbf{MProtoNet~\cite{vuorio2019multimodal}}\textsuperscript{**} & \textbf{MProtoNet+KML (ours)} \\
				\midrule
				
				\multirow{2}{*}{\textbf{2Mode\textsuperscript{\dag}}} &
				{1-shot} &
				43.05$\pm$0.58\%&
				43.42$\pm$0.56\%&
				43.57$\pm$0.59\%&
				\textbf{44.40$\pm$0.65\%}\\
				
				\arrayrulecolor{black!30}\cmidrule(r){3-6}
				& {5-shot} &
				57.70$\pm$0.59\%&
				56.73$\pm$0.64\%&
				56.03$\pm$0.64\%&
				\textbf{59.31$\pm$0.62\%}\\
				
				\arrayrulecolor{black!100}\midrule
				
				\multirow{2}{*}{\textbf{2Mode}} &
				{1-shot} &
				69.55$\pm$0.54\%&
				70.17$\pm$0.61\%&
				70.60$\pm$0.56\%&
				\textbf{73.69$\pm$0.52\%}\\
				
				\arrayrulecolor{black!30}\cmidrule(r){3-6}
				& {5-shot} &
				75.12$\pm$0.41\%&
				75.33$\pm$0.46\%&
				75.72$\pm$0.47\%&
				\textbf{79.82$\pm$0.40\%}\\
				\arrayrulecolor{black!100}\midrule
				
				\multirow{2}{*}{\textbf{3 Mode}} &
				{1-shot} &
				58.14$\pm$0.49\%&
				59.89$\pm$0.50\%&
				59.62$\pm$0.54\%&
				\textbf{62.08$\pm$0.54\%}\\
				
				\arrayrulecolor{black!30}\cmidrule(r){3-6}
				& {5-shot} &
				66.84$\pm$0.44\%&
				67.03$\pm$0.44\%&
				67.51$\pm$0.47\%& 
				\textbf{70.03$\pm$0.43\%}\\
				\arrayrulecolor{black!100}\midrule
				
				\multirow{2}{*}{\textbf{5Mode}} &
				{1-shot} &
				49.31$\pm$0.53\%&
				50.69$\pm$0.57\%&
				51.75$\pm$0.52\%&
				\textbf{56.72$\pm$0.46\%}\\
				
				\arrayrulecolor{black!30}\cmidrule(r){3-6}
				& {5-shot} &
				58.91$\pm$0.51\%&
				59.88$\pm$0.54\%&
				59.95$\pm$0.42\%&
				\textbf{64.91$\pm$0.38\%}\\
				\arrayrulecolor{black!100}
				\bottomrule
				
							\end{tabular}
		}
	\end{center}
\end{threeparttable}
\end{table*}

{\bf Multimodal Few-shot Classification Results.}
For multimodal few-shot classification, similar to ~\cite{vuorio2019multimodal},  we train and evaluate models on the meta-datasets with two modes (Omniglot and mini-Imagenet), three modes (Omniglot, mini-Imagenet, and FC100), and five modes (all the five datasets). This meta-dataset includes a few-shot classification of characters, natural objects with different statistics and also fine-grained classification of different bird and aircraft types. Due to the large discrepancies between datasets, the possibility of negative transfer increases which makes few-shot classification more challenging. We also combine the mini-ImageNet and FC100 datasets to construct another two-mode meta-dataset. Both mini-ImageNet and FC100 include samples of natural objects and few-shot tasks generated from them probably have a similar underlying structure. However, the few-shot classification of FC-100 tasks is a bit challenging due to the smaller image size. So, here we aim to investigate the possibility of knowledge transfer from mini-ImageNet to FC100. We use two different meta-learners namely ProtoNet ~\cite{snell2017prototypical}, and MAML ~\cite{finn2017model}, and several baselines considering each meta-learner. The first baseline is training and testing multiple separate meta-learners, one for each dataset. The second baseline is considered when meta-learner has no access to task mode information and simply trained and tested in the combination of tasks from different modes. The third baseline is the MMAML ~\cite{vuorio2019multimodal} which to the best of our knowledge is the only proposed algorithm for multimodal few-shot learning. A variant of this baseline (MProtoNet) is also used here by replacing the ProtoNet meta-learner with MAML in the general framework of ~\cite{vuorio2019multimodal}.

Multimodal few-shot image classification results are shown in Table \ref{table:Multimodal_Classification}. Comparing MProtoNet with MMAML, we can see that replacing ProtoNet with MAML considerably improves the classification accuracy. We believe that most of the extra accuracy gained in the multimodal framework by replacing MAML with ProtoNet results from the proper training of modulation network due to improved gradient flow from meta-learner. A detailed experiment can be found in supplementary. Additionally, as the results in Table \ref{table:Multimodal_Classification} suggest, the proposed KML scheme considerably improves the performance of multimodal few-shot classification with both meta-learners. The improvement in \emph{5-shot} scenarios is higher because more clues about the task provide more accurate embedding vectors to generate the modulation parameters. We have also included the detailed accuracy for 2Mode\textsuperscript{\dag} scenario consisting of mini-ImageNet and FC100 in table \ref{table:Multimodal_Classification_2}. This shows on average a positive knowledge transfer from mini-ImageNet to FC100 dataset in terms of increased classification accuracy compared to a single meta-learner trained and tested on FC100. The improved accuracy of proposed KML over previous modulation comes at the cost of parameter and computational overhead. A detailed discussion can be found in the supplementary. In addition to multimodal scenario, KML also brings considerable improvement in the conventional unimodal scenario (e.g., mini-ImageNet). Please see supplementary for more details and results.

\begin{table*}
\caption{Meta-test accuracies including the performance on each dataset.}
	\label{table:Multimodal_Classification_2}
\begin{threeparttable}
	\begin{center}
		\resizebox{\linewidth}{!}{
			\begin{tabular}{lcccccc}
				\toprule
				
				\multirow{3}{*}{\textbf{Method}} & \multicolumn{6}{c}{\textbf{Datasets}} \\
				\cmidrule(r){2-7}
				& \multicolumn{2}{c}{\textbf{mini-ImageNet}} & \multicolumn{2}{c}{\textbf{FC100}} & \multicolumn{2}{c}{\textbf{Overall}}\\
				\cmidrule(r){2-3}
				\cmidrule(r){4-5}
				\cmidrule(r){6-7}
				 & 1shot & 5shot & 1shot & 5shot & 1shot & 5shot  \\
				\midrule
				
				\textbf{MAML1} &
				41.70$\pm$0.91\%&
				\textbf{60.03$\pm$0.88\%}&
				---&---&
				\multirow{2}{*}{39.27$\pm$0.86\%}& \multirow{2}{*}{{53.51$\pm$0.82\%}}\\
				
				\textbf{MAML2} &---&---&
				36.84$\pm$0.81\%&
				47.08$\pm$0.74\%&&\\
				\arrayrulecolor{black!30}\midrule
				
				\textbf{MAML~\cite{finn2017model}}& 
				\textbf{44.11$\pm$0.88\%} & 
				59.34$\pm$0.78\% &
				37.01$\pm$0.82\% &
				\textbf{49.01$\pm$0.73\%} &
				40.53$\pm$0.68\% & 
				\textbf{54.11$\pm$0.63\%}\\
				\midrule
				
				\textbf{MMAML~\cite{vuorio2019multimodal}}& 
				42.70$\pm$0.86\% & 
				56.93$\pm$0.78\% & 
				35.50$\pm$0.75\% & 
				47.11$\pm$0.77\% & 
				39.11$\pm$0.72\% & 
				52.02$\pm$0.63\% \\
				\midrule
				\textbf{MAML+KML(ours)} &
				{44.01$\pm$0.82\%} & 
				{58.95$\pm$0.76\%} & 
				\textbf{37.47$\pm$0.86\%} & 
				{48.45$\pm$0.76\%} & 
				\textbf{40.73$\pm$0.66\%} & 
				{53.72$\pm$0.60\%}\\

				\arrayrulecolor{black!100}\midrule
				
				\textbf{ProtoNet1} &
				47.19$\pm$0.58\%&
				61.11$\pm$0.53\%&
				---&---&
				\multirow{2}{*}{43.42$\pm$0.56\%}& \multirow{2}{*}{56.73$\pm$0.64\%}\\
				
				\textbf{ProtoNet2} &---&---&
				39.61$\pm$0.54\%&
				52.35$\pm$0.57\%&&\\
				\arrayrulecolor{black!30}\midrule
				
				\textbf{ProtoNet~\cite{snell2017prototypical}}& 
				45.62$\pm$0.82\% & 
				\textbf{61.74$\pm$0.75\%} &
				40.36$\pm$0.76\%&
				53.66$\pm$0.77\%&
				43.05$\pm$0.58\%& 
				57.70$\pm$0.59\%\\
				\midrule
				
				\textbf{MProtoNet~\cite{vuorio2019multimodal}}& 
				47.06$\pm$ 0.80 & 
				60.56$\pm$ 0.77 & 
				40.09$\pm$ 0.77& 
				51.50$\pm$ 0.83 & 
				43.57$\pm$ 0.59 & 
				56.03$\pm$ 0.64 \\
				\midrule
				\textbf{MProto+KML(ours)} &
				\textbf{48.34$\pm$ 0.86\%} & 
				{61.20$\pm$ 0.77\%}& 
				\textbf{40.48$\pm$ 0.77\%}& 
				\textbf{54.42$\pm$ 0.76\%}& 
				\textbf{44.40$\pm$ 0.65\%}& 
				\textbf{59.31$\pm$ 0.62\%}\\
				\arrayrulecolor{black!100}\bottomrule
			\end{tabular}
		}
	\end{center}

\end{threeparttable}
\end{table*}

\begin{figure*}
\centering
\begin{subfigure}{.45\textwidth}
  \centering
  \includegraphics[clip, trim={0.5cm 0.1cm 0.5cm 0.5cm}, width=6cm]{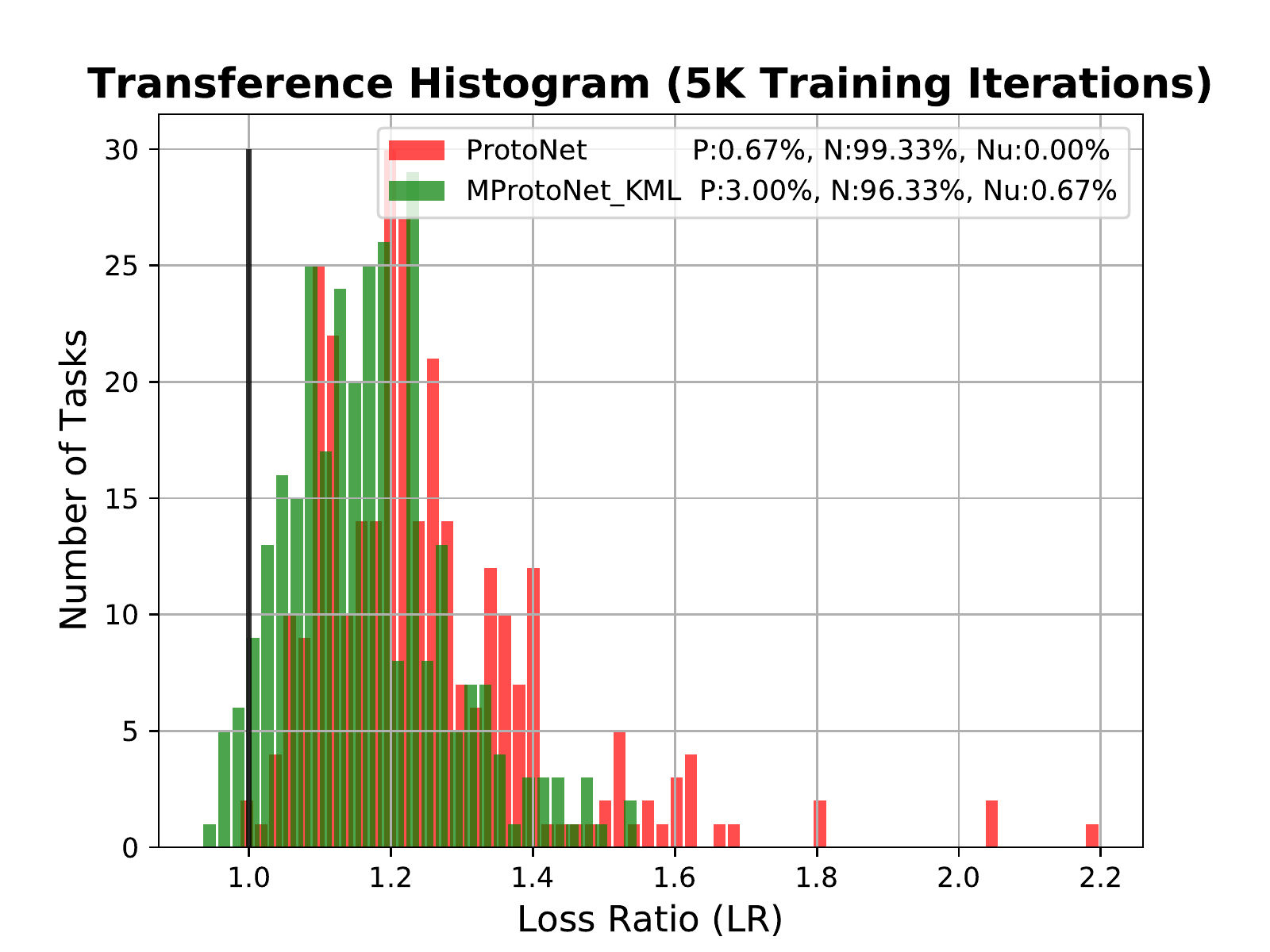}
  \caption{}
  \label{fig:sfig1}
\end{subfigure}%
\begin{subfigure}{.45\textwidth}
  \centering
  \includegraphics[clip, trim={0.5cm 0.1cm 0.5cm 0.5cm}, width=6cm]{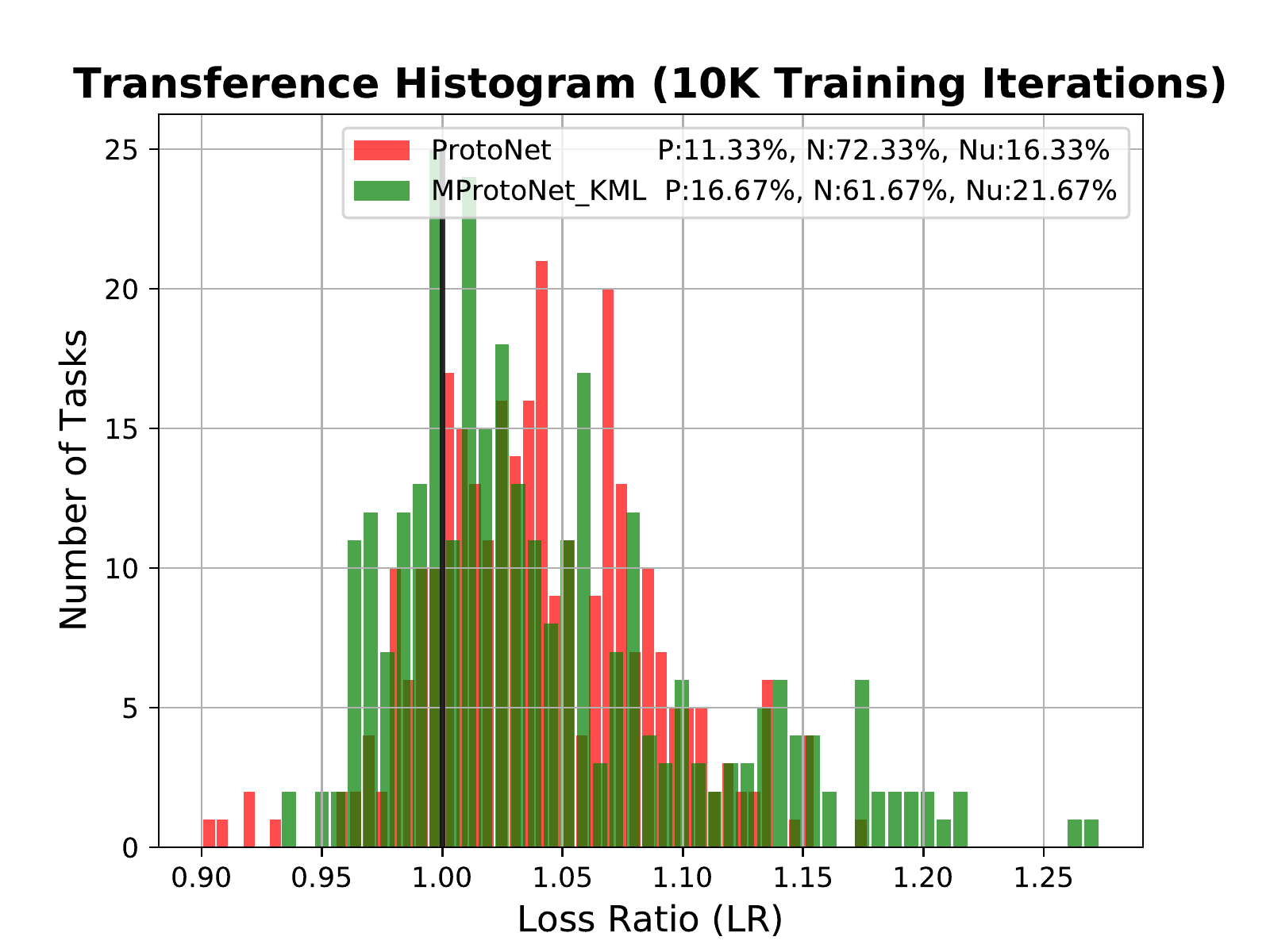}
  \caption{}
  \label{fig:sfig2}
\end{subfigure}
\begin{subfigure}{.45\textwidth}
  \centering
  \includegraphics[clip, trim={0.5cm 0.1cm 0.5cm 0.5cm}, width=6cm]{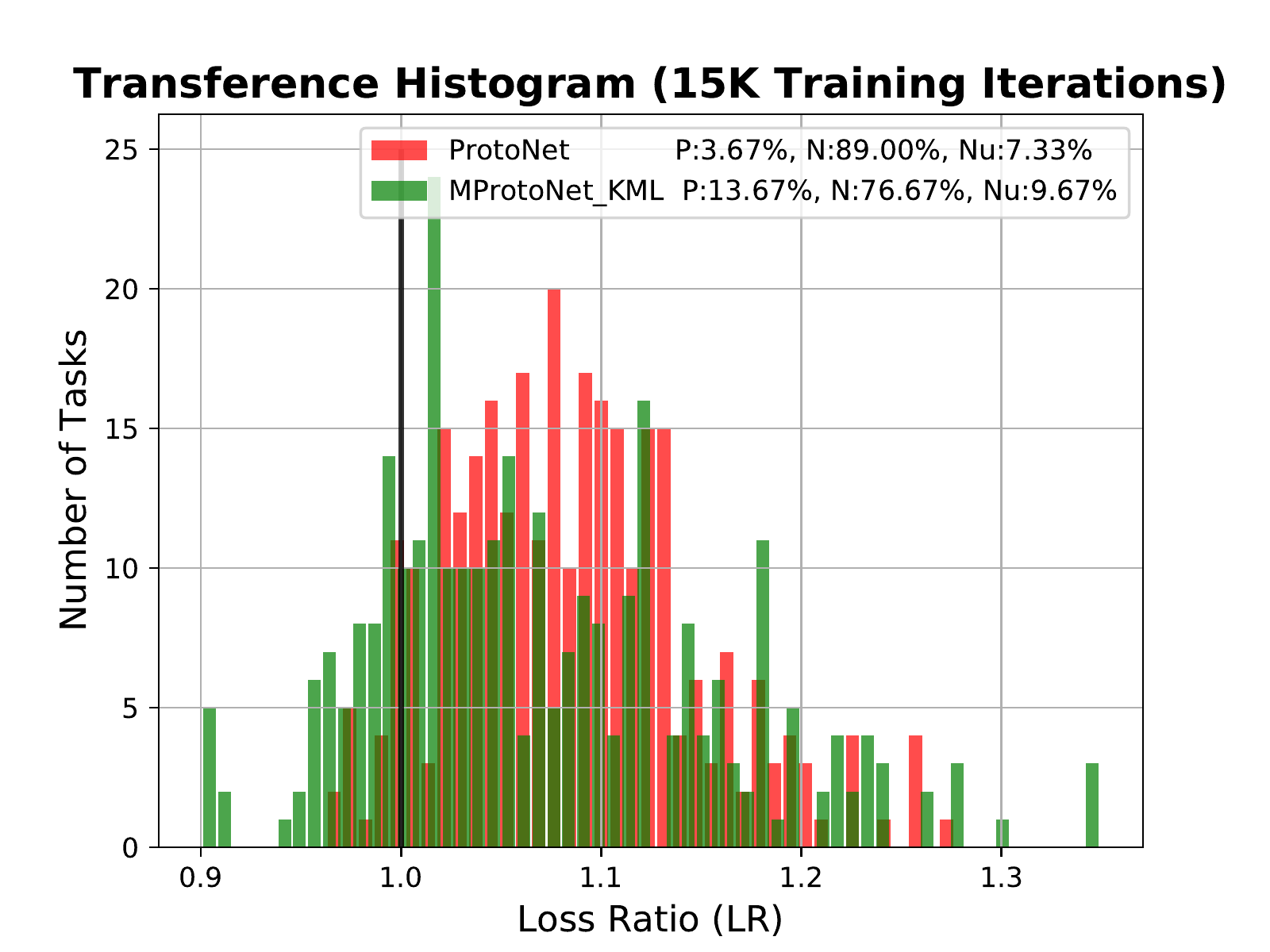}
  \caption{}
  \label{fig:sfig3}
\end{subfigure}%
\begin{subfigure}{.45\textwidth}
  \centering
  \includegraphics[clip, trim={0.5cm 0.1cm 0.5cm 0.5cm}, width=6cm]{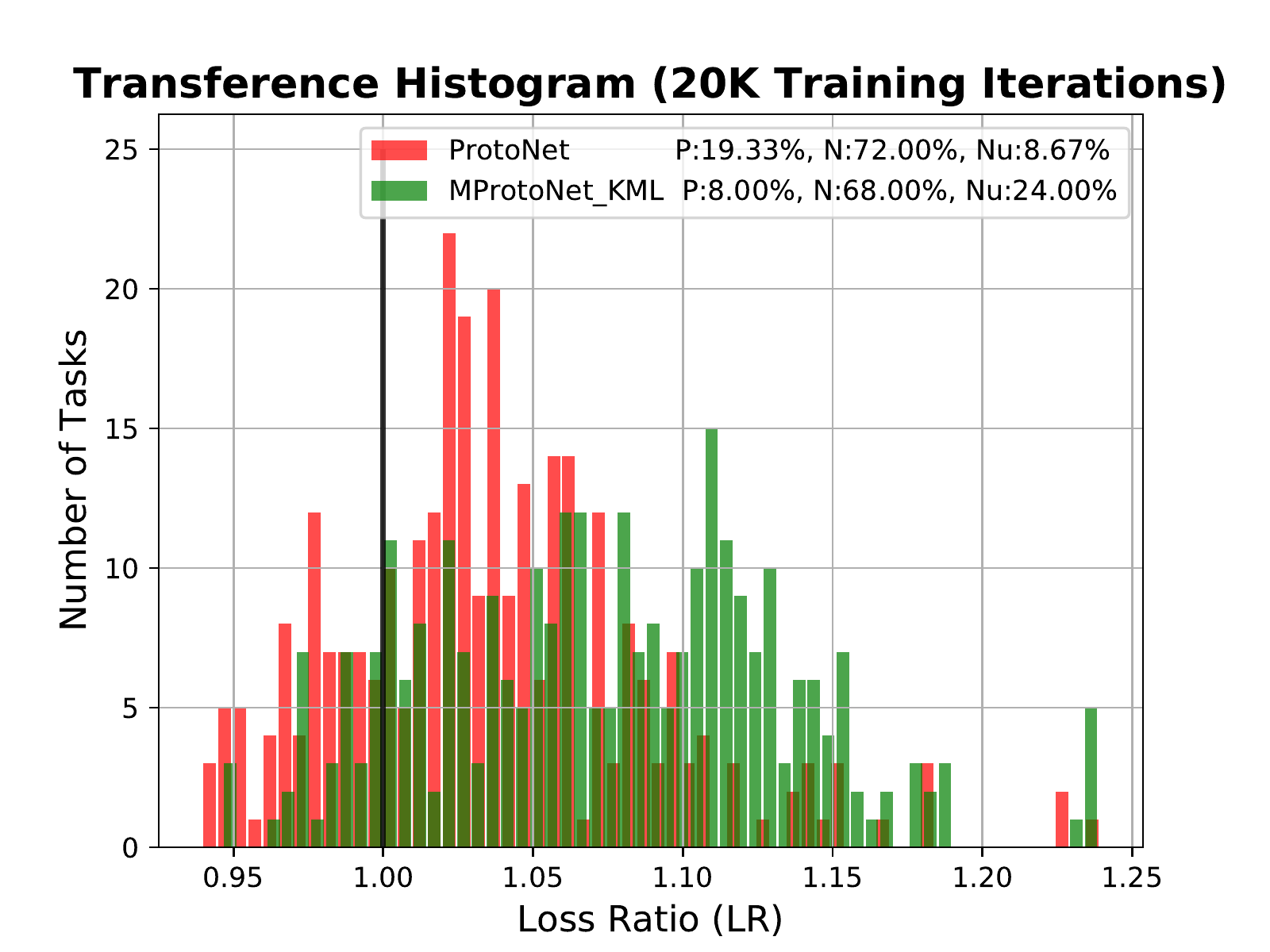}
  \caption{}
  \label{fig:sfig4}
\end{subfigure}

\caption{Transference Histogram from mini-ImageNet to FC100 task.}
\label{fig:transference_MC1}
\end{figure*}

{\bf Transference Results.} In addition to classification results, here we analyse the performance of the proposed KML method on improving knowledge transfer. To this end, we apply the proposed transference metric to analyse the knowledge transfer from 300 randomly sampled mini-ImageNet meta-train tasks to another random meta-test FC100 task. We use Algorithm \ref{alg:transference} to calculate the transference during training and the transference histogram is plotted in figure \ref{fig:transference_MC1} for both ProtoNet and proposed MProtoNet+KML. In each sub-figure, we have included the results for both methods for ease of comparison. A vertical line is also drawn to specify the threshold between positive and negative transference. The percentage of the Positive (P), Negative (N) and Neutral (Nu) transference is also shown for each method. For classifying a source task as neutral, we use an interval around the threshold that the loss change is negligible and corresponds to less than 0.05\% change in the meta-test accuracy. Note that in average, proposed KML also performs better in micro-level in terms of increasing positive knowledge transfer and reducing the negative transfer.

The cross mode knowledge transfer can be investigated between different datasets. An interesting behavior that  we observed during our analysis is that even Omniglot few-shot tasks --which are considered as easy tasks and can be handled very well with current meta-learning algorithms--
can have positive knowledge transfer to challenging FC100 tasks specially in later training iterations. We believe in this case Omniglot tasks act more like a regularizer as they emphasize on simple yet strong features and prevent the model from overfitting to FC100 meta-train task (please find the analysis results in supplementary). Note that knowledge transfer can also be analysed within tasks of a single dataset as in some datasets like mini-ImageNet classes vary so much in few-shot setup (please see supplementary for more details).

\section{Discussion and Conclusion}
In this work, we propose a new quantification method for understanding knowledge transfer in multimodal meta-learning. We then propose a new interpretation of the modulation mechanism in MMAML. These lead us to propose a new multi-modal meta-learner inspired by hard parameter sharing in multi-task learning. Our extensive experimental results show that the proposed method achieves substantial improvement over the best results in multimodal meta-learning. While our major focus is on multimodal meta-learning, our work also attempts to shed light on conventional unimodal meta-learning. Limitations are discussed in the supplementary.

\section*{Acknowledgement}

This project was supported by SUTD project
PIE-SGP-AI-2018-01. This research was also supported by the National Research Foundation Singapore under its AI Singapore Programme [Award Number: AISG-100E2018-005]. The authors are grateful for the fruitful discussion with Yuval Elovici, and Alexander Binder.

{\small
\bibliographystyle{unsrtnat}
\bibliography{egbib.bib}
}
\newpage

\appendix

\section{Additional Few-Shot Classification Results}
\label{sec:few_shot_experiemtns}
In this section we include additional experimental results for both unimodal and multimodal few-shot classification.
Note that here we focus on the analysis of our proposed meta-learner considering the  hard parameter sharing concept in MTL. We recall that, by hard parameter sharing for a layer, similar to MTL, we mean the layer that is not modulated using proposed KML and all parameters are shared between different few-shot tasks. For example, in the case of "{\bf 1\textsuperscript{st} Layer Shared}", we mean we do not apply the modulation on the parameters of the first layer (lines 5,6,7,11 and 12 are bypassed for these parameters in Algorithm 2). While for the remaining layers, the modulated parameters are generated and applied in the inner-loop, and then the modulation network is updated in the outer-loop following the procedure in Algorithm 2. So, when all of the layers are shared, the algorithm reduces to the vanilla meta-learner.

{\bf Unimodal Few-Shot Classification.} When compared to a multimodal scenario, in conventional unimodal few-shot setup, there could be less negative knowledge transfer between few-shot tasks (see Sec. \ref{sec:additional_transference_analysis} for an example). In this case, from the experiences in the MTL domain, we expect the performance to be increased when some layers are shared between tasks (specially earlier layers which encode low-level features ~\cite{zeiler2014visualizing}). The meta-test accuracies for different number of shared layers in a 4-layer CNN are shown in table \ref{table:Unimodal_Classification_miniimagenet2}. In this case, no shared layers means that all of the layers are modulated using task-aware KML scheme. In contrast, all layers shared means that no modulation is applied. Based on these results, proposed KML improves the unimodal few-shot classification by up to 2.5\% compared to the vanilla meta-learner. Additionally,
modulating only third and fourth layers of the CNN on average yields the best results. The details of network architecture and hyperparameters used for this experiment can be found in Sec. \ref{sec:experimental_details}.

\begin{table}[h]
\caption{Meta-test accuracies on unimodal few-shot classification for different number of shared layers.}
\label{table:Unimodal_Classification_miniimagenet2}
\begin{threeparttable}
\begin{center}
    \resizebox{0.90\linewidth}{!}{
			\begin{tabular}{lcccc}
				\toprule
				\multirow{2}{*}{\textbf{Shared Layers}} & \multicolumn{2}{c}{\textbf{mini-ImageNet}} &
				\multicolumn{2}{c}{\textbf{tiered-ImageNet}}\\
				\cmidrule(l){2-3}
				\cmidrule(r){4-5}
                 & 1-shot & 5-shot & 1-shot & 5-shot\\
				\midrule
				
				\textbf{No Shared Layers}&
				53.18$\pm$0.51\%&
				67.18$\pm$0.39\%&
				54.36$\pm$ 0.39\% &
				71.84$\pm$ 0.27\% \\
				\arrayrulecolor{black!30}\midrule
				
				\textbf{1\textsuperscript{st} Layer}&
				53.54$\pm$ 0.66\% &
				\textbf{68.07$\pm$0.45\%}&
				54.22$\pm$ 0.35\%&
				71.93$\pm$ 0.28\% \\
				\arrayrulecolor{black!30}\midrule

				\textbf{1\textsuperscript{st} \& 2\textsuperscript{nd} Layers} &
				\textbf{54.10$\pm$ 0.61\%} &
				{67.31$\pm$ 0.35\%}&
				\textbf{54.67$\pm$ 0.39\%} &
				\textbf{72.09$\pm$ 0.27\%}\\
				\arrayrulecolor{black!30}\midrule
				
				\textbf{1\textsuperscript{st}, 2\textsuperscript{nd} \& 3\textsuperscript{rd} Layers}&
				{52.83$\pm$ 0.57\%} &
				{66.98$\pm$ 0.44\%}&
				54.10$\pm$ 0.37\% &
				71.68$\pm$ 0.29\%\\
				\arrayrulecolor{black!30}\midrule
				
				\textbf{All Layers (ProtoNet)} &
				51.55$\pm$0.51\% &
				65.83$\pm$0.36 \%&
				53.01$\pm$0.33\% &
				70.11$\pm$0.29\%\\
				
				\arrayrulecolor{black!100}\bottomrule
			\end{tabular}
		}
\end{center}
\end{threeparttable}
\end{table}


{\bf Multimodal Few-Shot Classification.} In contrast, in multimodal scenario, few-shot tasks are more diverse. So, the negative transfer can happen more often due to the larger discrepancy between the characteristics of the different datasets. Therefore, here we expect the performance to degrade by sharing layers between tasks from different modes. The experimental results for sharing different number of layers in a 4-layer CNN are shown in table \ref{table:multimodal_Classification_2mode2} for 2Mode multimodal classification (including mini-ImageNet and FC100). Similar results are obtained for other modes. As the results suggest, sharing layers between a diverse set of tasks degrades the performance, and the best results obtained when all of the layers are modulated. 


\begin{table}[h]
\caption{Meta-test accuracies on multimodal few-shot classification by including hard parameter sharing.}
\label{table:multimodal_Classification_2mode2}
\begin{threeparttable}
\begin{center}
    \resizebox{0.55\linewidth}{!}{
			\begin{tabular}{lcc}
				\toprule
				\multirow{2}{*}{\textbf{Shared Layers}} & \multicolumn{2}{c}{\textbf{2Mode\textsuperscript{$\dag$}}}\\
				
				\cmidrule(l){2-3}
                 
                 & 1-shot & 5-shot\\
				\midrule
				
				\textbf{No Shared Layers}&
				{\bf 44.40$\pm$0.65\%}&
				{\bf 59.31$\pm$0.62\%}\\
				\arrayrulecolor{black!30}\midrule
				
				\textbf{1\textsuperscript{st} Layer}&
				44.18$\pm$ 0.64\% &
				59.07$\pm$0.60\%\\
				\arrayrulecolor{black!30}\midrule

				\textbf{1\textsuperscript{st} \& 2\textsuperscript{nd} Layers} &
				43.77$\pm$ 0.65\%&
				58.65$\pm$ 0.59\%\\
				\arrayrulecolor{black!30}\midrule
				
				\textbf{1\textsuperscript{st}, 2\textsuperscript{nd} \& 3\textsuperscript{rd} Layers}&
				43.59$\pm$ 0.60\%&
				58.40$\pm$ 0.61\%\\
				\arrayrulecolor{black!30}\midrule
				
				\textbf{All Layers (ProtoNet)} &
				43.05$\pm$0.58\% &
				57.70$\pm$0.59 \%\\
				
				\arrayrulecolor{black!100}\bottomrule
			\end{tabular}
		}
\end{center}
\end{threeparttable}
\end{table}

Considering this behavior, for multimodal few-shot classification, the optimal layer sharing configuration is to modulate all layers and use no shared layers. Also for unimodal scenario, sharing first two layers produces the best results due to less negative transfer in unimodal scenario. These results justify the configurations used in the main paper.

\section{Additional Experimental Results on Transference Analysis}
\label{sec:additional_transference_analysis}

Here we provide some additional results on transference analysis that could not be included in the main text due to the lack of space. First, we provide some additional results on the transference from mini-ImageNet meta-train tasks to FC100 target tasks. Then, the results for transference analysis from Omniglot to FC100 is discussed. Finally, we provide some results for transference between tasks within a conventional unimodal few-shot learning.

{\bf Network Structure.} In the transference analysis experiments using ProtoNet and MProtoNet+KML networks, we exactly use the same structure and hyperparameters discussed in Sec. \ref{sec:experimental_details} of this supplementary. 

\begin{figure}[h]
\centering
\begin{subfigure}{.5\textwidth}
  \centering
  \includegraphics[clip, trim={0.5cm 0.1cm 0.5cm 0.5cm}, width=7cm]{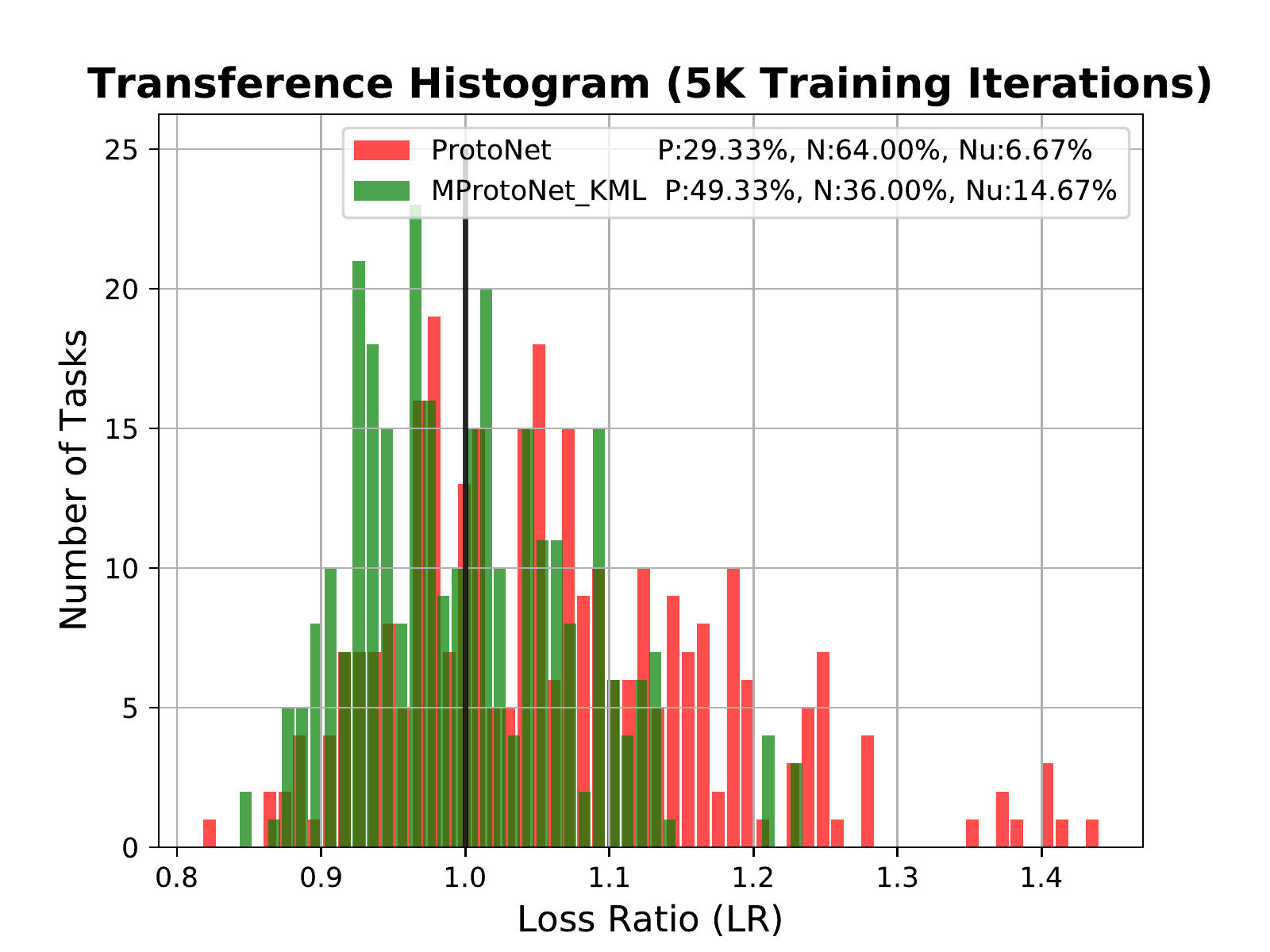}
  \caption{}
\end{subfigure}%
\begin{subfigure}{.5\textwidth}
  \centering
  \includegraphics[clip, trim={0.5cm 0.1cm 0.5cm 0.5cm}, width=7cm]{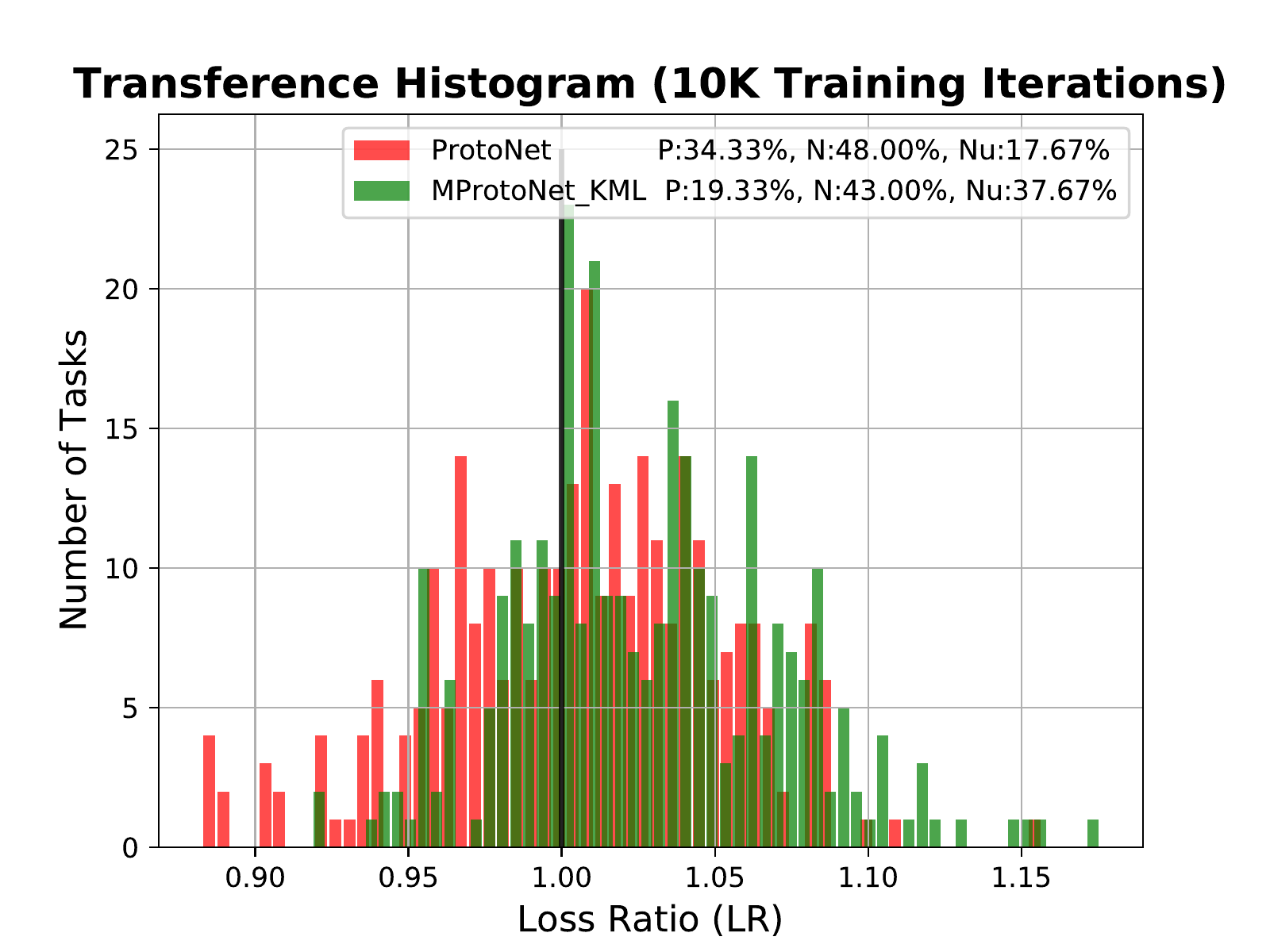}
  \caption{}
\end{subfigure}
\begin{subfigure}{.5\textwidth}
  \centering
  \includegraphics[clip, trim={0.5cm 0.1cm 0.5cm 0.5cm}, width=7cm]{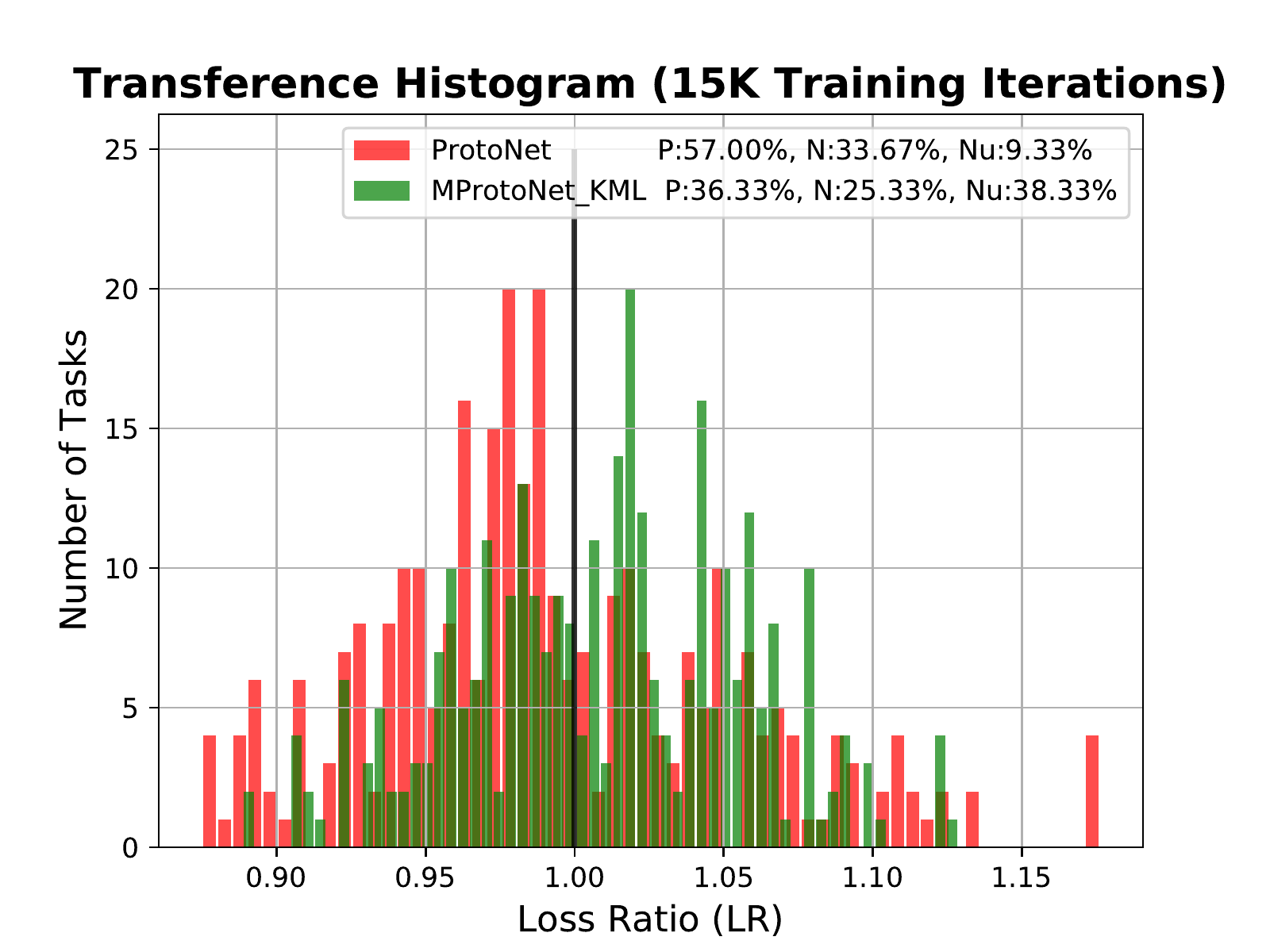}
  \caption{}
\end{subfigure}%
\begin{subfigure}{.5\textwidth}
  \centering
  \includegraphics[clip, trim={0.5cm 0.1cm 0.5cm 0.5cm}, width=7cm]{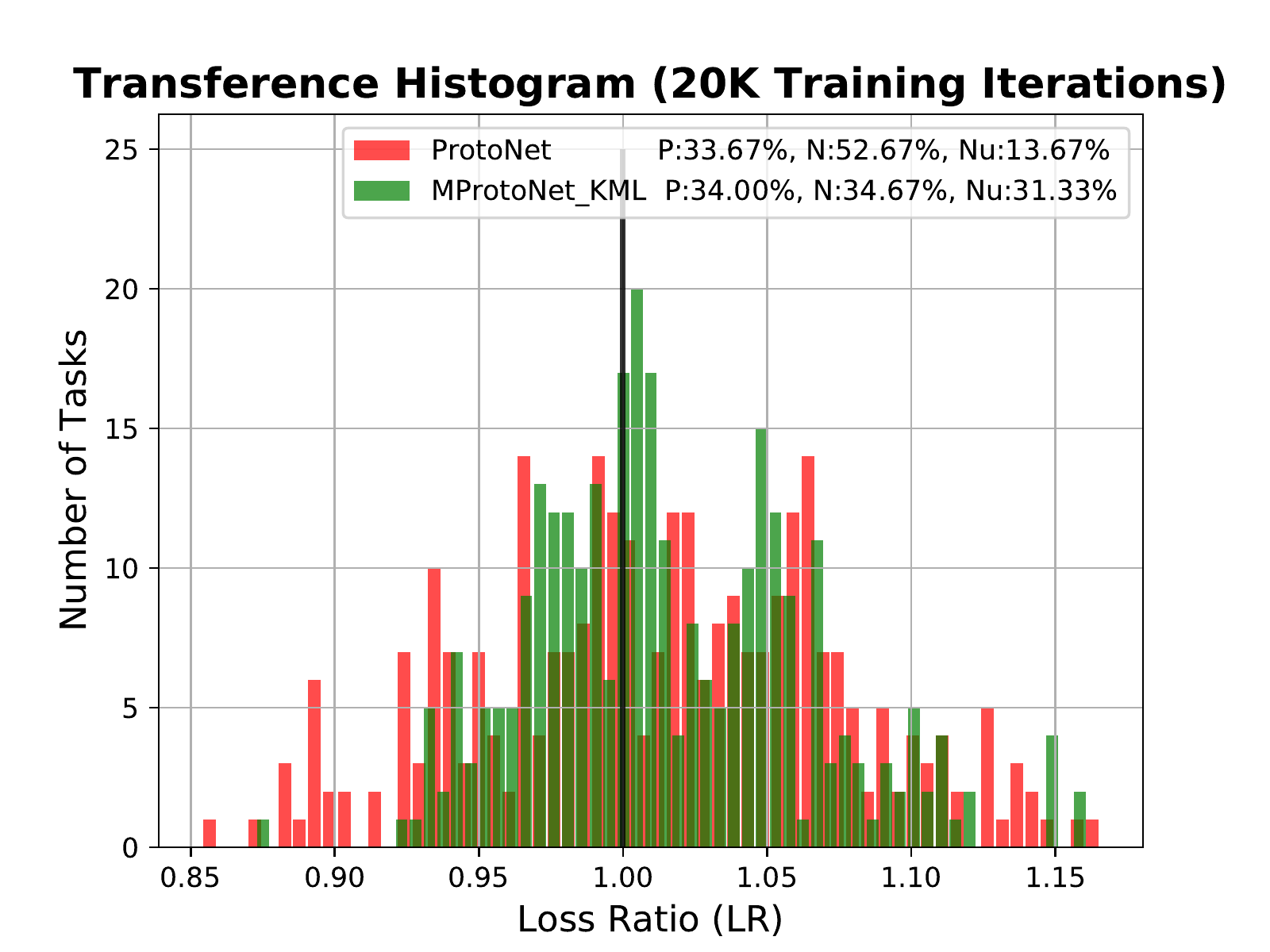}
  \caption{}
\end{subfigure}
\begin{subfigure}{.5\textwidth}
  \centering
  \includegraphics[clip, trim={0.5cm 0.1cm 0.5cm 0.5cm}, width=7cm]{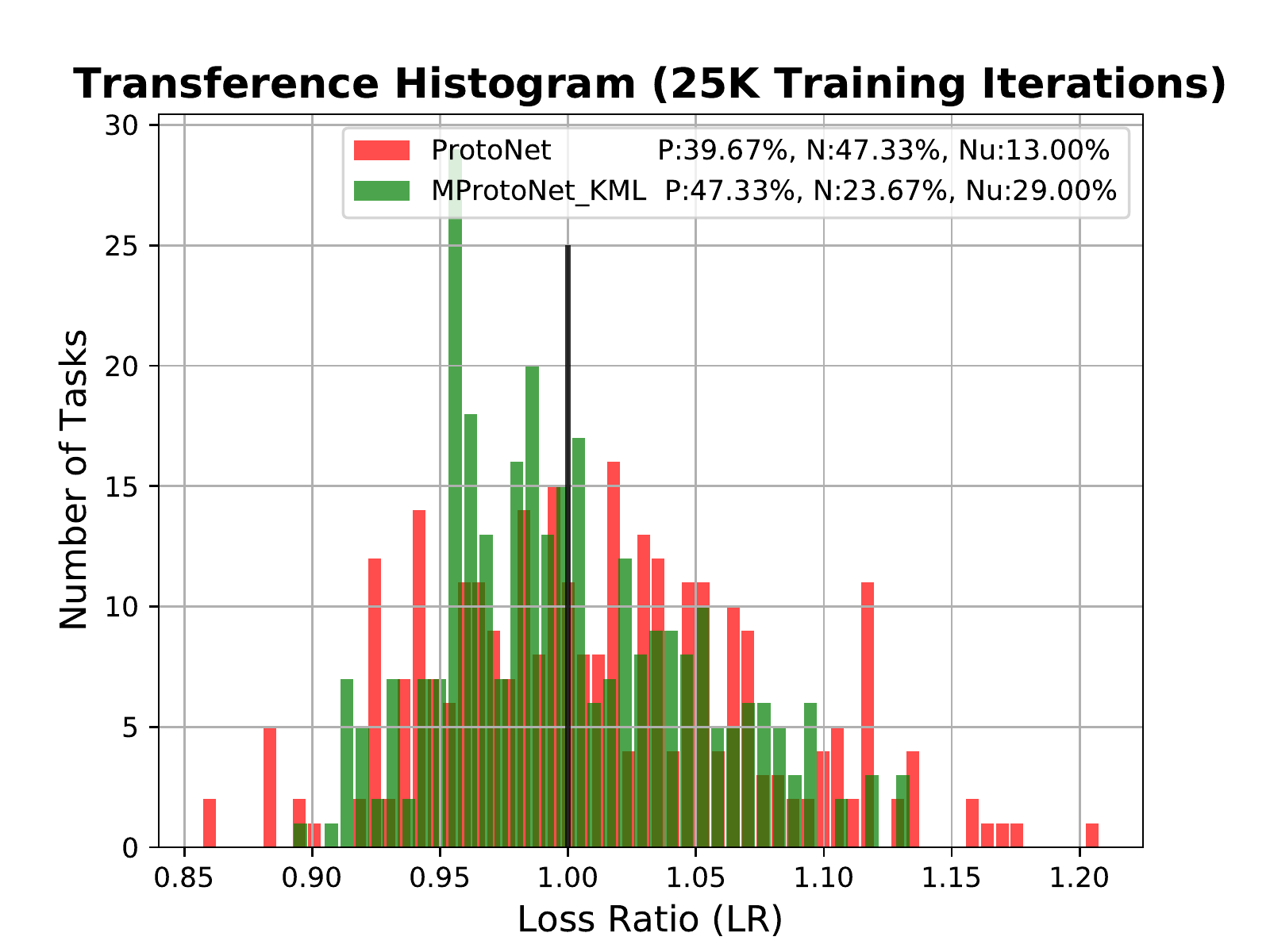}
  \caption{}
\end{subfigure}%
\begin{subfigure}{.5\textwidth}
  \centering
  \includegraphics[clip, trim={0.5cm 0.1cm 0.5cm 0.5cm}, width=7cm]{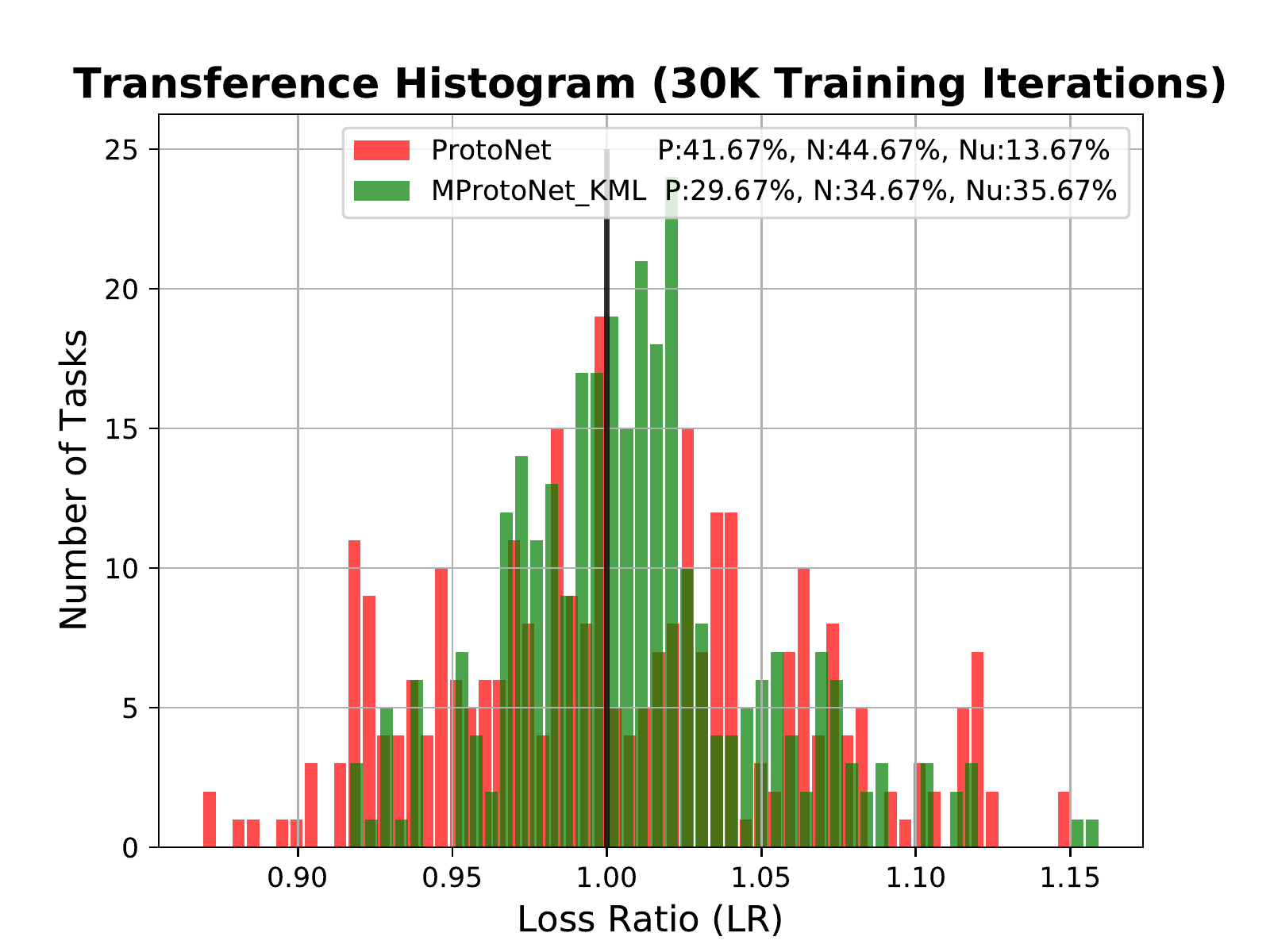}
  \caption{}
\end{subfigure}
\caption{Transference Histogram from mini-ImageNet to FC100 task.}
\label{fig:transference_MC2}
\end{figure}

\subsection{Cross-Mode Transference in Multi-Modal Few-Shot Learning}

{\bf Transference from miniImageNet to FC100.}
Here, we repeat the transference analysis from 300 randomly sampled mini-ImageNet meta-train source tasks into a randomly sampled target tasks from FC100. We train the model with multimodal dataset (including both mini-ImageNet and FC100 tasks). For extracting the transference histogram in an specific epoch, we use the network parameters in this epoch as initial parameters for transference analysis. Then using Algorithm 1 (main paper), first we calculate the loss on target meta-test task using initial parameters. Then, for each source task, we use the data from that task to calculate the adapted parameters to that task and then calculate the loss of target task on adapted parameters. Then, transference from each task is the ratio between the loss of target task after and before adapting.

This target classification task contains samples from “otter, girl, dolphin, raccoon, and skunk”. The classes are disjoint enough and the sampled data points are clean which makes classification task a less challenging one to handle. On the other hand, most of the classes share similar underlying structure as in the animal subcategories of the mini-ImageNet tasks. So, as we expect, the ProtoNet also performs better in meta-target task with the information provided by the source mini-ImageNet tasks. The transference histograms during training are shown in figure \ref{fig:transference_MC2}.
Based on the transference results, following points can be considered:

\begin{itemize}
    
    \item In the beginning iterations of the training, most of the source tasks from mini-ImageNet have negative transference on the target task. The probable reason could be that at the beginning stages, the model has not learned to generalize and still tries to remember the meta-training tasks and corresponding samples (Please note that the LR is measured on a meta-test task which consists of unseen classes under the few-shot problem setting).

    \item As training proceeds, the percentage of the positive transference increases. After learning the useful features, the network begins to overfit to training tasks and as a result its generalization performance (knowledge transfer to FC100 meta-test task) degrades.
    
\end{itemize}

Overall, our additional results are consistent with those in the main paper, and support the observation: {\bf In the case of cross mode knowledge transfer, negative transference occurs at the beginning iterations and increasingly more positive transference occurs as training proceeds.}
As the major advantage of cross mode knowledge transfer is to improve generalization to better handle  unseen test tasks, it is reasonable to observe positive transference in the later iterations when the networks learn features for generalization.

\begin{figure*}
 \begin{center}
 \includegraphics[clip, trim={2.1cm 2.5cm 2.2cm 3cm}, width=14.5cm]{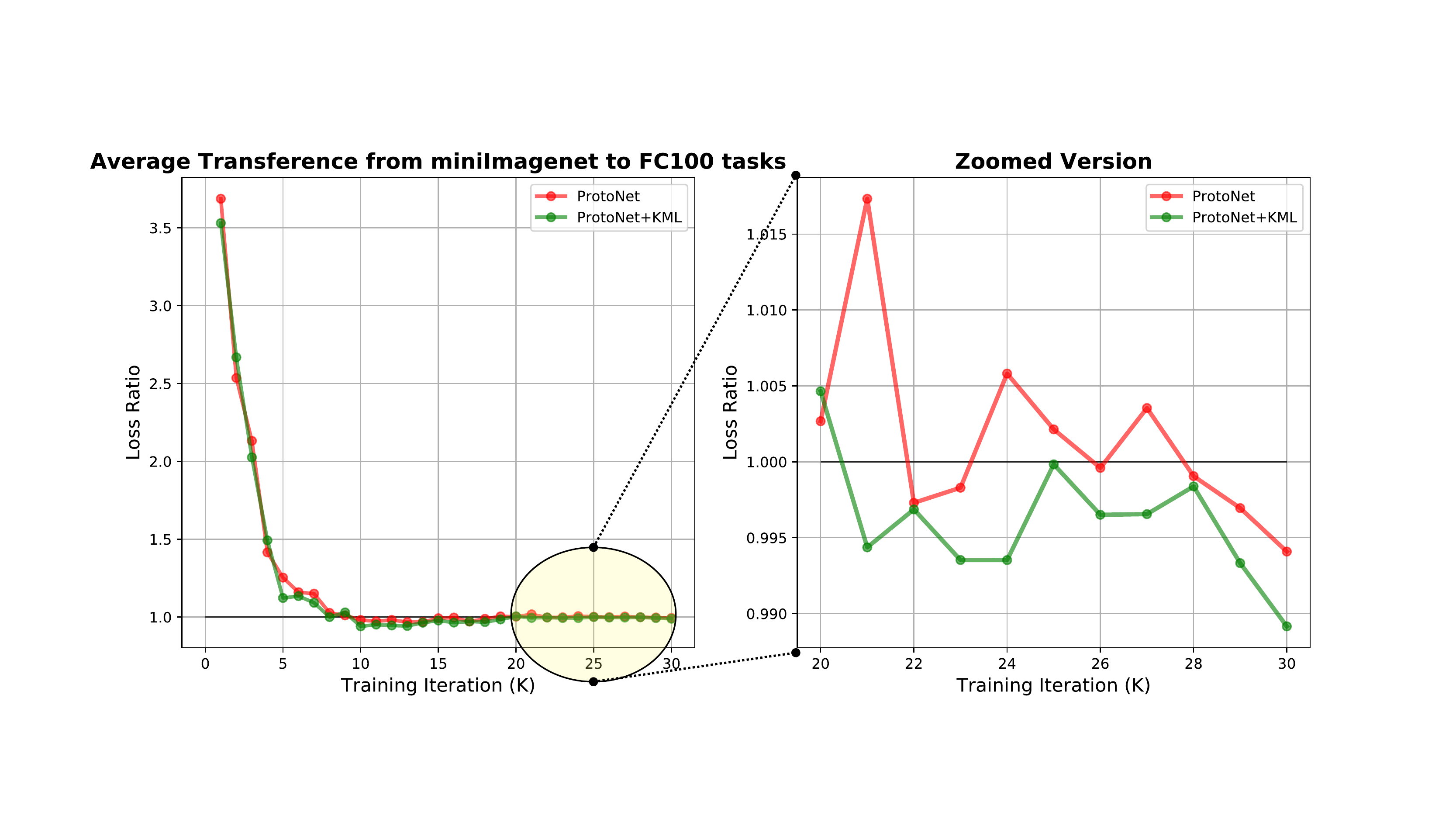}
 \end{center}
   \caption{Average transference from 300 mini-ImageNet task to 100 meta-test FC100 tasks.}
\label{fig:avergae_transference_MC}
\end{figure*}

{\bf Average Transference.} The average transference values from 300 meta-train mini-ImageNet tasks to 100 meta-test FC100 tasks are shown in figure \ref{fig:avergae_transference_MC}. For calculating each point in this figure, we have averaged the transference results from 300 source tasks to each task and then computed the average along all target tasks in that training iteration. This figure shows that in the beginning of the training, the information provided by the mini-ImageNet are almost negative for generalization performance on the FC100 meta-test tasks. As training proceeds and the network learns necessary  features (and probably low and mid-level ones), it can make good use of additional information provided by the external mini-ImageNet dataset. This means that {\em cross-mode knowledge transfer occurs on the later iterations of training}. Results also show that on average the proposed method performs better in terms of obtaining the information provided by the source meta-train tasks.

\begin{figure}[h]
\centering
\begin{subfigure}{.5\textwidth}
  \centering
  \includegraphics[clip, trim={0.3cm 0.1cm 0.1cm 0.5cm}, width=7cm]{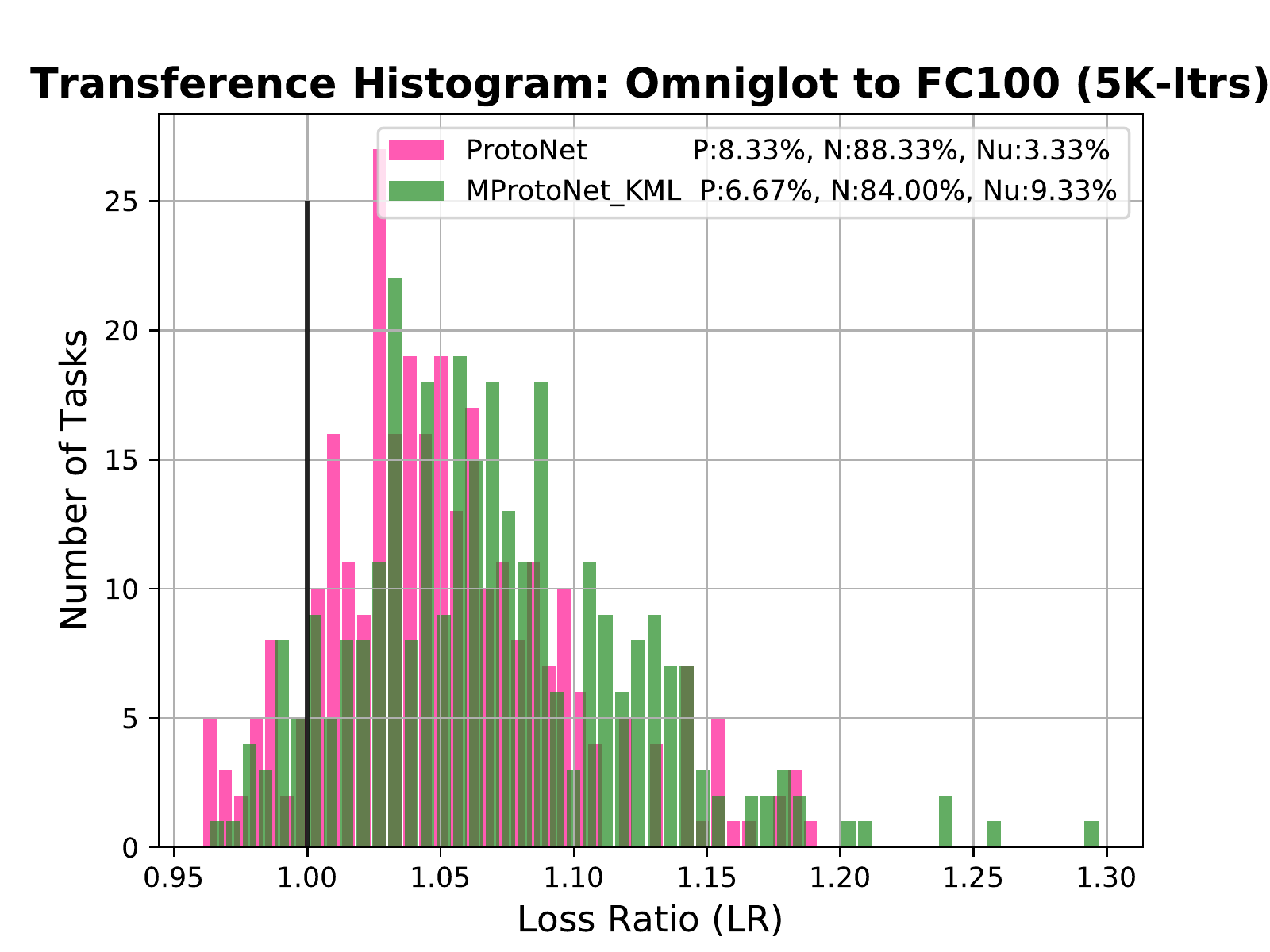}
  \caption{}
\end{subfigure}%
\begin{subfigure}{.5\textwidth}
  \centering
  \includegraphics[clip, trim={0.5cm 0.1cm 0.4cm 0.5cm}, width=7cm]{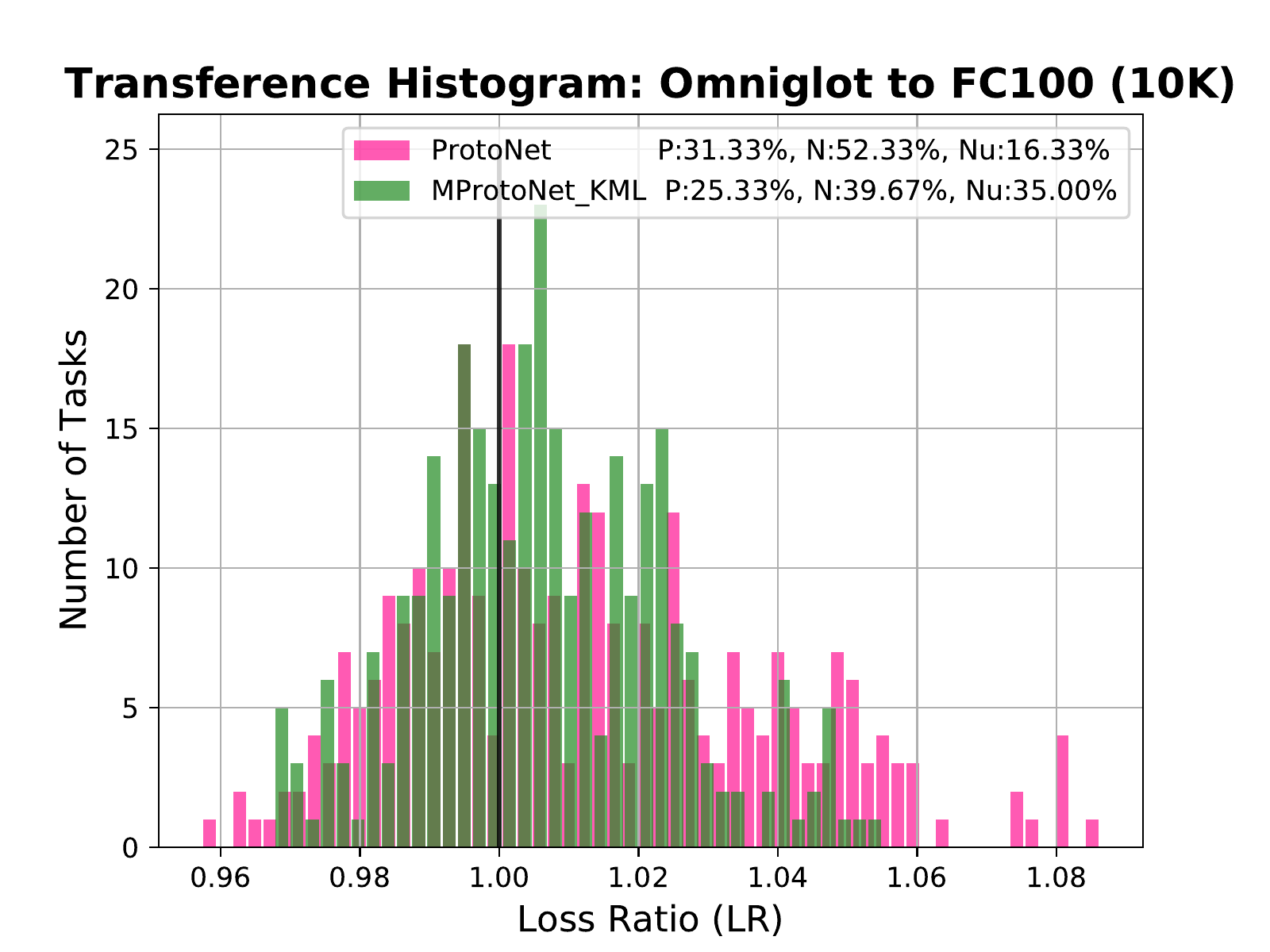}
  \caption{}
\end{subfigure}
\begin{subfigure}{.5\textwidth}
  \centering
  \includegraphics[clip, trim={0.5cm 0.1cm 0.4cm 0.5cm}, width=7cm]{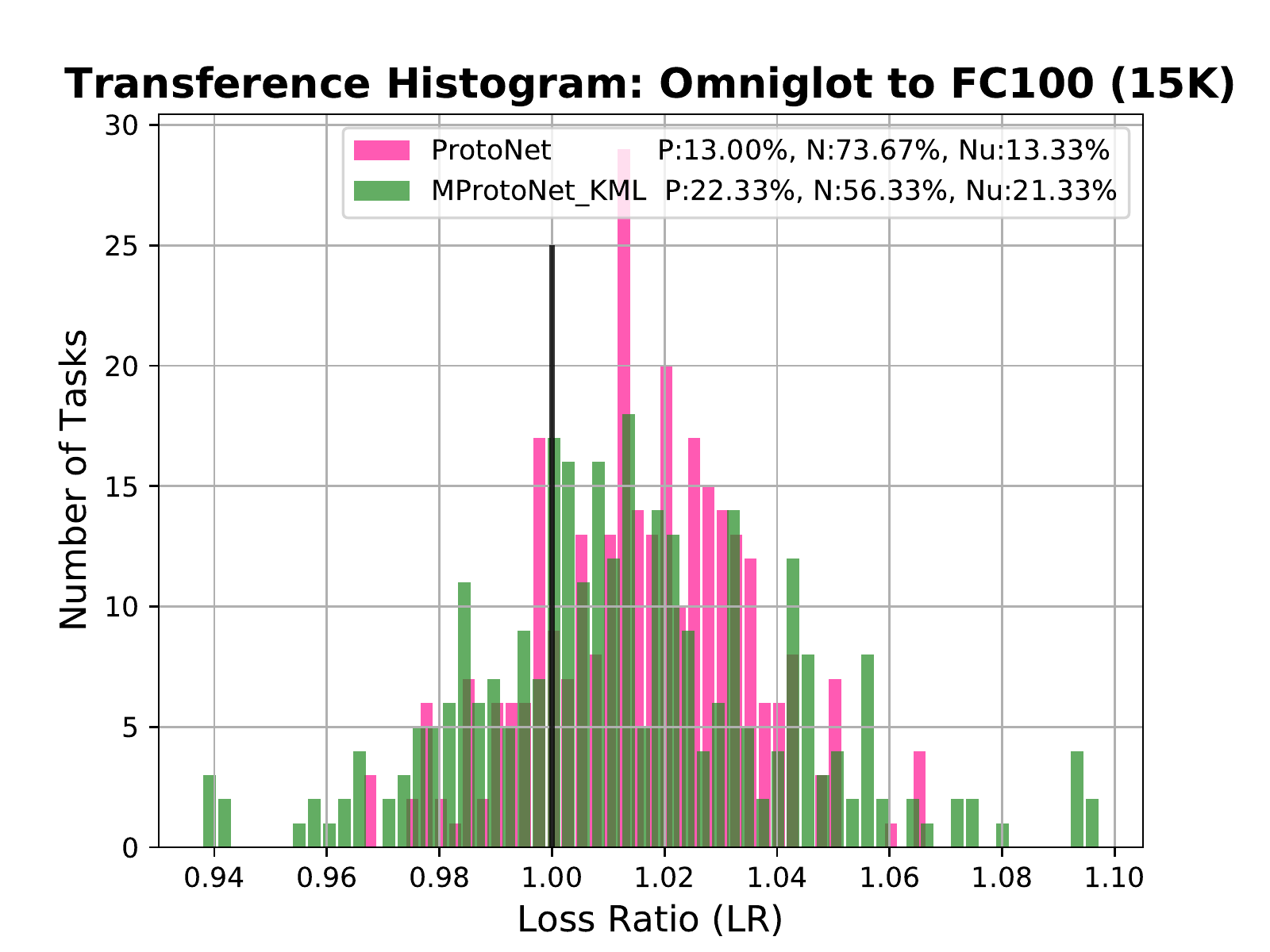}
  \caption{}
\end{subfigure}%
\begin{subfigure}{.5\textwidth}
  \centering
  \includegraphics[clip, trim={0.5cm 0.1cm 0.4cm 0.5cm}, width=7cm]{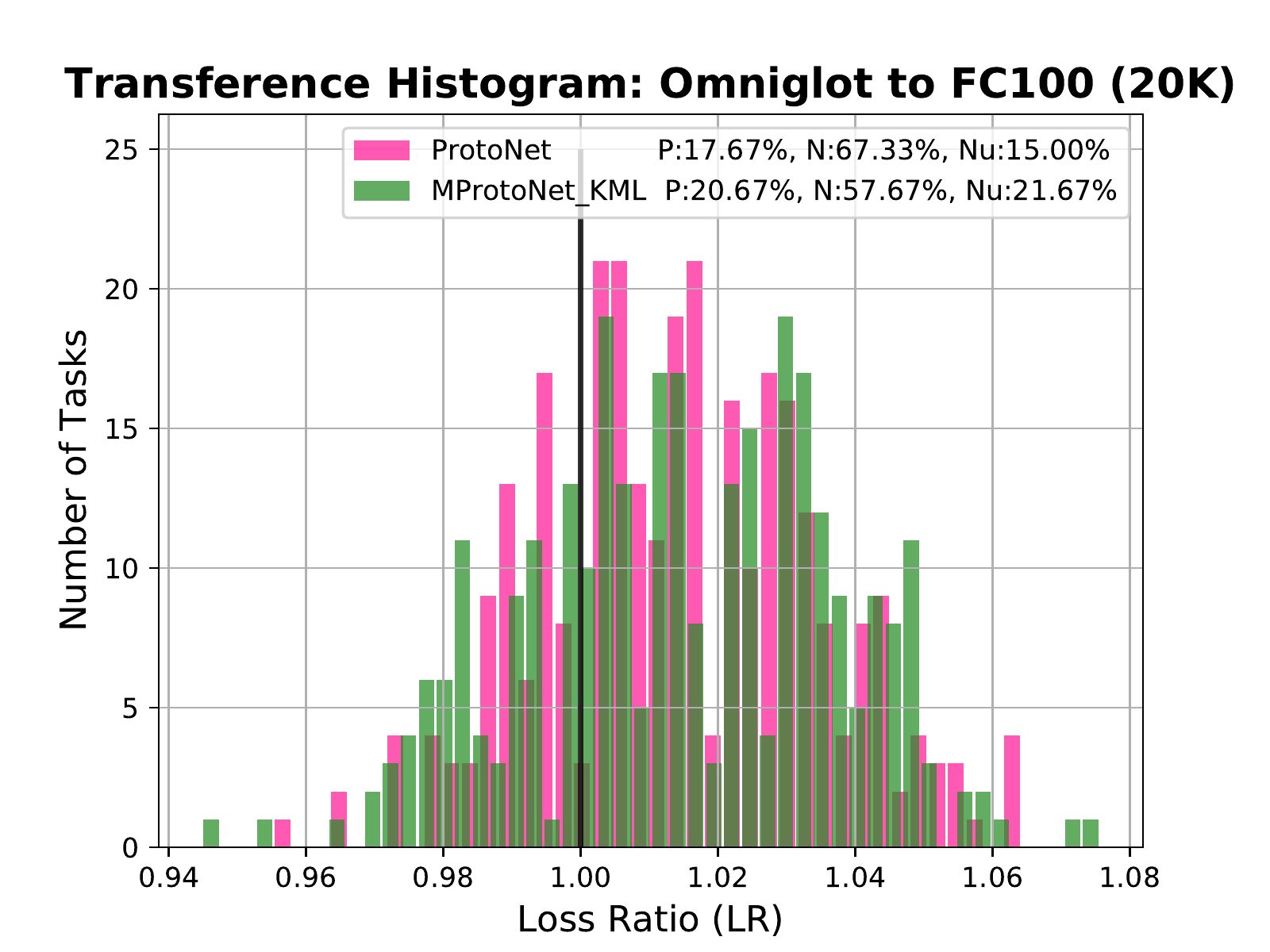}
  \caption{}
\end{subfigure}
\begin{subfigure}{.5\textwidth}
  \centering
  \includegraphics[clip, trim={0.5cm 0.1cm 0.4cm 0.5cm}, width=7cm]{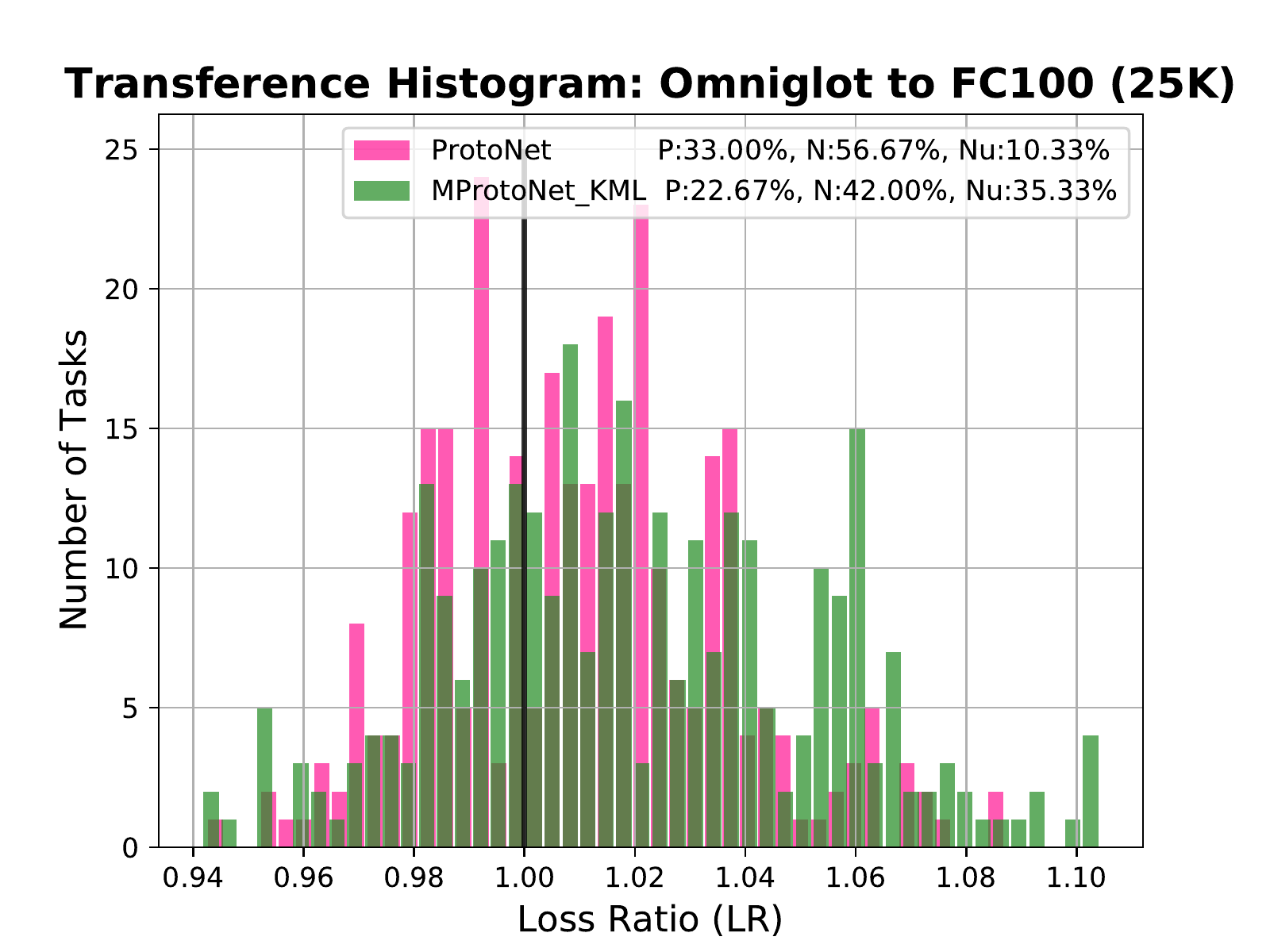}
  \caption{}
\end{subfigure}%
\begin{subfigure}{.5\textwidth}
  \centering
  \includegraphics[clip, trim={0.5cm 0.1cm 0.4cm 0.5cm}, width=7cm]{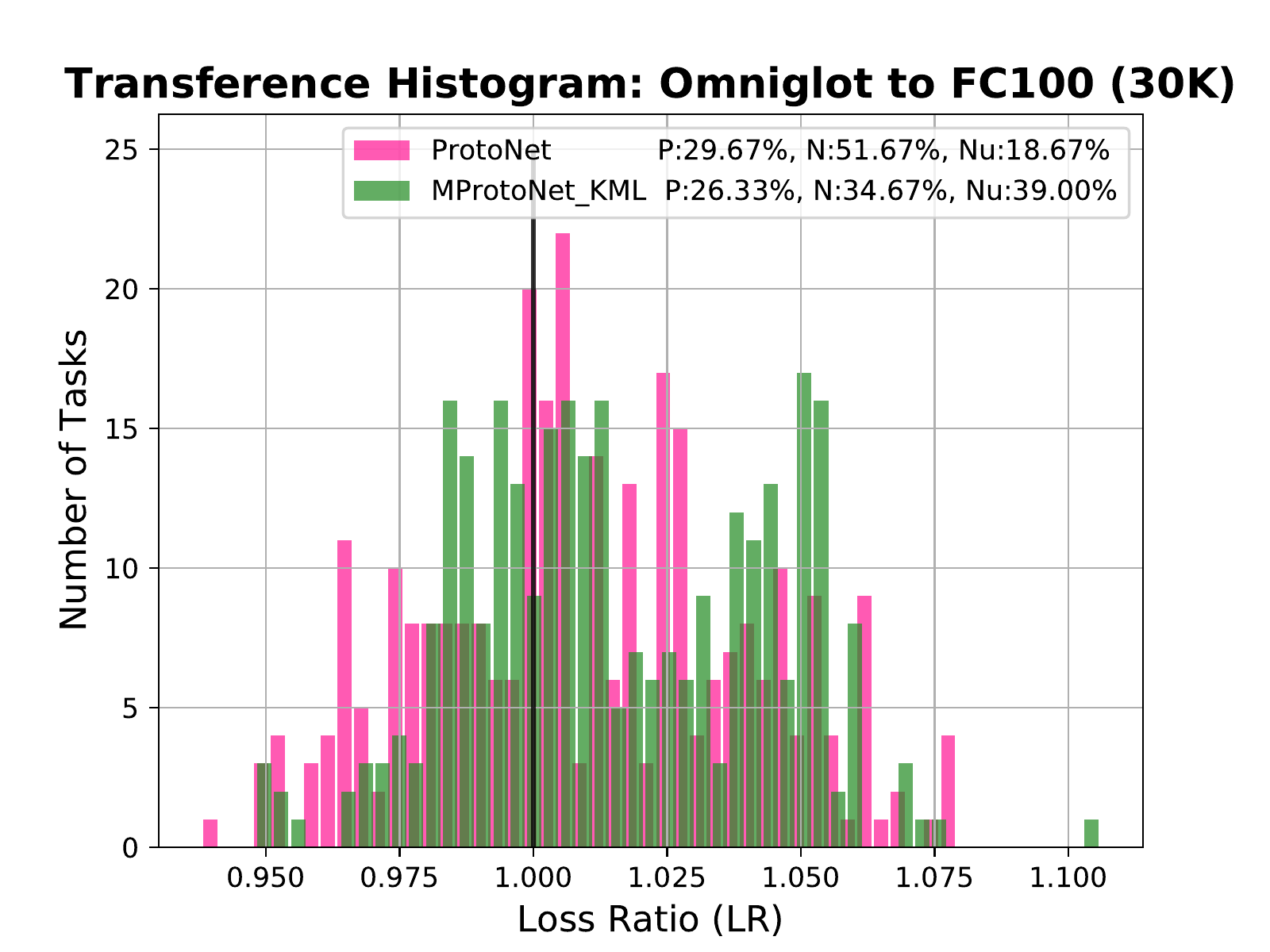}
  \caption{}
\end{subfigure}
\caption{Transference Histogram from Omniglot to FC100 task.}
\label{fig:transference_OC}
\end{figure}

\paragraph{Transference from Omniglot to FC100.}
Here we investigate the transference from Omniglot meta-train tasks to FC100 meta-test target task. Omniglot is a set of black-and-white images of handwritten characters with clean background. Therefore, classification of these tasks requires not much complicated feature representation, and the major challenge is limited number of data samples in each task. Please note that current meta-learning algorithms can easily handle the Omniglot few-shot classification tasks, and their performance is almost saturated on this dataset. However, by including the Omniglot in our analysis, we aim to investigate the dynamics of knowledge transfer from Omniglot to challenging FC100 few-shot tasks.

In this experiment we train both ProtoNet and proposed MProtoNet+KML on a meta-dataset constructed by combining Omniglot and FC100 few-shot tasks. We samples 300 meta-train Omniglot tasks, and a meta-test FC100 task as target task to perform transference analysis. The transference results are shown in figure \ref{fig:transference_OC}. Considering these results, similarly, as training proceeds the negative knowledge transfer reduces. An interesting phenomena is the high rate of positive transfer in the later training stages. In the later training stages, probably the model overfitts to meta-train classes by learning some features that can not generalize well for meta-test tasks. On the other hand, source meta-train tasks from Omniglot require strong but simpler features due to their samples type. So, a potential reason is that since Omniglot meta-train tasks emphasize on these features, they can prevent overfitting and have a positive impact on the generalization performance on the FC100 meta-test task.

\subsection{Transference within the same mode in Unimodal Few-Shot Learning}

\begin{figure}[h]
\centering
\begin{subfigure}{.5\textwidth}
  \centering
  \includegraphics[clip, trim={0.5cm 0.1cm 0.5cm 0.5cm}, width=7cm]{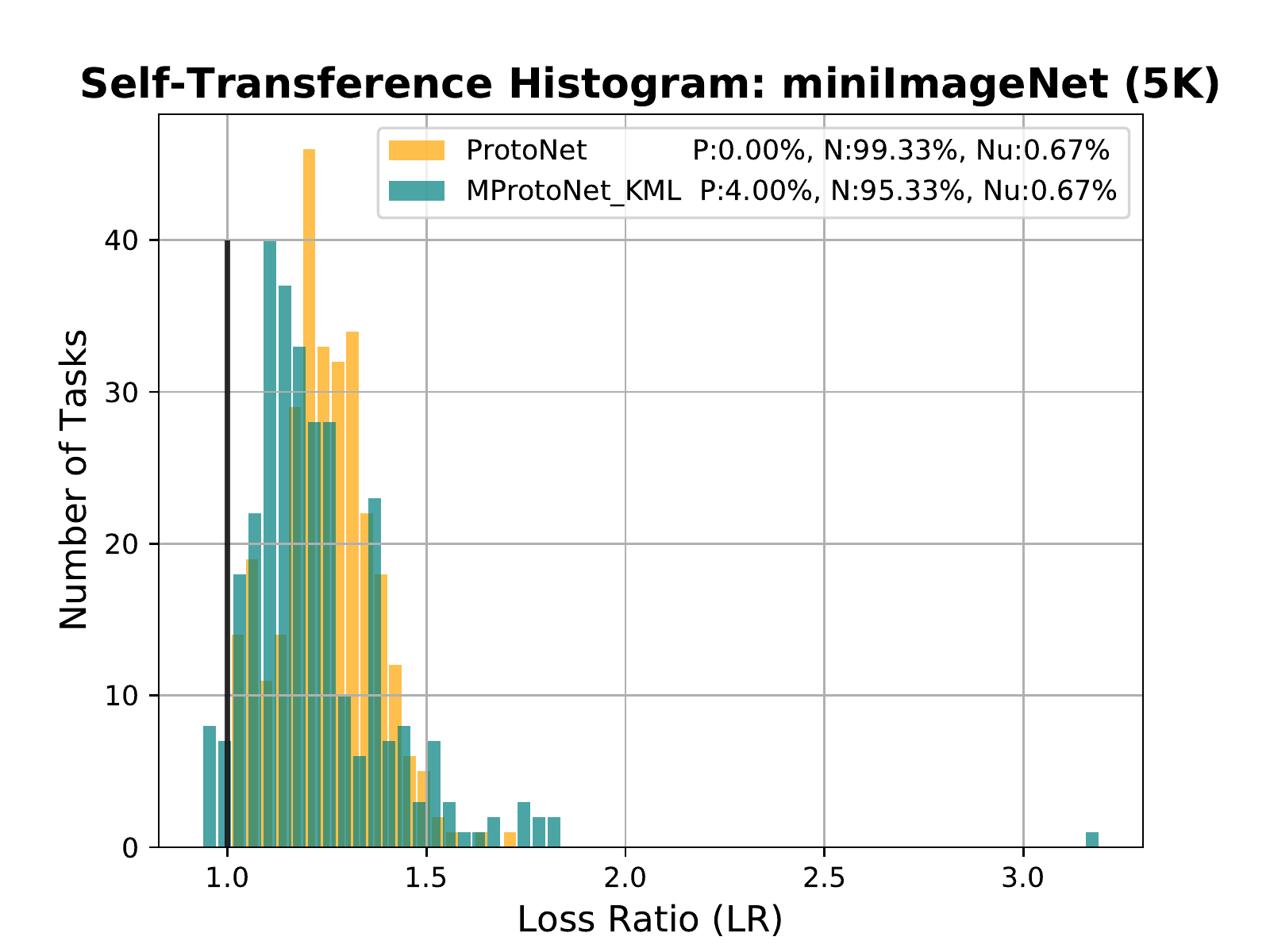}
  \caption{}
\end{subfigure}%
\begin{subfigure}{.5\textwidth}
  \centering
  \includegraphics[clip, trim={0.5cm 0.1cm 0.5cm 0.5cm}, width=7cm]{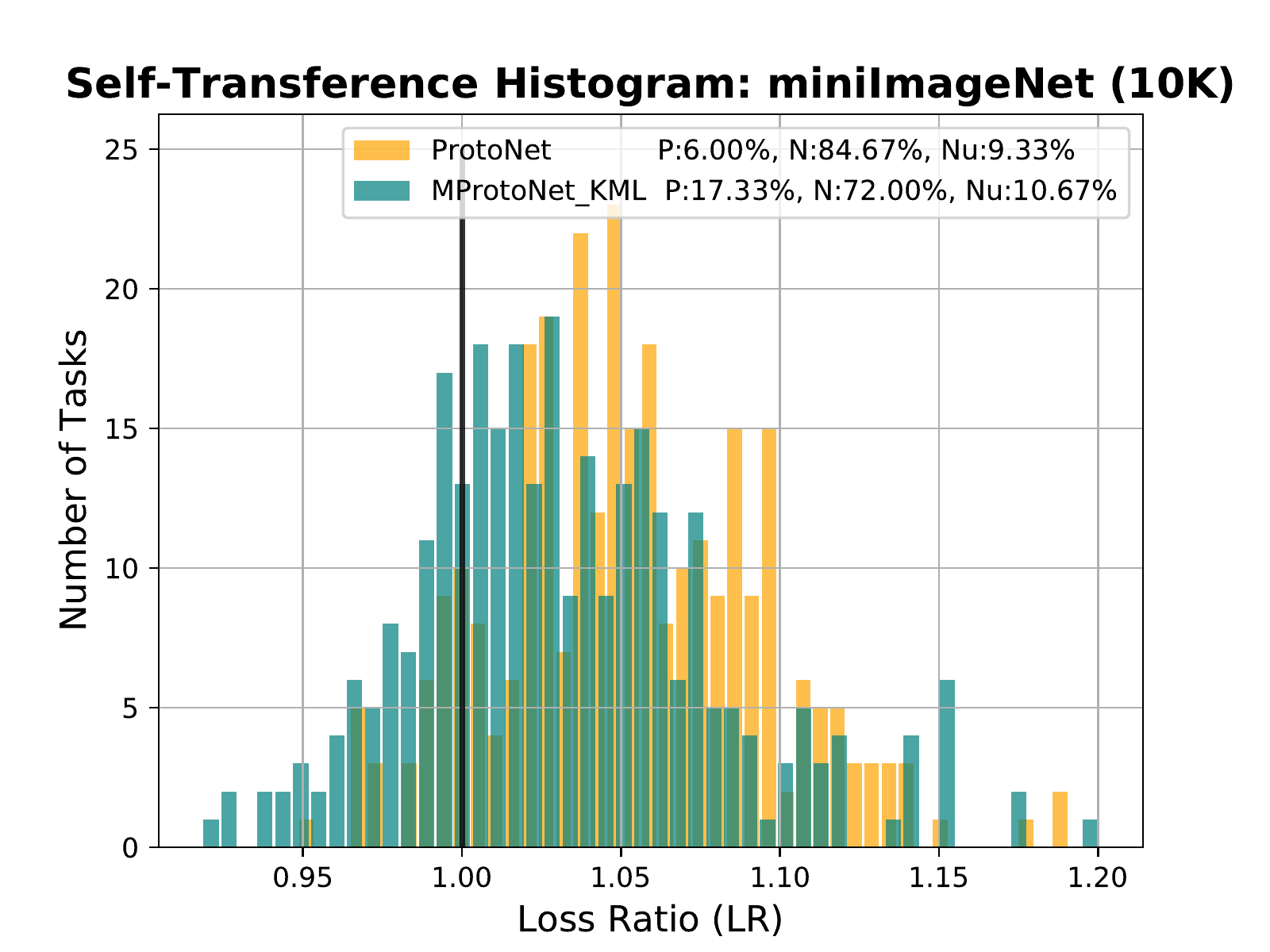}
  \caption{}
\end{subfigure}
\begin{subfigure}{.5\textwidth}
  \centering
  \includegraphics[clip, trim={0.5cm 0.1cm 0.5cm 0.5cm}, width=7cm]{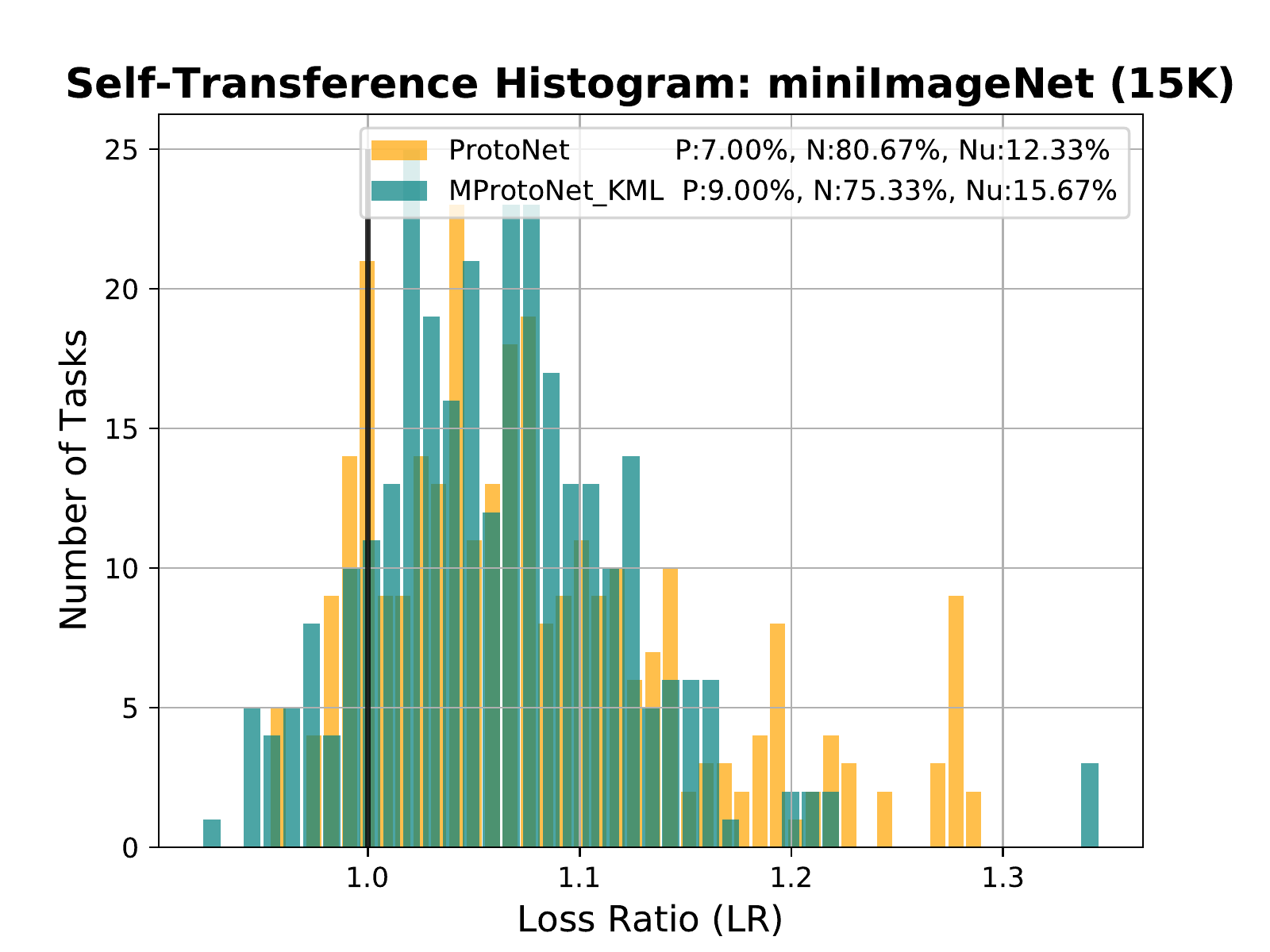}
  \caption{}
\end{subfigure}%
\begin{subfigure}{.5\textwidth}
  \centering
  \includegraphics[clip, trim={0.5cm 0.1cm 0.5cm 0.5cm}, width=7cm]{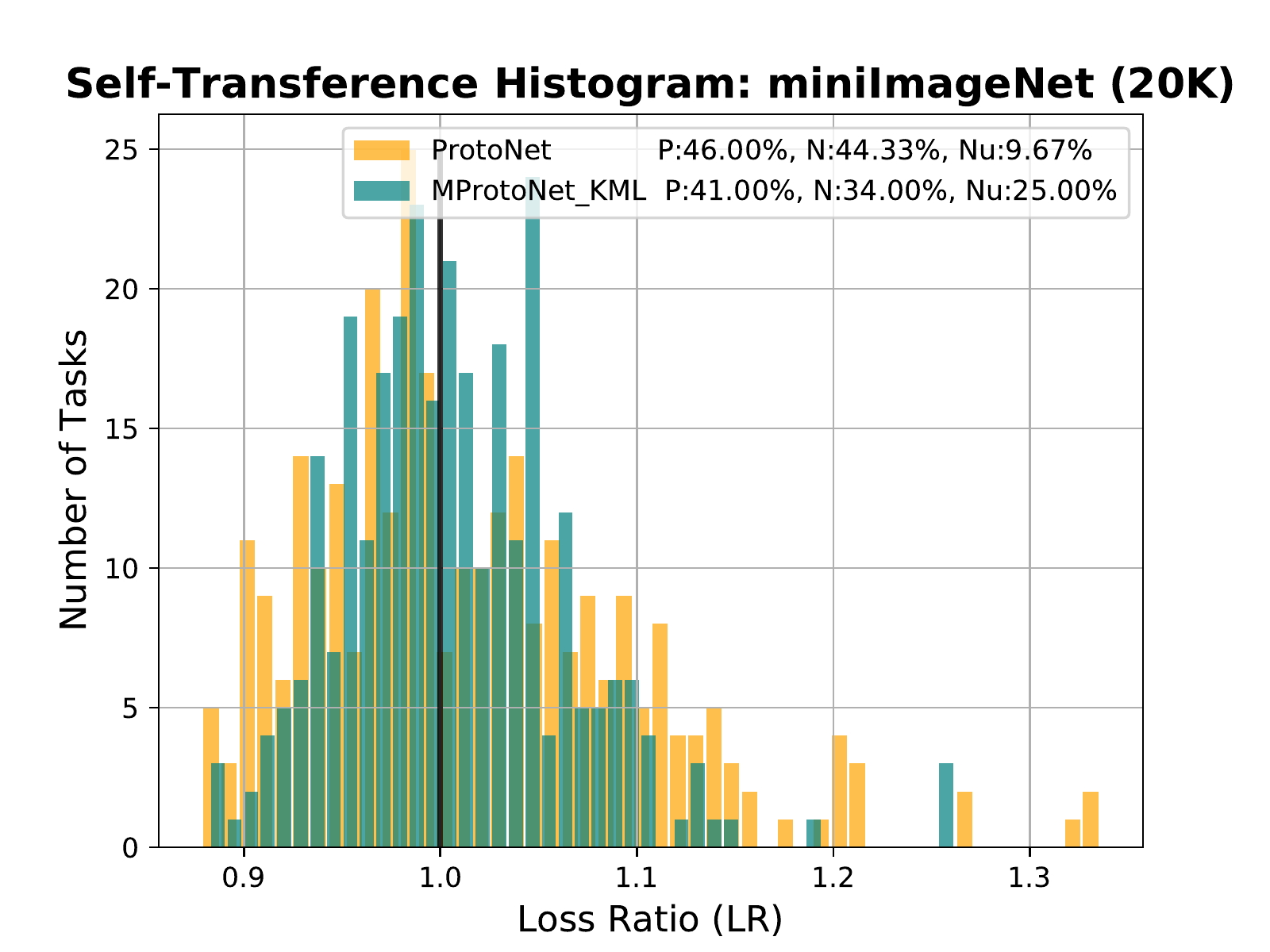}
  \caption{}
\end{subfigure}
\begin{subfigure}{.5\textwidth}
  \centering
  \includegraphics[clip, trim={0.5cm 0.1cm 0.5cm 0.5cm}, width=7cm]{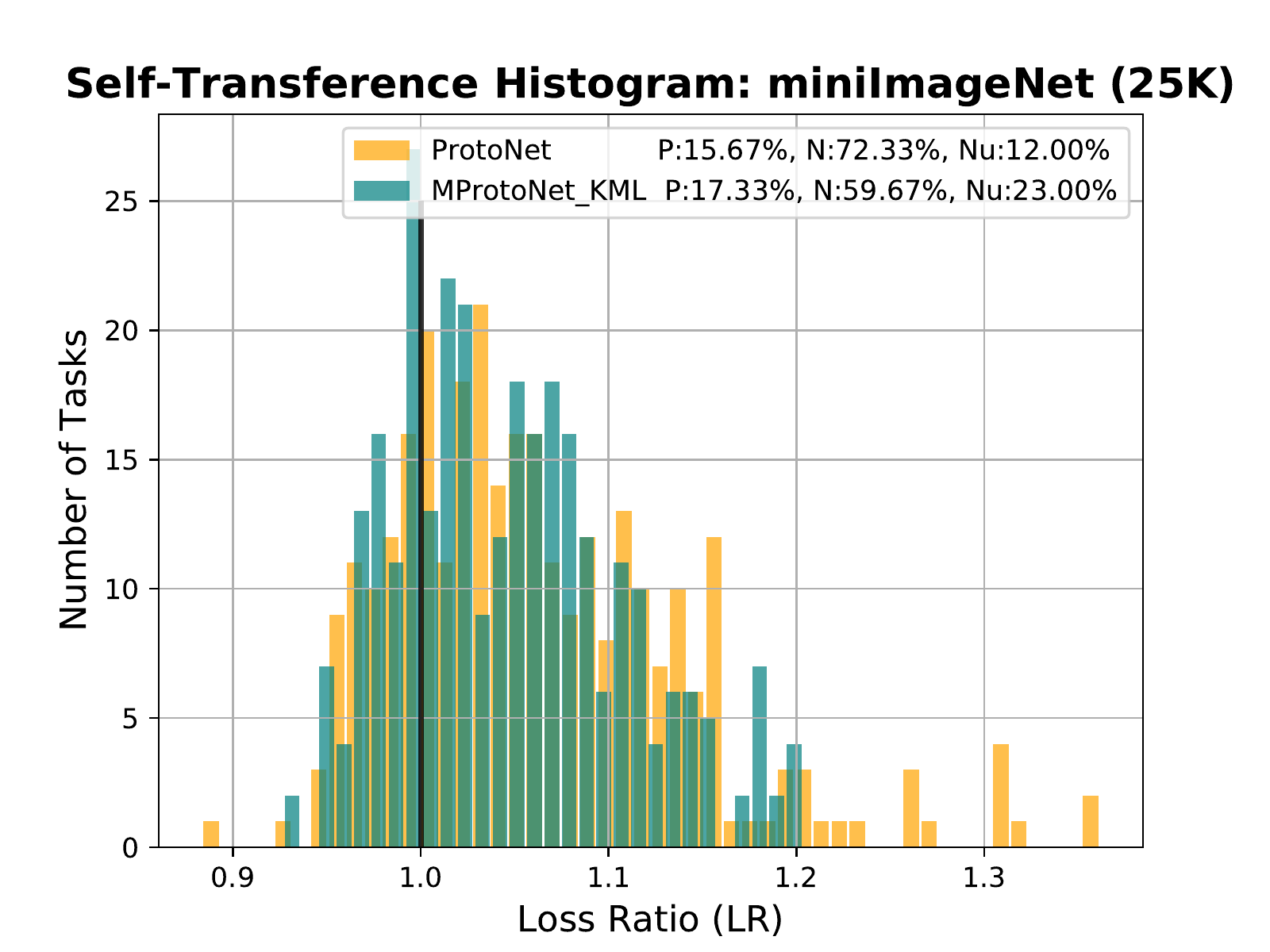}
  \caption{}
\end{subfigure}%
\begin{subfigure}{.5\textwidth}
  \centering
  \includegraphics[clip, trim={0.5cm 0.1cm 0.5cm 0.5cm}, width=7cm]{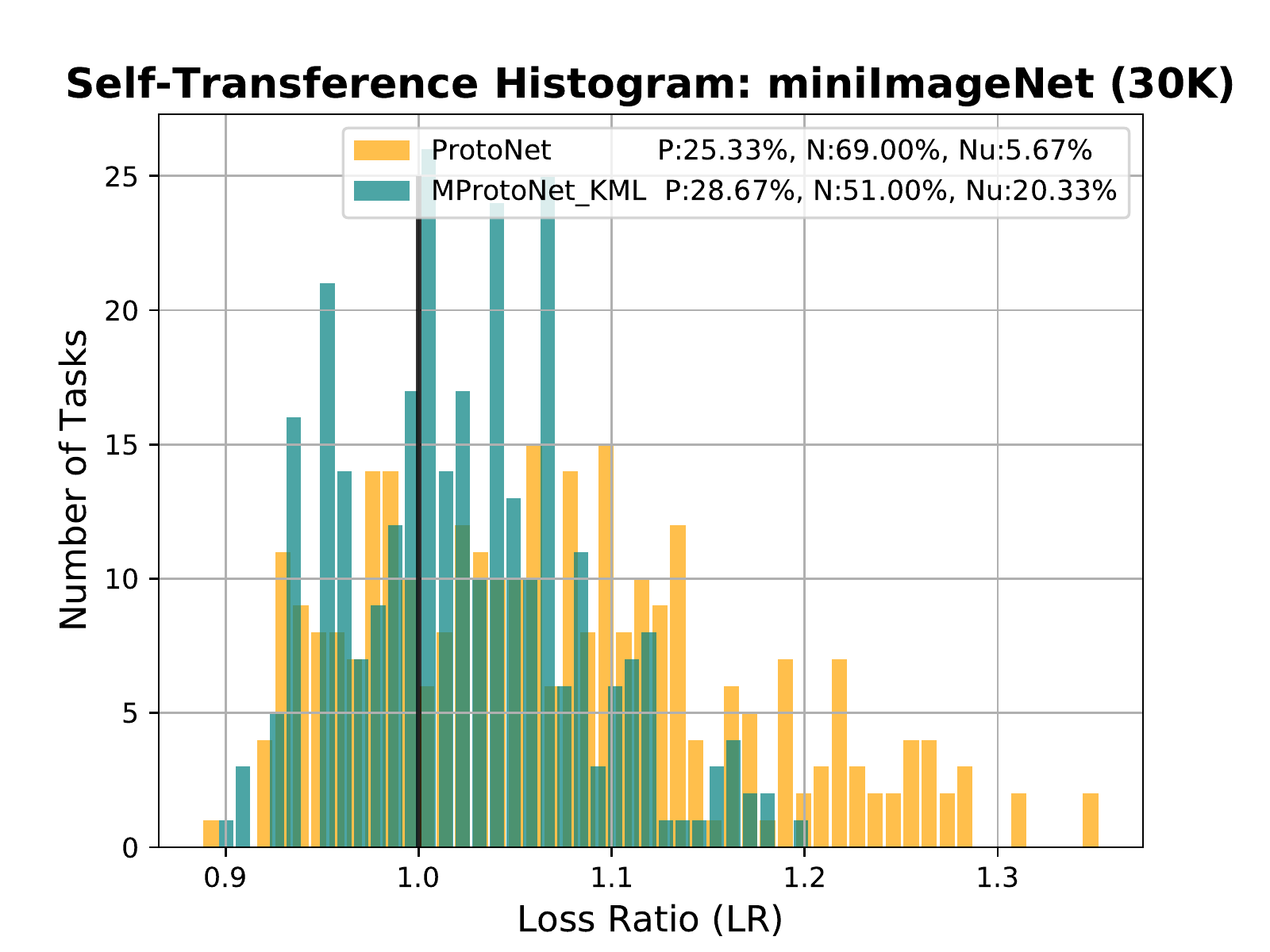}
  \caption{}
\end{subfigure}
\caption{Self-Transference Histogram within mini-ImageNet dataset.}
\label{fig:self-transference}
\end{figure}

In this section, we analyze the knowledge transfer between tasks within a conventional unimodal dataset. While a single dataset is defined to be one mode following the definition in~\cite{vuorio2019multimodal}, tasks from different classes can have a negative or positive impact on each other within a dataset. To investigate this, we analyze the transference from a number of meta-train miniImageNet tasks to a meta-test task from the same dataset. The experimental setup is exactly like the previous experiments, but the only difference is that we just train the model on the miniImageNet dataset (not combination of datasets). The transference results from 300 randomly sampled miniImageNet meta-train tasks on a target miniImageNet meta-test task is shown in figure \ref{fig:self-transference} for different training iterations.

Transference histograms show that there is a multimodality in terms of transference from meta-train tasks of a dataset to a target meta-test task from the same dataset. This means that a group of tasks have positive knowledge and others have negative. This can be interpreted with our previous findings. For example, in the simplest form, if the target meta-test task includes samples from animal classes, we expect the meta-train tasks from animal classes to have positive knowledge transfer, and tasks from non-animal classes to have negative transfer. Considering this behaviour within tasks in mini-ImageNet dataset, applying proposed KML scheme boosts the performance by reducing negative transfer (also shown in histograms) through assigning task-aware layers.

\section{Reinforcement Learning Results}
\label{sec:RL_results}
The proposed KML idea can be extended to Reinforcement Learning (LR) environments,  when using an optimization-based meta-learner like MAML. we have applied our KML algorithm on the official code of MMAML for RL experiments on three different environments used in ~\cite{vuorio2019multimodal}: Point Mass, Reacher, and Ant. Similar to ~\cite{vuorio2019multimodal}, for each environment, the goals are sampled from a multimodal goal distribution, with similar environment-specific parameters as ~\cite{vuorio2019multimodal}. To have a fair comparison, we have kept all other hyperparameters the same as the ~\cite{vuorio2019multimodal}. The mean and standard deviation of cumulative reward per episode for multimodal reinforcement learning problems with 2, 4 and 6 modes are shown in table \ref{table:RL_experiments}. Results show that our proposed KML can achieve gain over ~\cite{vuorio2019multimodal} in all of the RL experiment setups.

\begin{table}[h]
\caption{The mean and standard deviation of cumulative reward per episode for multimodal reinforcement learning problems with 2, 4 and 6 modes reported across 3 random seeds.}
\label{table:RL_experiments}
\begin{threeparttable}
\begin{center}
    \resizebox{\linewidth}{!}{
			\begin{tabular}{lcccccccc}
				\toprule
				\multirow{2}{*}{\bf Method} & \multicolumn{3}{c}{\bf Point Mass 2D} &
				\multicolumn{3}{c}{\bf Reacher}&
				\multicolumn{2}{c}{\bf Ant}
				\\
				\cmidrule(l){2-4}
				\cmidrule(r){5-7}
				\cmidrule(r){8-9}
                 & {\bf 2Modes} & {\bf 4Modes} & {\bf 6Modes} & {\bf 2Modes} & {\bf 4Modes} & {\bf 6Modes} & {\bf 2Modes} & {\bf 4Modes}\\
				\midrule
				
				\textbf{MMAML}&
				-136$\pm$8&
				-209$\pm$32&
				-169$\pm$48&
				-10.0$\pm$1.0&
				-11.0$\pm$0.8&
				-10.9$\pm$1.1&
				-711$\pm$25&
				-904$\pm$37\\
				\midrule
				
				\textbf{MMAML+KML(ours)}&
				-121$\pm$9&
				-197$\pm$30&
				-161$\pm$41&
				-9.6$\pm$1.0&
				-10.6$\pm$0.7&
				-10.6$\pm$1.0&
				-689$\pm$23&
				-891$\pm$36\\
				\bottomrule
			\end{tabular}
		}
\end{center}
\end{threeparttable}
\end{table}

\section{Verification of New Interpretation}
\label{sec:verification}

In Sec. 5.1. of the main paper, we propose a new interpretation of modulation scheme used in MMAML. Briefly, we show that feature-wise linear modulation (FiLM) ~\cite{perez2018film} applied to each channel of feature map in MMAML, can be considered as convolving the input with modulated kernel $\hat{\mathbf{W}}_{i} = \eta_{i}\mathbf{W}_{i}$ and adding the modulated bias term $\hat{b}_{i} = \eta_{i} b_{i} + \gamma_{i}$. In addition to the mathematical formulation provided in the main paper, we also verify this new interpretation with experiments.

\begin{figure}[h]
 \begin{center}
 \includegraphics[ clip, trim={1.85cm 18cm 5cm 1.65cm},   width=6cm]{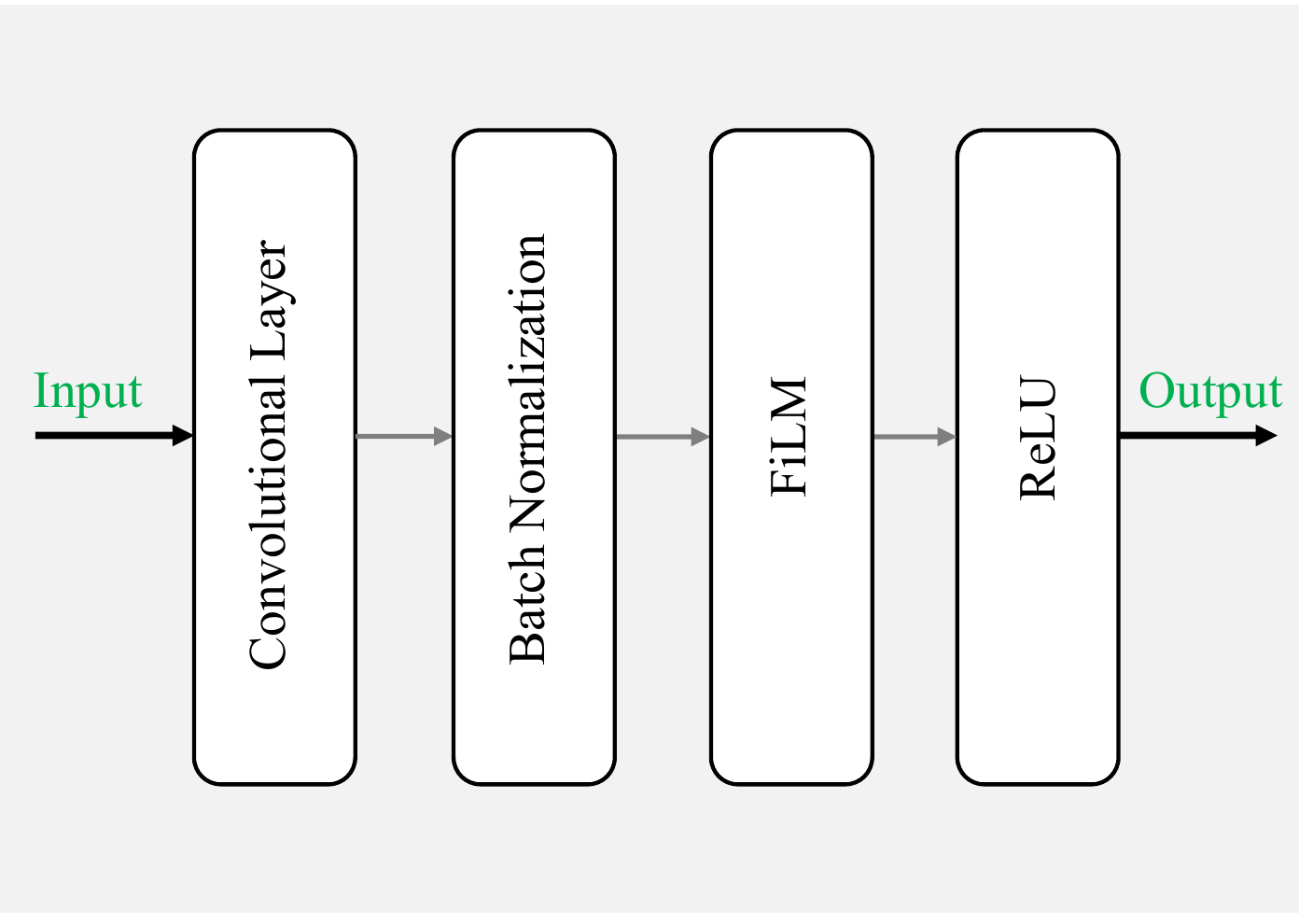}
 \end{center}
   \caption{The structure of each layer in MMAML.}
\label{fig:MMAML_Layer}
\end{figure}

{\em First,}, we compare generated featuremaps by the proposed new interpretation for MMAML (\emph{new}-MMAML) with the ones generated by the original implementation of MMAML. For implementing the new interpretation of MMAML, we use the official code provided by authors with exactly same hyperparameters. Figure \ref{fig:MMAML_Layer} shows the structure of each Convolutional Block used in MMAML for few-shot image classification. Original MMAML implementation uses the Batch Normalization (BN) layer before modulation. So, for a fair comparison, we disable the BN layer in MMAML and extract the feature maps after applying FiLM. Then for \emph{new}-MMAML, we simply perform convolution with modulated parameters $\hat{\mathbf{W}}_{i}$ and $\hat{b}_{i}$ to produce featuremaps. The produced featuremaps are same for all CNN layers. For example, for a mini-ImageNet classification task, the average error between two produced featuremaps in first layer is around $\mathbf{3.7\mathrm{e}{-3}}$ while the average absolute value of feature maps is around $\mathbf{1.4\mathrm{e}{+2}}$. Note that this minor error probably stems from the floating-point round-off error in PyTorch ~\cite{paszke2019pytorch}.

\begin{table}[h]
	\caption{Comparison of meta-test accuracies for original implementation of MMAML with proposed new interpretation (\emph{new}-MMAML) for few-shot classification on multimodal scenario. }
	\label{table:MMAML vs new-MMAML}
	\begin{center}
		\resizebox{0.99\linewidth}{!}{
			\begin{tabular}{lcccccc}
				\toprule
				\multirow{2}{*}{\textbf{Method}} & \multicolumn{2}{c}{\textbf{2 Mode}} & \multicolumn{2}{c}{\textbf{3 Mode}} & \multicolumn{2}{c}{\textbf{5 Mode}}\\
				\cmidrule(r){2-3}
				\cmidrule(r){4-5}
				\cmidrule(r){6-7}
				 \multirow{2}{*}{} & 1-shot & 5-shot & 1-shot & 5-shot & 1-shot & 5-shot  \\
				\midrule
				\textbf{MMAML} &
				67.67$\pm$0.63\% & 
				73.52$\pm$0.71\% &  57.35$\pm$0.61\%& 
				64.21$\pm$0.57\%& 
				49.53$\pm$0.50\%& 
				58.89$\pm$0.47\%\\
				\midrule
				\textbf{\emph{new}-MMAML} & 
				67.43$\pm$0.61\%& 
				73.64$\pm$0.66\% & 
				57.44$\pm$0.60\%& 
				64.09$\pm$0.61\%& 
				49.23$\pm$0.51\%& 
				58.71$\pm$0.44\%\\
				\bottomrule
			\end{tabular}
		}
	\end{center}
\end{table}

{\em Second}, we compare the training performance of MMAML and \emph{new}-MMAML. In the official implementation of MMAML, the affine transform of the BN layer ~\cite{ioffe2015batch} is disabled, due to the similar functionality performed by FiLM. However, since in \emph{new}-MMAML we are using modulated parameters, we enable the affine transform of the BN layer. Meta-test results are shown in table \ref{table:MMAML vs new-MMAML} for different multimodal image classification modes. As the results suggest the \emph{new}-MMAML has almost the same performance as the MMAML which also verifies the proposed interpretation through experiments. The minor difference between the results is due to the difference between the BN layer in two implementations (as discussed).

\section{Parameter Reduction in Proposed Parameter Generator}
\label{sec:param_reduction}
In Sec. 5.2 of the main paper we have proposed a structure to reduce the number of parameters in the modulation parameter generator network $\mathbf{g}_{\phi}$. As discussed, proposed structure includes three smaller MLP modules. Here a more detailed comparison between the proposed simplified structure and a single MLP is provided. 
A standard convolutional layer is parameterized by convolution kernel of size $N_{k} \times N_{k} \times N_{i} \times N_{o}$ and a bias term of size $N_{o}$, where $N_{k}$ is the spatial dimension of kernel, $N_{i}$ is the number of input channels and $N_{o}$ is the number of output channels. Then the required number of parameters for an MLP with single hidden layer that takes the task embeddings with size $N_{\upsilon}$ to produce the whole elements for this layer is:
$N_{\upsilon} \times (N_{k} \times N_{k} \times N_{i} \times N_{o} + N_{o})$.
Instead using the proposed structured MLP, we use three smaller MLPs to produce $N_{o}$, $N_{i} \times N_{k} \times N_{k}$ and $N_{o}$ parameters, respectively. Then the parameter reduction ratio compared to single MLP is:

\begin{align*}
\frac{N_{k} \times N_{k} \times N_{i} \times N_{o} + N_{o}}{N_{k} \times N_{k} \times N_{i} + 2N_{o}}
\end{align*}

Following the structure proposed in ~\cite{vuorio2019multimodal}, the base network consists of four convolutional layer with the channel size 32, 64, 128 and 256. In our KML scheme we intend to produce a  modulation parameter for each parameter of the base network using task embedding $\boldsymbol{\upsilon}$ (with a dimension of 128) as input. Table \ref{table:MLPvsProposed} compares the number of parameters in single  MLP and proposed structure when used as parameter generation network for each layer. The total number is also provided. As one can see, proposed structure reduces the number of the parameters by a factor of \textbf{152}.

\begin{table}[h]
	    \caption{Number of parameters required in each structure as modulation parameter generator $\mathbf{g}_{\phi}$.}
	\label{table:MLPvsProposed}
	\begin{center}
		\resizebox{0.45\linewidth}{!}
		{
			\begin{tabular}{lcc}
				\toprule
				\multirow{2}{*}{\textbf{Layer}} & \multicolumn{2}{c}{\textbf{Number of parameters}} \\
				\cmidrule(r){2-3}
				 & \textbf{MLP} & \textbf{Proposed Structure}\\
				\midrule
				\#1 & 114,688 &
				11,648\\
				\midrule
				\#2 &
				2,367,488 &
				45,056\\
				\midrule
				\#3 & 9,453,568 &
				90,112\\
				\midrule
				\#4 &
				37,781,504 &
				180,224\\
				\midrule
				\textbf{Total} & \textbf{49,717,248} &
				\textbf{327,040}\\
				\bottomrule
			\end{tabular}
		}
	\end{center}
\end{table}

\begin{figure}[h]
 \begin{center}
 \includegraphics[clip, trim={0.4cm 0.1cm 1cm 1cm},   width=9cm]{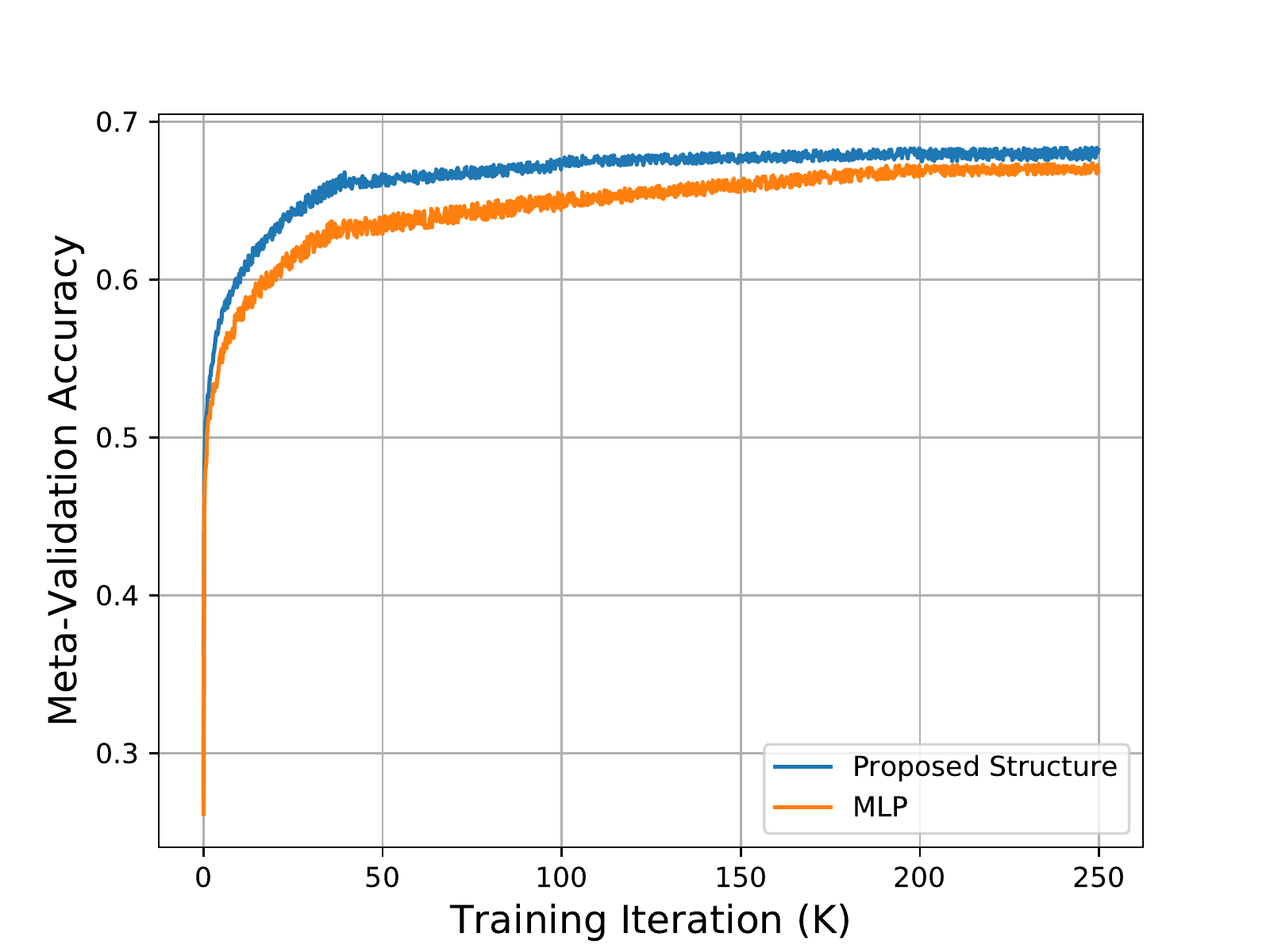}
 \end{center}
   \caption{The meta-validation accuracy during meta-training.}
\label{fig:meta-validation_mlp_vs_proposed}
\end{figure}

We also compare using MLP with proposed structure in terms of convergence
speed. We use the same hyperparameters for training both models.
The accuracy of meta-validation set during meta-training on 3Mode,
5-way 1-shot setting is plotted in figure \ref{fig:meta-validation_mlp_vs_proposed}.
We can clearly see that using the proposed structure as modulation parameter generator, the network converges faster and also yields better results in term of accuracy compared to single MLP.
We have also provided the meta-test accuracy results for 3Mode few-shot classification in Table \ref{table:MLPvsProposed_meta-test_accuracy}. These results also support the improved performance of the proposed simplified structure versus single MLP.

\begin{table}[h]
	    \caption{2Mode Meta-test accuracy of the proposed simplified structure versus single MLP for 2Mode 5-way scenario.}
	\label{table:MLPvsProposed_meta-test_accuracy}
	\begin{center}
		\resizebox{0.65\linewidth}{!}
		{
			\begin{tabular}{lcc}
				\toprule
				\multirow{2}{*}{\textbf{Method}} & \multicolumn{2}{c}{\textbf{Setup}} \\
				\cmidrule(r){2-3}
				 & \textbf{1shot} & \textbf{5shot}\\
				\midrule
				
				{\bf MLP} &
				61.22$\pm$0.56\% &
				69.38$\pm$0.48\%\\
				\midrule
				{\bf Proposed Simplified Structure} &
				{\bf 62.08$\pm$0.54\%}&
				{\bf 70.03$\pm$0.43\%}\\
				\bottomrule
			\end{tabular}
		}
	\end{center}
\end{table}

\section{KML vs FiLM: Parameter and Computational Overhead}
\label{sec:kml_vs_film_overhead}
Previously, we have demonstrated that replacing FiLM with the proposed KML method significantly improves the accuracy of the meta-learner in both multimodal and conventional unimodal few-shot classification. KML achieves this substantial improvement by modulating the whole elements of the kernel instead of applying the affine transform on the feature maps (FiLM in ~\cite{vuorio2019multimodal}). This is done by generating a larger number of parameters compared to FiLM. Here we analyze the overhead introduced by replacing FiLM (existing method in ~\cite{vuorio2019multimodal}) with KML (our proposed method).

First, we discuss the number of additional parameters introduced by KML. Since the only difference between the two methods is on the generator, we consider this module for comparison. Recalling from section D of the supplementary, the number of parameters required in proposed simplified structure in KML (for a layer) is $N_{\upsilon} \times (N_{k} \times N_{k} \times N_{i} + 2N_{o})$. Since in FiLM, only two parameters are generated for each channel of the convolutional layer, the number of parameters in the generator is $N_{\upsilon} \times (2N_{o})$. Therefore, the additional overhead of KML for the generator becomes $N_{\upsilon} \times N_{k} \times N_{k} \times N_{i}$ for each layer. Considering we have four convolutional layers in our structure and for each layer, a separate generator is used, in total, KML adds around 261.5 K parameters to the ones in FiLM. Considering this number, for example, the total number of parameters in MProtoNet+KML (our method) increases by 22.9\% compared to MProtoNet (existing method in ~\cite{vuorio2019multimodal}).

\begin{table}[h]
	    \caption{Training time for MProtoNet and MProtoNet+KML (proposed method) for 2Mode setup.}
	\label{table:FiLMvsKML_overhead}
	\begin{center}
		\resizebox{0.65\linewidth}{!}
		{
			\begin{tabular}{lcc}
				\toprule
				\textbf{Method} & \textbf{Training Time} &
				\textbf{Accuracy}
				\\
				\midrule
				
				{\bf MProtoNet} &
				173 minutes &
				56.03$\pm$0.64\%\\
				\midrule
				{\bf MProtoNet+KML(proposed)} &
				182 minutes&
				59.31$\pm$0.62\%\\
				\bottomrule
			\end{tabular}
		}
	\end{center}
\end{table}
Second, in terms of computational overhead, the table \ref{table:FiLMvsKML_overhead} shows the total training time for MProtoNet (existing method in ~\cite{vuorio2019multimodal}) and MProtoNet+KML (ours) for 2Mode (combination of Omniglot and miniImageNet), 5-way 5-shot scenario. As the results show, the computational overhead of the proposed method in training time is around 5.2\%. Similar training results are obtained for the other setups (3 Mode, 5Mode). Also please note that the inference time of our method and existing method ~\cite{vuorio2019multimodal} are almost the same (on average 0.087 seconds for each mini-batch of few-shot tasks).


\section{Simplified Parameter Generator as Low-Rank Approximation}
\label{sec:simplified_structure}

The proposed simplified structure (figure 4 of the main paper) can be considered as a low-rank approximation (1-rank). In the section \ref{sec:param_reduction} of this supplementary, we have shown that this 1-rank approximation achieves better meta-test results compared to a full-rank version. Here we check the results for higher ranks (2-rank and 3-rank). For producing the 2-rank approximation, we produce two different matrices: $M_1$ and $M_2$ using a similar method as (8) in our paper, and then add these two matrices to generate the final modulation matrix $M = M_1 + M_2$. Please note that this time instead of 3 modules, we have 5 modules in our simplified structure. Two pairs of modules are used to generate the $M_1$ and $M_2$, and the fifth one is used to generate the bias term. We have checked these vectors to be independent. A similar procedure is used to design a 3-rank approximation of the MLP using three different pairs.

\begin{table}[h]
	    \caption{Meta-test accuracies for 2Mode setup with different rank approximation in simplified parameter generator $g_\phi$.}
	\label{table:ranks_parameter_generator}
	\begin{center}
		\resizebox{\linewidth}{!}
		{
			\begin{tabular}{lcccc}
				\toprule
				\textbf{Setup} & \textbf{MProtoNet} &
				\textbf{MProtoNet+KML(1-rank)} & \textbf{MProtoNet+KML(2-rank)} &
				\textbf{MProtoNet+KML(3-rank)}
				\\
				\midrule
				
				{\bf 5way-1shot} &
				70.60$\pm$0.56\% &
				73.69$\pm$0.52\% &
				72.12$\pm$0.54\% &
				72.06$\pm$0.52\%\\
				\midrule
				{\bf 5way-5shot} &
				75.72$\pm$0.47\% &
				79.82$\pm$0.40\% &
				78.94$\pm$0.43\% &
				78.70$\pm$0.46\%\\
				\bottomrule
			\end{tabular}
		}
	\end{center}
\end{table}
The meta-test results for 2Mode classification are shown in table \ref{table:ranks_parameter_generator}. As results suggest, the 2-rank and 3-rank approximations still have better performance compared to the MProtoNet. However, the performance is degraded compared to the 1-rank approximation. The possible reason could be overfitting of 2-rank and 3-rank versions due to more parameters.

\section{KML for Visual Reasoning}
\label{sec:kml_vr_results}

We remark that based on the proposed new interpretation of the FiLM scheme, our proposed KML can be seen as a generalization of FiLM. So we can expect that applying KML to the areas improved by FiLM may bring some further improvement. We also declare that the amount of improvement depends on the underlying structure of learning tasks. For example in the case of few-shot learning (especially multimodal distribution), since there could be a significant difference between different tasks (e.g., digit classification vs natural object classification), KML brings a large improvement over FiLM by letting the more powerful adaption of kernels for each few-shot task. Intuitively, this improvement may be less for the applications where more similar kernels are required for different tasks, e.g., visual reasoning on CLEVR dataset where there is a significantly lower variation on image statistics compared to our multimodal few-shot distribution, and the main difference originates from the question, and probably program signal.

\begin{table}[h]
	    \caption{Results of applying the proposed KML on visual reasoning dataset CLEVR.}
	\label{table:visual_reasoning}
	\begin{center}
		\resizebox{\linewidth}{!}
		{
			\begin{tabular}{lcccccc}
				\toprule
				\textbf{Method} & \textbf{Count} &
				\textbf{Exist} & \textbf{Compare Numbers} &
				\textbf{Query Attribute} &
				\textbf{Compare Attribute} &
				\textbf{Average}
				\\
				\midrule
				CNN+GRU+FiLM~\cite{perez2018film}&
				94.3\% &
				99.1\% &
				96.8\% &
				99.1\% &
				{\bf99.1\%} &
				97.7\%\\
				\midrule
			    CNN+GRU+KML(ours)&
				{\bf96.1\%} &
				{\bf99.5\%} &
				{\bf97.1\%} &
				{\bf99.3\%} &
				{\bf99.1\%} &
				{\bf98.2\%} \\
				\bottomrule
			\end{tabular}
		}
	\end{center}
\end{table}

We applied the KML to the CLEVR dataset by replacing KML with the FiLM in the official code of the FiLM paper ~\cite{perez2018film}. The results are shown in table \ref{table:visual_reasoning}. As the results show, KML on average improves the FiLM by 0.5\% in the 5 question types. Note that KML obtains this improvement while FiLM has achieved very high accuracy already.

\section{Gradient Flow From Meta-learner to Modulation Network}
\label{sec:gradient_flow_metalearner}
In general multimodal meta-learning framework, meta-learning algorithm plays two important  roles. {\em First}, its meta-learning capability directly affects the multimodal meta-learning performance. {\em Second}, modulation network is fed with the gradient propagated from the meta-learner in each training iteration. In the case of good gradient propagation, the modulation network can be trained well to predict the task mode and generate powerful modulation parameters. So, good gradient flow from meta-learner to modulation network is as important as the good performance of the meta-learner itself. 

{\bf Limitation of MMAML in Terms of Gradient Flow.} While proposing a generic framework, another limitation of MMAML is due to using MAML as meta-learner. MAML is a simple and elegant meta-learning algorithm, however, backpropagating the gradients through inner-loop (with multiple updates) requires the second order gradients for optimization. This increases the computational complexity of meta-learning and makes it difficult for gradients to propagate ~\cite{antoniou2018how}. This problem gets even worse when using MAML in multimodal meta-learning framework where the modulation network and meta-learner are supposed to be trained together in an end-to-end fashion. We empirically found that multimodal meta-learning framework can benefit from using a more simpler meta-learner like ProtoNet ~\cite{snell2017prototypical} which has better gradient propagation due to replacing the inner-loop adaptation with prototype construction.

\begin{figure}[h]
\centering
\begin{subfigure}{.5\textwidth}
  \centering
  \includegraphics[clip, trim={1.3cm 0.5cm 1.5cm 1.35cm}, width=7cm]{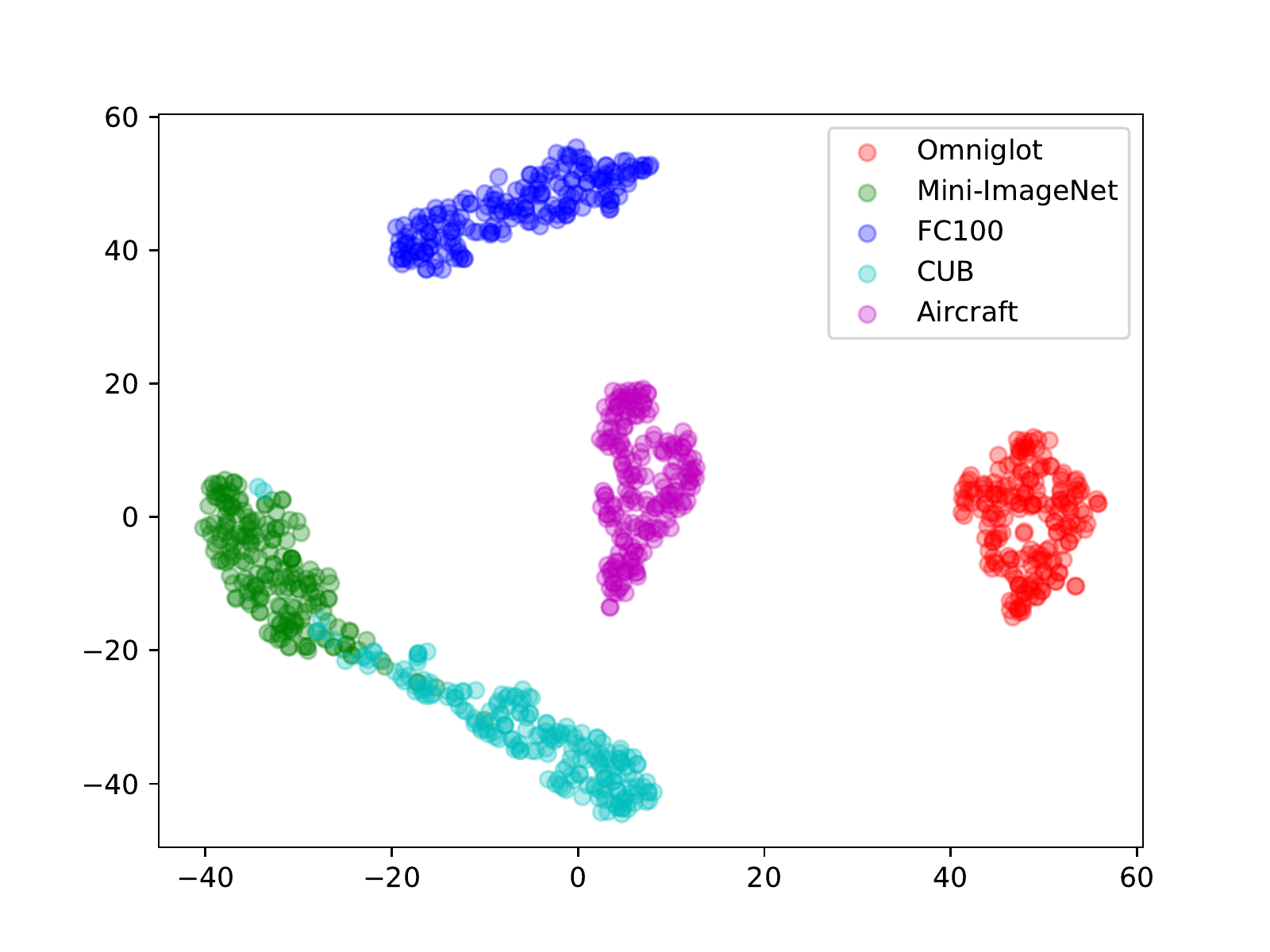}
  \caption{MMAML}
  \label{fig:sfig1}
\end{subfigure}%
\begin{subfigure}{.5\textwidth}
  \centering
  \includegraphics[clip, trim={1.3cm 0.5cm 1.5cm 1.35cm}, width=7cm]{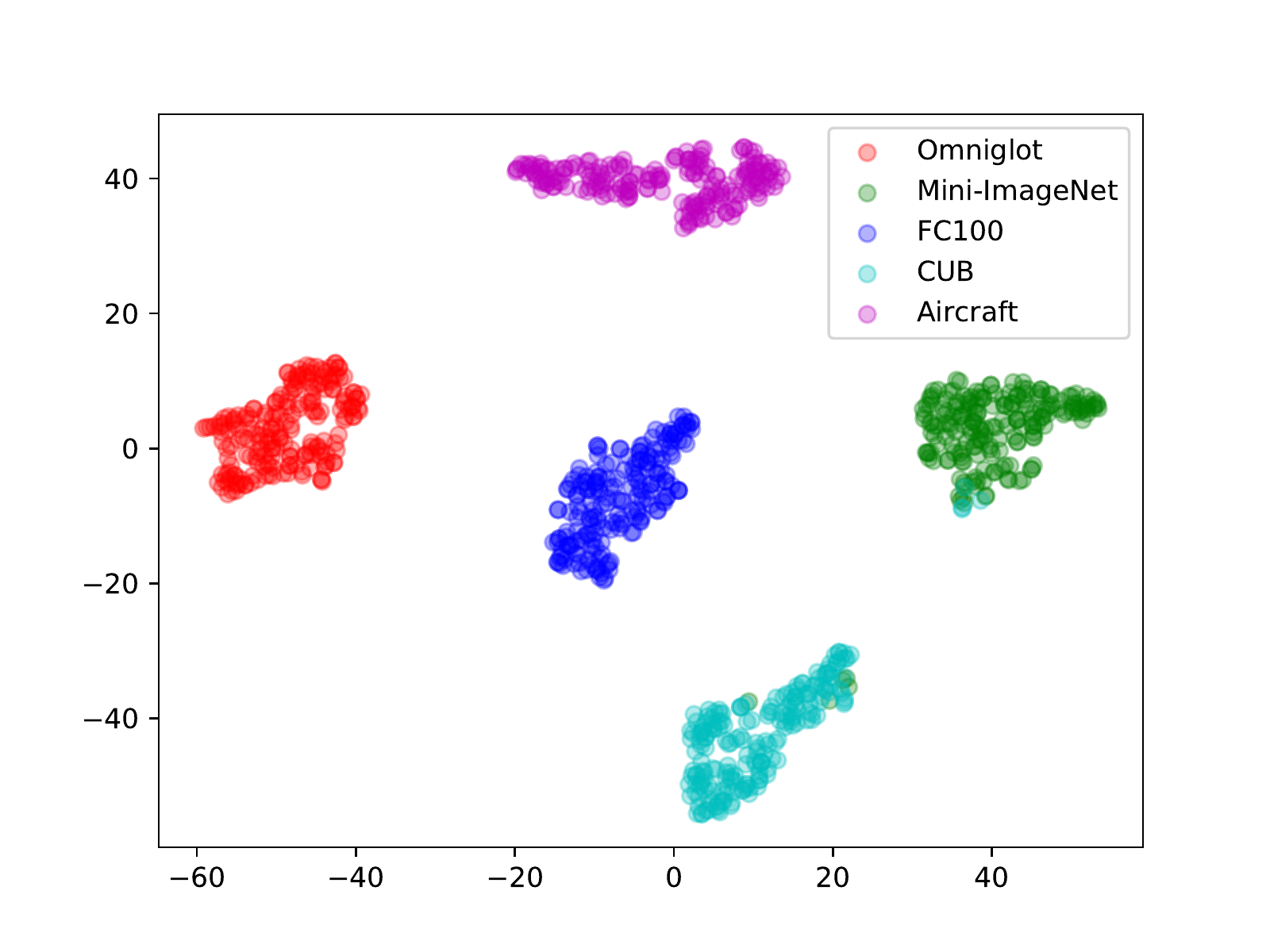}
  \caption{MProtoNet}
  \label{fig:sfig1}
\end{subfigure}
\caption{t-SNE plots for task embedding vectors $\boldsymbol{\upsilon}$ produced for 1000 randomly generated test tasks in 5Mode, 5-way, 1-shot setup by MMAML and MProtoNet.}
\label{fig:t-SNE}
\end{figure}

{\bf Experiment.} To gain a better understanding on the effective modulation network training, we randomly sample 1000 5-mode, 5-way 1-shot meta-test tasks and calculate the task embeddings $\boldsymbol{\upsilon}$ for each task using both MMAML and MProtoNet. Then we employ the t-SNE ~\cite{van2008visualizing} to visualize $\boldsymbol{\upsilon}$ in Figure \ref{fig:t-SNE}. As one can see, in the task embeddings produced by MProtoNet, the embeddings for different modes are separated better.

\section{Experimental Details}
\label{sec:experimental_details}

\subsection{Meta-Dataset}
\begin{figure}[h]
 \begin{center}
 \includegraphics[width=14cm]{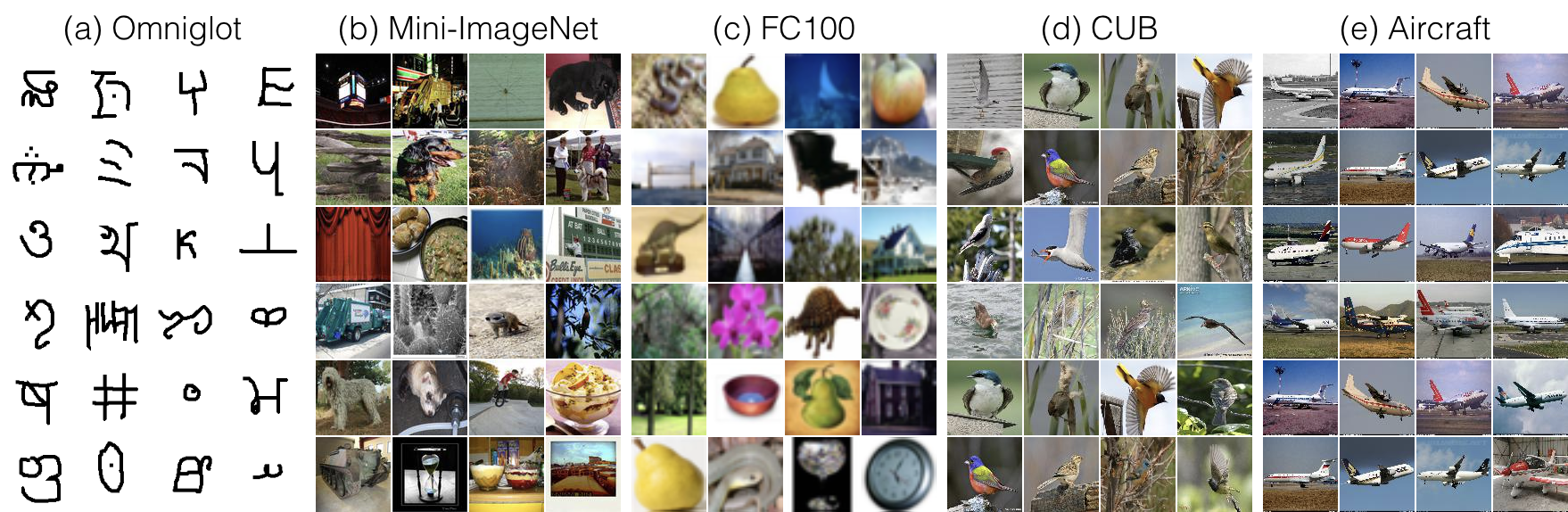}
 \end{center}
   \caption{Examples of images used to create multimodal meta-dataset.}
\label{fig:meta-dataset}
\end{figure}

To create a meta-dataset for multi-modal few-shot classification, we utilize five popular datasets: OMNIGLOT, MINI-IMAGENET, FC100, CUB, and AIRCRAFT. The detailed information of all the datasets are summarized in Table \ref{table:datasets}. To fit the images from all the datasets to a model, we resize all the images to 84 × 84. The images randomly sampled from all the datasets are shown in Figure \ref{fig:meta-dataset}, demonstrating a diverse set of modes.

\begin{table}[h]
	\caption{Details of Datasets. }
	\label{table:datasets}
	\begin{center}
		\resizebox{0.99\linewidth}{!}{
		\begin{tabular}{ccccccc}
		\toprule
        \textbf{Dataset}&
        \textbf{Train Classes}&
        \textbf{Validation Classes}&
        \textbf{Test Classes}&
        \textbf{Image Size}&
        \textbf{Image Channel}&
        \textbf{Image Content}\\
        \midrule
        
        Omniglot&
        4112&
        688&
        1692&
        28$\times$28&
        1&
        handwritten digits\\
        
        mini-ImageNet&
        64&
        16&
        20&
        84$\times$84&
        3&
        natural objects\\
        
        FC100&
        64&
        16&
        20&
        32$\times$32&
        3&
        natural objects\\
        
        CUB&
        140&
        30&
        30&
        $\sim$500$\times$500&
        3&
        species of birds\\
        
        Aircraft&
        70&
        15&
        15&
        $\sim$ 1-2 Mpixels &
        3&
        types of aircrafts\\
        
		\bottomrule
		\end{tabular}
		}
	\end{center}
\end{table}

\subsection{Meta-Learning Algorithms in Multimodal Framework}
Here we provide more details about the meta-learners used in experiments including: MAML ~\cite{finn2017model} and ProtoNet ~\cite{snell2017prototypical}.

\paragraph{MAML.} 
MMAML uses the model-agnostic meta-learning (MAML) algorithm ~\cite{finn2017model} as meta-learner in base network. Given a network $f_{\theta}$, MAML aims to learn a common initialization for weights $\theta$ under a certain task distribution such that it can adapt to new unseen task with a few steps of gradient descent. In multimodal scenario of MMAML, for each task $\mathcal{T}_i$, first the modulation parameters $\boldsymbol{\tau}_{i}$ are generated. Then base network uses these parameters together with the samples from support set $\mathcal{S}_{i}$ to adapt the network weights to that task using gradient descent:

\begin{align*}
    \theta_{i}^{\prime} = \theta - \alpha \nabla_{\theta} \mathcal{L}_{i} (f_{\theta} ({\mathcal{S}_i};\boldsymbol{\tau}_i))   
\end{align*}

Where $\mathcal{L}_{i} (f_{\theta} ({\mathcal{S}_i};\boldsymbol{\tau}_i))$ denotes the loss function on the support set of task $\mathcal{T}_i$. Then the adapted model is evaluated on the query samples of the same task $\mathcal{Q}_i$ which provides the feedback in the form of loss of gradients for generalization performance on that task. The feedback from a batch of tasks is used to update the base network weights $\theta$ to achieve better generalization:

\begin{align*}
    \label{eq:MAML_outer}
    \theta \leftarrow \theta - \beta \nabla_{\theta} \sum_{\mathcal{T}_i} \mathcal{L}_{i} (f_{\theta_{i}^{\prime}} ({\mathcal{Q}_i}))
\end{align*}

The modulation network parameters $(\phi, \varphi)$ are also updated in a same manner using the feedback from the generalization performance of the adapted model. 

\paragraph{ProtoNet.} As another multimodal meta-learner, we replace the ProtoNet with MAML due to its ease of training and better gradient flow.
In this algorithm, first modulation parameters are generated by processing task samples. Then network parameters are modulated using these parameters to produce modulated parameters ${\hat{\theta}_{\mathcal{T}}}$ \footnote{Note that, while in our KML algorithm these parameters are produced explicitly, in MProtoNet (variant of MMAML produced in this paper) these are implicitly generated by applying FiLM~\cite{perez2018film} on each featuremap.}.
ProtoNet computes a \emph{prototype} for each class through an embedding function $f_{\hat{\theta}_{\mathcal{T}}}:\mathbb{R}^D \rightarrow \mathbb{R}^F$ which maps an input sample $\mathbf{x} \in \mathbb{R}^D$ to an $F$-dimensional feature space. The prototype $\mathbf{c}_n$ of a class $n=1, \dots , N$, is the mean vector of the embedded support samples belonging to that class: 
\begin{align*}
\mathbf{c}_n= \frac{1}{\lvert \mathcal{S}^n_{\mathcal{T}} \rvert } \sum_{x \in \mathcal{S}^n_{\mathcal{T}}} f_{\hat{\theta}_{\mathcal{T}}}(\mathbf{x})
\end{align*}

After generating prototypes, it uses a distance function to produce the distribution over classes for each query sample $\mathbf{\Tilde{x}}$ as follows:

\begin{align*}
    p_{n}(\mathbf{\Tilde{x}}) = \frac{exp(-d(f_{\hat{\theta}_{\mathcal{T}}}(\mathbf{\Tilde{x}}),\mathbf{c}_n))} {\sum_ {n^{\prime}} exp(-d(f_{\hat{\theta}_{\mathcal{T}}}(\mathbf{\Tilde{x}}),\mathbf{c}_n^{\prime}))}
\end{align*}

This predicted distribution is compared with class labels to compute the loss. Then this loss optimized by SGD with respect to network parameters ($\theta, \phi, \varphi$).

\subsection{Network Structures}
\subsubsection{Base Network}

\paragraph{Multimodal Few-shot Classification Experiments.}
In multimodal experiments, for the base network (as meta-learner), we use the exactly same architecture as the MMAML convolutional network proposed in ~\cite{vuorio2019multimodal}. It consists of four convolutional layers with the channel size 32, 64, 128, and 256, respectively. All the convolutional layers have a kernel size of 3 and stride of 2. A batch normalization layer follows each convolutional layer, followed by ReLU. With the input tensor size of $(n \cdot k) \times 84 \times 84 \times 3$ for an {\em n-way, k-shot} task, the output feature maps after the final convolutional layer have a size of $(n \cdot k) \times 6 \times 6 \times 256$. 
For ProtoNet-based architectures, these featuremaps are directly used for constructing prototypes and performing classification. For MAML-based architectures, the featuremaps are average pooled along spatial dimensions, resulting feature vectors with a size of $(n \cdot k) \times 256$. In this case, a linear fully-connected layer takes the feature vector as input, and produce a classification prediction with a size of \emph{n} for {\em n-way} classification task.

\paragraph{Unimodal Few-Shot Classification Experiments.}
In conventional unimodal few-shot classification, we use the more standard architecture which is slightly different from the one used in MMAML for multimodal scenario. Here, for ProtoNet-based experiments, following the original implementation of the ProtoNet ~\cite{snell2017prototypical}, 4 similar convolutional blocks are used. Each block comprises a 64-filter 3$\times$3 convolution, batch normalization layer, a ReLU nonlinearity and a 2$\times$2 max-pooling layer. As also discussed in ~\cite{vuorio2019multimodal}, the slight difference between the multimodal results and unimodal ones reported in previous works is due to the difference in network structure and hyperparameters.

\subsubsection{Modulation Network}
The modulation network includes a task encoder network $h_{\varphi}$ and a modulation parameter generator network $g_{\phi}$.

\paragraph{Task Encoder.}
Similar to ~\cite{vuorio2019multimodal},  for the task encoder, we use the exactly same architecture as the base network. It consists of four convolutional layers with the channel size 32, 64, 128, and 256, respectively. All the convolutional layers have a kernel size of 3, stride of 2, and use valid padding. A batch normalization layer follows each convolutional layer, followed by ReLU. With the input tensor size of $(n \cdot k) \times 84 \times 84 \times 3$ for a n-way k-shot task, the output feature maps after the final convolutional layer have a size of $(n \cdot k) \times 6 \times 6 \times 256$. The feature maps are then average pooled along spatial dimensions, resulting feature vectors with a size of $(n\cdot k)\times256$. To produce an aggregated embedding vector from all the feature vectors representing all samples, we perform an average pooling, resulting a feature vector with a size of 256. Finally, a fully-connected layer followed by ReLU takes the feature vector as input, and produce a task embedding vector $\upsilon$ with a size of 128.

\paragraph{Modulation Parameter Generator.} Modulation parameter generator structure varies based on the multimodal algorithm. For MMAML and MProtoNet, we follow the design in ~\cite{vuorio2019multimodal}. In this design, the modulating each channel requires producing two parameters ($\eta_{i}$ for scaling and $\gamma_{i}$ for shifting featuremap). Considering the channel size 32, 64, 128 and 256 in base network, four linear fully-connected layers are used to convert task embedding vector $\upsilon$ (with a size of 128) to required modulation parameters. The size of these layers are as follows: $128\times64$, $128\times128$, $128\times256$ and $128\times512$ \footnote{For unimodal experiments, these numbers are changed based on the number of filters in the base network.}. For KML-based meta-learners (MMAML+KML and MProtoNet+KML), we produce a modulation number for each parameter in the network using the proposed simplified structure. The details explanation on the number of required parameters are discussed in Sec. \ref{sec:param_reduction} of this supplementary material.

\begin{table}[h]
\caption{Hyperparameters used in the experiments. \textsuperscript{$\dag$} Halve every 10K Iterations.}
\label{table:hyperparameters}
\begin{center}
	\resizebox{0.99\linewidth}{!}{
		\begin{tabular}{lcccccc}
		
		\toprule
		
        \textbf{Method}&
        \textbf{DataSet group}&
        \textbf{Inner lr}&
        \textbf{Outer lr}&
        \textbf{Meta batch-size}&
        \textbf{Number of updates}&
        \textbf{Training Iterations}\\
        \midrule
        
        \textbf{MMAML}&
        --&
        0.05& 
        0.001&
        10&
        5&
        60000\\
        \midrule
        
        \textbf{MMAML+KML(ours)}&
        --&
        0.05&
        0.001&
        10&
        5&
        60000\\
        \midrule
        
        \multirow{2}{*}{\textbf{Multi-MAML}}&
        Grayscale&
        0.4&
        0.001&
        10&
        1&
        60000\\
        
        &
        RGB&
        0.01&
        0.001&
        4&
        5&
        60000\\
        
        \bottomrule
        \textbf{MProtoNet}&
        --&
        --&
        0.001\textsuperscript{$\dag$}&
        10&
        --&
        30000\\
        
        \midrule
        \textbf{MProtoNet+KML(ours)}&
        --&
        --&
        0.001\textsuperscript{$\dag$}&
        10&
        --&
        30000\\
        
        \midrule
        \multirow{2}{*}{\textbf{Multi-ProtoNet}}&
        Grayscale&
        --&
        0.001\textsuperscript{$\dag$}&
        10&
        --&
        30000\\
        
        &
        RGB&
        --&
        0.001\textsuperscript{$\dag$}&
        4&
        --&
        30000\\
        
		\bottomrule
		\end{tabular}
		}
\end{center}
\end{table}

\subsubsection{Hyperparameters}
The hyperparameters for all the experiments are shown in Table \ref{table:hyperparameters}. For comparing our algorithm with previous work, we use exactly the same hyperparameters.
We use 15 examples per class for evaluating the post-update meta-gradient for all the experiments, following ~\cite{finn2017model, Ravi2017OptimizationAA, vuorio2019multimodal, snell2017prototypical}. In the training of all networks, we use the Adam optimizer with default hyperparameters.

\section{Information for checklist}
\label{sec:limitations}

{\bf Limitations:} We have followed the procedures in  \cite{vuorio2019multimodal} to construct multimodal datasets in our experiments for fair comparison with their work.
Specifically, the following popular datasets have been used and we have also followed the combination procedures discussed in \cite{vuorio2019multimodal}: 
OMNIGLOT, MINI-IMAGENET, FC100, CUB, and AIRCRAFT.
Potentially, additional datasets can be used in experiments to further demonstrate our ideas.
Furthermore, our work has focused on few shot image classification. Our proposed ideas could be applicable to other few shot learning problems. 

{\bf Broader Impact:} 
Multimodal meta-learning is an extension of conventional few-shot meta-learning. 
Importantly, it mimics humans' ability to acquire a new skill via prior knowledge of a set of diverse skills. 
Research findings in this problem are very meaningful and important  in machine learning.
Furthermore, few shot classification studies the problem  
to classify samples from novel categories given only a few labeled data 
from each category.
The setup is significantly different from other modern deep learning problems, but important for many domains where labeled data is difficult to obtain. For example, in clinical disease diagnosis, data needs to be labeled by medical experts and labeled data is expensive to obtain.

{\bf Amount of compute:} 
All the results in this paper are produced by a machine with a single RTX 2080 Ti GPU. The amount of compute in this project is documented in Table~\ref{table:carbon}. We follow submission guidelines to include the amount of compute for different experiments and CO2 emission.

\begin{table}[h]
    \centering
    \caption{Amount of compute in this project. The GPU hours include computations for early explorations and experiments to produce the reported values. The carbon emission values are computed using \url{https://mlco2.github.io/}.}
        
    \resizebox{\linewidth}{!}{
        \begin{tabular}{lccc}\toprule
        {\bf Experiment} & {\bf Hardware} & {\bf GPU hours} & {\bf Carbon emitted in kg}  \\ 
        
        \toprule
        
        Multimodal Classification Results: Main paper Table 1  & RTX 2080 Ti & 480 & 51.84  \\ 
        \midrule
        
        Unimodal Classification Results: Main paper Table 3  & RTX 2080 Ti & 36 & 3.89  \\ 
        
        \midrule
        
        Visualization: Main paper Figure 1, Figure 2 and supplementary & RTX 2080 Ti & 5 & 0.54  \\ 
        
        \midrule
        
        Hard Parameter Sharing: Supplementary Table 1 and Table 2  & RTX 2080 Ti & 90 & 9.72  \\ 
        \midrule
        
        Verification of new-MMAML interpretation: Supplementary Table 3 & RTX 2080 Ti & 45 & 4.86  \\ 
        
        \midrule
        
        Visualization: Supplementary t-SNE plot & RTX 2080 Ti & 10 & 1.08  \\
        \bottomrule
        \end{tabular}

    }
    \label{table:carbon}
\end{table}

\end{document}